\documentclass{article}
\usepackage{algorithmic}

\usepackage{etoolbox}
\newtoggle{neurips}
\togglefalse{neurips}
\newcommand{\neurips}[1]{\iftoggle{neurips}{#1}{}}
\newcommand{\arxiv}[1]{\iftoggle{neurips}{}{#1}}

\neurips{
  \PassOptionsToPackage{numbers, compress, square}{natbib}
  \usepackage[]{arXiv-2407.18676v2/neurips_2024}
  \usepackage[numbers]{natbib}
  \bibliographystyle{abbrvnat}
}
\usepackage{caption}

\arxiv{
\usepackage[letterpaper, left=1in, right=1in, top=1in,
bottom=1in]{geometry}
  \usepackage{parskip}
}

\neurips{
  \usepackage{parskip}
  }

\PassOptionsToPackage{hypertexnames=false}{hyperref}  %

\usepackage[svgnames]{xcolor}
\usepackage[colorlinks=true, linkcolor=blue!70!black, citecolor=blue!70!black,urlcolor=black,breaklinks=true]{hyperref}
\colorlet{txblue}{RoyalBlue!70!NavyBlue}
\hypersetup{linkcolor=txblue,
            citecolor=txblue}
\usepackage{microtype}
\usepackage{hhline}

\usepackage{amsthm}
\usepackage{mathtools}
\usepackage{amsmath}
\usepackage{bbm}
\usepackage{amsfonts}
\usepackage{amssymb}
\usepackage[nameinlink,capitalize]{cleveref}

\makeatletter
\newcommand{\neutralize}[1]{\expandafter\let\csname c@#1\endcsname\count@}
\makeatother

\newtheorem*{lemma*}{Lemma}

\usepackage{algorithm}

\arxiv{
\usepackage{natbib}
\bibliographystyle{plainnat}
\bibpunct{(}{)}{;}{a}{,}{,}
}

\usepackage{xpatch}

\usepackage{thmtools}
\usepackage{thm-restate}
\declaretheorem[name=Theorem,parent=section]{theorem}
\declaretheorem[name=Lemma,parent=section]{lemma}
\declaretheorem[name=Assumption, parent=section]{assumption}
\declaretheorem[name=Condition, parent=section]{condition}

\declaretheorem[name=Remark, parent=section]{remark}
\declaretheorem[name=Proposition, parent=section]{proposition}

\usepackage{crossreftools}
\makeatletter
  \renewenvironment{proof}[1][Proof]%
  {%
   \par\noindent{\bfseries\upshape {#1.}\ }%
  }%
  {\qed\newline}
  \makeatother

\theoremstyle{definition}  %

\newtheorem{claim}{Claim}

\newtheorem{corollary}{Corollary}[section]

\theoremstyle{plain}
\newtheorem{definition}{Definition}[section]

\xpatchcmd{\proof}{\itshape}{\normalfont\proofnameformat}{}{}
\newcommand{\proofnameformat}{\bfseries}

\renewcommand{\eqref}[1]{\texorpdfstring{\hyperref[#1]{(\ref*{#1})}}{(\ref*{#1})}}
\crefformat{equation}{#2Eq.\,(#1)#3}
\Crefformat{equation}{#2Eq.\,(#1)#3}

\Crefformat{figure}{#2Figure~#1#3}
\Crefformat{assumption}{#2Assumption~#1#3}

\Crefname{assumption}{Assumption}{Assumptions}

\crefname{fact}{Fact}{Facts}

\Crefformat{figure}{#2Figure #1#3}
\Crefformat{assumption}{#2Assumption #1#3}

\usepackage{crossreftools}
\usepackage{xparse}

\ExplSyntaxOn
\DeclareDocumentCommand{\XDeclarePairedDelimiter}{mm}
 {
  \__egreg_delimiter_clear_keys: %
  \keys_set:nn { egreg/delimiters } { #2 }
  \use:x %
   {
    \exp_not:n {\NewDocumentCommand{#1}{sO{}m} }
     {
      \exp_not:n { \IfBooleanTF{##1} }
       {
        \exp_not:N \egreg_paired_delimiter_expand:nnnn
         { \exp_not:V \l_egreg_delimiter_left_tl }
         { \exp_not:V \l_egreg_delimiter_right_tl }
         { \exp_not:n { ##3 } }
         { \exp_not:V \l_egreg_delimiter_subscript_tl }
       }
       {
        \exp_not:N \egreg_paired_delimiter_fixed:nnnnn 
         { \exp_not:n { ##2 } }
         { \exp_not:V \l_egreg_delimiter_left_tl }
         { \exp_not:V \l_egreg_delimiter_right_tl }
         { \exp_not:n { ##3 } }
         { \exp_not:V \l_egreg_delimiter_subscript_tl }
       }
     }
   }
 }

\keys_define:nn { egreg/delimiters }
 {
  left      .tl_set:N = \l_egreg_delimiter_left_tl,
  right     .tl_set:N = \l_egreg_delimiter_right_tl,
  subscript .tl_set:N = \l_egreg_delimiter_subscript_tl,
 }

\cs_new_protected:Npn \__egreg_delimiter_clear_keys:
 {
  \keys_set:nn { egreg/delimiters } { left=.,right=.,subscript={} }
 }

\cs_new_protected:Npn \egreg_paired_delimiter_expand:nnnn #1 #2 #3 #4
 {%
  \mathopen{}
  \mathclose\c_group_begin_token
   \left#1
   #3
   \group_insert_after:N \c_group_end_token
   \right#2
   \tl_if_empty:nF {#4} { \c_math_subscript_token {#4} }
 }
\cs_new_protected:Npn \egreg_paired_delimiter_fixed:nnnnn #1 #2 #3 #4 #5
 {
  \mathopen{#1#2}#4\mathclose{#1#3}
  \tl_if_empty:nF {#5} { \c_math_subscript_token {#5} }
 }
\ExplSyntaxOff

\XDeclarePairedDelimiter{\supnorm}{
  left=\lVert,
  right=\rVert,
  subscript=\infty
  }

\let\Pr\undefined

\DeclareMathOperator{\Pr}{Pr}

\def\ddefloop#1{\ifx\ddefloop#1\else\ddef{#1}\expandafter\ddefloop\fi}
\def\ddef#1{\expandafter\def\csname bb#1\endcsname{\ensuremath{\mathbb{#1}}}}
\ddefloop ABCDEFGHIJKLMNOPQRSTUVWXYZ\ddefloop
\def\ddefloop#1{\ifx\ddefloop#1\else\ddef{#1}\expandafter\ddefloop\fi}
\def\ddef#1{\expandafter\def\csname b#1\endcsname{\ensuremath{\mathbf{#1}}}}
\ddefloop ABCDEFGHIJKLMNOPQRSTUVWXYZ\ddefloop
\def\ddef#1{\expandafter\def\csname sf#1\endcsname{\ensuremath{\mathsf{#1}}}}
\ddefloop ABCDEFGHIJKLMNOPQRSTUVWXYZ\ddefloop
\def\ddef#1{\expandafter\def\csname c#1\endcsname{\ensuremath{\mathcal{#1}}}}
\ddefloop ABCDEFGHIJKLMNOPQRSTUVWXYZ\ddefloop
\def\ddef#1{\expandafter\def\csname h#1\endcsname{\ensuremath{\widehat{#1}}}}
\ddefloop ABCDEFGHIJKLMNOPQRSTUVWXYZ\ddefloop
\def\ddef#1{\expandafter\def\csname hc#1\endcsname{\ensuremath{\widehat{\mathcal{#1}}}}}
\ddefloop ABCDEFGHIJKLMNOPQRSTUVWXYZ\ddefloop
\def\ddef#1{\expandafter\def\csname t#1\endcsname{\ensuremath{\widetilde{#1}}}}
\ddefloop ABCDEFGHIJKLMNOPQRSTUVWXYZ\ddefloop
\def\ddef#1{\expandafter\def\csname tc#1\endcsname{\ensuremath{\widetilde{\mathcal{#1}}}}}
\ddefloop ABCDEFGHIJKLMNOPQRSTUVWXYZ\ddefloop
\def\ddefloop#1{\ifx\ddefloop#1\else\ddef{#1}\expandafter\ddefloop\fi}
\def\ddef#1{\expandafter\def\csname scr#1\endcsname{\ensuremath{\mathscr{#1}}}}
\ddefloop ABCDEFGHIJKLMNOPQRSTUVWXYZ\ddefloop

\input{arXiv-2407.18676v2/widebar}

\usepackage{hyperref}       
\usepackage{url}            
\usepackage{natbib}         
\usepackage[english]{babel}
\usepackage{graphicx}       

\usepackage{arXiv-2407.18676v2/math_commands}  
\usepackage{caption}
\usepackage[toc,page,header]{appendix}
\usepackage{minitoc}

\title{DP‑NCB: Privacy Preserving Fair Bandits}

\author{
  Dhruv Sarkar\thanks{Indian Institute of Technology Kharagpur, \texttt{dhruv.sarkar@kgpian.iitkgp.ac.in}} 
  \and
  Nishant Pandey\thanks{Indian Institute of Technology Kanpur, \texttt{nishantp22@iitk.ac.in}} 
  \and
  Sayak Ray Chowdhury\thanks{Indian Institute of Technology Kanpur, \texttt{sayakrc@iitk.ac.in}} 
}

\usepackage{amssymb,amsfonts,amsmath,amsthm} 
\usepackage{amsmath}
\usepackage{amsfonts} 
\usepackage{mathtools} 
\usepackage{bm} 
\usepackage{mathrsfs}
\usepackage{thmtools}
\usepackage{thm-restate}

\usepackage{enumerate}
\usepackage{comment}
\usepackage{multirow}
\usepackage{xcolor}         
\usepackage{xfrac}

\usepackage{subcaption}

\begin{document}

\maketitle

\begin{abstract}
Multi-armed bandit algorithms are fundamental tools for sequential decision-making under uncertainty, with widespread applications across domains such as clinical trials and personalized decision-making. As bandit algorithms are increasingly deployed in these socially sensitive settings, it becomes critical to protect user data privacy and ensure fair treatment across decision rounds.
While prior work has independently addressed privacy and fairness in bandit settings, the question of whether both objectives can be achieved simultaneously has remained largely open. Existing privacy-preserving bandit algorithms typically optimize average regret, a utilitarian measure, whereas fairness-aware approaches focus on minimizing Nash regret, which penalizes inequitable reward distributions, but often disregard privacy concerns.

To bridge this gap, we introduce Differentially Private Nash Confidence Bound (DP-NCB)—a novel and unified algorithmic framework that simultaneously ensures $\epsilon$-differential privacy and achieves order-optimal Nash regret, matching known lower bounds up to logarithmic factors. The framework is sufficiently general to operate under both global and local differential privacy models, and is anytime, requiring no prior knowledge of the time horizon. We support our theoretical guarantees with simulations on synthetic bandit instances, showing that DP-NCB incurs substantially lower Nash regret than state-of-the-art baselines. Our results offer a principled foundation for designing bandit algorithms that are both privacy-preserving and fair, making them suitable for high-stakes, socially impactful applications.
\end{abstract}

\section{Introduction}

The multi-armed bandit framework \citep{bubeck2012regret,lattimorebanditalgorithms} is a foundational model for sequential decision-making under uncertainty. It has been widely applied across domains such as online advertising and product recommendations \citep{li2010contextual,schwartz2017customer}, education and tutoring systems \citep{clement2015multi}, and healthcare and clinical trials \citep{tewari2017ads,villar2015multi}.
Given its growing use in socially sensitive areas, there is increasing interest in integrating principles of privacy and fairness into bandit algorithms. Privacy concerns arise from the reliance on user feedback, which can inadvertently reveal sensitive information \citep{pan2019you}. Fairness is motivated by social welfare considerations — ensuring that all users perceive the system as treating them equitably \citep{moulin2004fair}.

To highlight the importance of privacy and fairness in bandit problems, consider a clinical trial scenario originally proposed by \citet{thompson1933likelihood}: given $k$ drugs and $T$ patients, the decision maker selects one drug to administer to the $t$-th patient in each round $t \le T$. The treatment outcome is inherently private and sensitive i.e. patients may not wish to disclose their medical conditions publicly post-trial. However, since future drug assignments are based on prior outcomes, an individual patient’s response can influence subsequent decisions. This means that even if outcomes are not explicitly revealed, they may still be inferred from changes in the drug selection policy. Therefore, the learning algorithm must ensure that individual patient data (e.g., treatment outcomes) remains private while taking future decisions (e.g., drug choices). At the same time, it is crucial to ensure that no patient is ex-ante disadvantaged by the learner, i.e., fairness must be maintained across rounds. A fair learning algorithm should guarantee that patients participating in different rounds of the trial benefit from progressively better drug choices, while still permitting sufficient exploration to accurately identify the most effective drug.

Achieving privacy in isolation is straightforward, for example, the learner could ignore treatment outcomes entirely and always administer a fixed, publicly announced drug. However, such an approach would likely lead to poor treatment efficacy. This highlights the need to carefully balance privacy with utility. Differential Privacy (DP) offers a principled way to do so by providing strong guarantees even against adversaries with access to arbitrary auxiliary information. It achieves this by adding calibrated random noise to obscure any individual’s influence on the algorithm's output, ensuring that no single user's data significantly alters the outcome \citep{dwork2014}. Importantly, DP includes a tunable parameter $\epsilon > 0$ that allows explicit control over the privacy-utility trade-off: smaller $\epsilon$ yields stronger privacy but reduced utility, and vice versa.
In recent years, DP has been effectively incorporated into the design of privacy-preserving bandit algorithms \citep{mishra2015nearly,tossou2016algorithms,azize2022privacy}, typically using average reward (e.g., treatment outcome) or, equivalently, regret, as the measure of utility.

Minimizing \textbf{average regret}, which is defined as the difference between the (a priori unknown) highest expected reward and the \emph{arithmetic mean} of the expected rewards obtained by the algorithm—is equivalent to maximizing social welfare in a utilitarian sense \citep{moulin2004fair}. While this objective promotes overall efficiency, it may inadvertently favor short-term reward maximization, potentially resulting in inequities across decision rounds. For example, achieving a high average treatment efficacy does not preclude some patients from receiving highly ineffective treatments.
To address this limitation, Nash social welfare (NSW) considers the \emph{geometric mean} of expected rewards, promoting a more equitable distribution across rounds in an ex-ante sense \citep{moulin2004fair}. The associated metric, \textbf{Nash regret}—the gap between the highest expected reward and the NSW offers a fairness-aware alternative to average regret. It penalizes disparities in individual rewards and has gained recent attention as a foundation for designing fair bandit algorithms \citep{barman2023fairness,sawarni2023nash,krishna2025p}.

While prior work has extensively explored privacy and fairness in multi-armed bandits, these concerns have been addressed in isolation. The fundamental question of whether these concerns can be simultaneously addressed in bandit settings remains largely unresolved. In this work, we address this gap by introducing a novel differentially private Nash regret minimization framework that combines controlled noise addition of DP mechanisms with fairness-aware reward aggregation. This ensures a learning process that is both private and fair, making it well-suited for high stakes, sensitive applications like healthcare and personalized decision-making.

\paragraph{Our Contributions}
\begin{itemize}
    \item We propose a general framework — Differentially Private Nash Confidence Bound (DP-NCB) — that guarantees $\epsilon$-differential privacy for any $\epsilon$, while achieving order-optimal Nash regret that vanishes as the number of rounds $T \to \infty$. 

\item In the global model of differential privacy, where the learner is trusted and has access to raw user data, Algorithm~\ref{algo:gdp-ncb} (GDP-NCB) achieves a Nash regret\footnote{$\widetilde{O}$ hides constant and polylogarithmic factors in $T$ and $k$.} of
$\widetilde O\Big(\sqrt{\frac{k}{T}} +\frac{k}{\epsilon\,T}\Big)$. In the stronger local privacy model, where the learner is not trusted and has access to only randomized user data, Algorithm~\ref{algo:ldp-ncb} (LDP-NCB) achieves the Nash regret $\widetilde O\Big(\sqrt{\frac{k}{T}} +\frac{1}{\epsilon}\sqrt{\frac{k}{T}} \Big)$.

\item Since Nash regret is a stricter benchmark than average regret (see Remark~\ref{rem:comp-regret}), existing lower bounds for average regret also apply to Nash regret. In particular, the known minimax lower bounds—$\Omega \Big( \max \big\lbrace \sqrt{\frac{k}{T}}, \frac{k}{\epsilon\, T} \big\rbrace\Big)$ in the global privacy model and $\Omega\Big(\frac{1}{\epsilon}\sqrt{\frac{k}{T}} \Big)$ in the local model—serve as fundamental performance limits.
This implies that our upper bound on Nash regret is nearly optimal (up to logarithmic factors), as it matches the best-known lower bounds even for the easier benchmark of average regret.

\item We further extend our algorithms to the anytime setting, where prior knowledge of the time horizon $T$ is not required. We show that this generalization incurs only an $O(\log T)$ multiplicative increase in Nash regret in both the global and local privacy models.
\item To empirically validate fairness across rounds, we evaluate the Nash regret of our algorithms under both the global and local privacy models. Our results show that the proposed algorithms incur significantly lower Nash regret compared to state-of-the-art private bandit algorithms designed for average regret, thereby supporting our theoretical claims.
\end{itemize}

\section{Preliminaries}
\label{gen_inst}

\paragraph{Multi-armed Bandit (MAB)}

In the stochastic multi-armed bandit problem, the learner has sample access to $k$ probability distributions (called arms) supported on the interval $[0,1]$. Each arm $i \in [k]:=\{1,\ldots,k\}$ has its mean $\mu_i \in [0,1]$ a-priori unknown to the learner. Let $\mu^* = \max_{i \in [k]} \mu_i$ denote the highest mean. We consider settings where the learning algorithm operates over a population of $T \ge 1$ unique users, one for each round. At each round $t \in [T]:=\{ 1,2,\ldots T\}$, the algorithms selects an arm $I_t \in [k]$ and observes a random reward $X_t$ drawn independently from the distribution with mean $\mu_{I_t}$. $I_t$ is selected based on past history of draws and rewards $\lbrace I_1,X_1,\ldots,I_{t-1},X_{t-1} \rbrace$.

Typically, the learner seeks to maximize the social welfare (arithmetic mean of expected rewards) $\sum_{t=1}^T \mathbb{E}[\mu_{I_t}]$. The average regret $\ARg_T$ measures the algorithm's shortfall relative to the optimal social welfare $\mu^*$, that is
\begin{equation}\label{eq:avg-reg}
    \ARg_T:=\mu^*-\frac{1}{T}\sum\nolimits_{t=1}^T \mathbb{E}[\mu_{I_t}]~.
\end{equation}
The average regret is a utilitarian metric and ignores fairness across rounds (e.g, users). The Nash social welfare (geometric mean of expected rewards)$\big(\prod_{t=1}^T \mathbb{E}[\mu_{I_t}]\big)^{\frac{1}{T}}$ achieves per-round fairness  since for the geometric mean to be large, the expected reward at each round should be large enough. Note that $\mu^*$ is also the optimal Nash social welfare,
which yields the Nash regret
\begin{equation}\label{eq:nash-reg}
    \NRg_T:=\mu^*-\big(\prod\nolimits_{t=1}^T \mathbb{E}[\mu_{I_t}]\big)^{\frac{1}{T}}~.
\end{equation}

\begin{remark}[$\NRg_T \ge \ARg_T$ \citep{barman2023fairness}]
\label{rem:comp-regret}
The AM-GM inequality yields that $\NRg_T \ge \ARg_T$, implying that Nash regret is a stricter metric than average regret. This further implies that any upper bound on $\NRg_T$ is also an upper bound on $\ARg_T$, and any lower bound on $\ARg_T$ is also a lower bound on $\NRg_T$.
\end{remark}

\paragraph{Differential Privacy (DP)}

DP is a rigorous framework that ensures an algorithm's output remains almost the same under a change in one input datum, thereby protecting the sensitive information about any individual data point. Let $\cD$ be the data universe. Two datasets $D, D' \in\cD$
are called neighboring if they differ only in a single datum. The standard definition of DP is as follows \citep{dwork2009complexity}.

\begin{definition}[Differential Privacy (DP)]
For any $\epsilon >0$, a randomized mechanism $\mathcal{M}$ satisfies $\epsilon$-differential privacy if for all neighboring datasets $D, D'$, and for all measurable subsets $S \subseteq \range(\cM)$,
\[
\Pr[\mathcal{M}(D) \in S] \leq e^{\epsilon} \Pr[\mathcal{M}(D') \in S]~.
\]
\end{definition}
\noindent Here $\epsilon$ controls the level of privacy; smaller values of $\epsilon$ imply stronger privacy and vice versa. Differential privacy in bandits is generally studied under two models: \textit{global differential privacy} and \textit{local differential privacy} to protect users' private and sensitive rewards.

In the global model, a trusted server (learning agent) has access to the user's rewards $X_{1:T}=\lbrace X_1,\ldots,X_T \rbrace$. It needs to ensure that its decisions $I_{1:T}=\lbrace I_1,\ldots I_T \rbrace$ are not distinguishable for two neighboring reward sequences. Formally, we call a bandit algorithm over $T$ rounds $\epsilon$-global DP if 
\begin{align*}
   \Pr[I_{1:T}|X_{1:T}] \leq e^{\epsilon} \Pr[I_{1:T}|X'_{1:T}] 
\end{align*}
for every arm sequence $I_{1:T} \in [k]^T$
and every pair of neighboring reward sequences $X_{1:T}, X'_{1:T} \in [0,1]^T$ such that $X_s \neq X'_s$ for some $s \in [T]$ and $X_t = X'_t$ for all $t \in [T]\!\setminus\! \lbrace s\rbrace $.

In contrast, the local model assumes no trusted aggregator or learning agent: each user $t$ perturbs her reward $X_t$ locally using a mechanism $\cM$ before sending it to the agent. The agent decides on which arm to pull based on the perturbed reward $\cM(X_t)$ and the history. Formally, we call a bandit algorithm over $T$ rounds $\epsilon$-local DP if
\begin{align*}
 \forall t \in [T]~,  \Pr[\mathcal{M}(X_t) = z] \leq e^{\epsilon} \Pr[\mathcal{M}(X'_t) = z]
\end{align*}
for every $z \in \mathbb{R}$ and every pair of rewards $X_t,X'_t \in [0,1]$ such that $X_t \neq X'_t$.

Local DP (LDP) offers stronger individual-level privacy than global DP (GDP) but often at the cost of slower learning due to higher noise and lack of centralized access to rewards, leading to worse regret performance. Differential privacy in bandits is a well-explored area and has been studied under both global \citep{azize2022privacy,  tossou2016algorithms} and local \citep{basu2019differential, ren2020localdifferentialprivacy} models. For more details on prior work in DP bandits, see Appendix \ref{app:rel_w}.

\section{A Primer on Algorithm Design}
Before presenting our differentially private algorithms under the global and local privacy models, we first introduce a common framework that underlines both. At the core of our approach is a Nash regret-minimizing dynamic, which guides the learning process. This section outlines a high-level blueprint for how privacy-preserving estimators are integrated into this dynamic. Understanding this foundation will help clarify the algorithmic design choices specific to the GDP and LDP variants. Our algorithms are structured in two main phases:

\paragraph{Phase I (Uniform Exploration)} 
    In this phase, we repeatedly sample each arm uniformly until the number of pulls $n_i$ times the private and randomized empirical mean $\widetilde{\mu_i}$ for some arm $i$ exceeds a threshold that depends on the horizon $T$ and the privacy level $\epsilon$. Specifically, for the GDP setting (Algorithm \ref{algo:gdp-ncb}), we stop Phase I when
    \[
        n_i \,\widetilde{\mu}_i \;\geq\; 1600\Bigl(c^2\log T + \tfrac{(\log T)^2}{\epsilon}\Bigr)
        \,\,\text{for some arm }i~.
    \]
     Similarly, for the LDP setting (Algorithm \ref{algo:ldp-ncb}), Phase I is terminated when the following holds for some arm $i$:
\begin{multline*}
    n_i\widetilde{\mu}_{i}\geq \max \Bigg \lbrace \frac{1}{\epsilon}\sqrt{8 n_i\alpha\log T},\,\, \frac{\sqrt{8n_i\alpha\log T}} {\epsilon} + 1600\Big(c^2\log T+\frac{n_i(\log T)^2}{\left(n_i\widetilde{\mu}_{i}-\frac{1}{\epsilon}\sqrt{8\,n_i\alpha\log T}\right)\epsilon^2}\Big) \Bigg\rbrace~.
\end{multline*}
The stopping conditions are carefully designed to ensure that the number of exploration rounds—while dependent on the underlying bandit instance—does not grow excessively large with high probability, thereby keeping the (Nash) regret due to random exploration in control.
     \paragraph{Phase II (Private Adaptive Exploitation)} In this phase, the algorithm switches to an upper confidence bound (UCB)-style arm selection. To minimize Nash regret,  the standard UCB computation for average regret minimization \citep{bubeck2012regret} is modified 
     in \citet{barman2023fairness} to define the Nash confidence bound at round $t$:
\begin{align}
	\NCB(i,t) & \coloneqq  \widehat\mu_i + 4 \sqrt{\frac{\widehat{\mu_i} \log T}{n_i}} \label{eq:ncb}~,
\end{align}
where $\widehat\mu_i$ is the non-private empirical mean of arm $i$ at round $t$ computed from $n_i$ reward samples. 

To ensure privacy and then to account for different amounts of noise added for privacy, we adjust the NCB differently for GDP and LDP settings. Since the observed rewards lie in $[0,1]$, to ensure GDP, we add Laplace noise with scale $\frac{\log T}{\epsilon n_i}$ to each empirical mean $\widehat \mu_i$ and obtain the private mean $\widetilde \mu_i$. A detailed proof of privacy is given in Appendix \ref{sec:appendix_privacy}. Similarly, to ensure LDP, Laplace noise with scale $1/\epsilon$ is directly added to each observed reward, which leads to a higher amount of overall noise in the private mean $\widetilde \mu_i$. 

For the global setup, we follow the batch arm selection idea of \citet{azize2022privacy}, where a single arm is selected for a fixed duration (denoted by episode $\ell$). To account for the added noise, the modified NCB is computed using the private mean estimate ($\widetilde{\mu}_{i}$) made at the end of the previous episode ($c,\alpha$ are constants):
\begin{align*}     
        \NCB_{\GDP}(i,t) \;=\; \widetilde{\mu}_i 
        \;+\; 2c \sqrt{\frac{2 \,\widetilde{\mu}_i \,\log T}{n_{i}}} 
        \;\\+\; \frac{\alpha (\log T)^2}{\epsilon\,n_{i}}
        \;+ 4\sqrt{\frac{2\alpha}{\epsilon}}\;\frac{(\log T)^{3/2}}{n_{i}}~,
    \end{align*}
We compute $\NCB_\GDP(i,t)$ for each arm $i$ at the beginning of each episode $\ell$, and the arm with the highest NCB is the one that is pulled throughout the episode. At the end of the episode, the number of pulls and the empirical mean are reset to those obtained in Phase I.

For the local setup, we follow the standard sequential arm selection rule, where the modified NCB is computed using the private mean $\widetilde{\mu}_{i}$ at round $t$:
    \begin{align*}
        {\NCB}_{\LDP}(i, t) \;=\; \widetilde{\mu}_i 
        \;+\; 2c \sqrt{\frac{2 \,\widetilde{\mu}_i \,\log T}{n_{i}}} 
        \; +\frac{1}{\epsilon}\sqrt\frac{8\alpha\log T}{n_i}
        \;+ 4c\frac{(2\alpha)^{\frac{1}{4}}(\log T)^{\frac{3}{4}}}{\sqrt{\epsilon}(n_i)^{\frac{3}{4}}}~,
    \end{align*}
and the arm with the highest $\NCB_\LDP(i,t)$ is pulled. Finally, as a post-processing step, we clip the private means to $[0,1]$ to keep them in the desired range for both algorithms. 

The first two terms in our modified indices arise from substituting the empirical mean in the original NCB expression with its randomized and private counterpart. The remaining two terms are carefully constructed to ensure that, with high probability, the overall index remains a valid upper bound on the true mean, despite the added noise. This design is essential to uphold the core principle of the Upper Confidence Bound (UCB) framework, which relies on optimistic estimates—that is, selecting arms based on high-probability overestimates of their expected rewards.
In effect, these additional terms compensate for the injected noise to maintain optimism: they bound the potential deviation due to privacy-induced randomness with probability at least $1 - 1/T^\alpha$, and are instrumental in the transition from private to non-private mean estimates during the regret analysis.

\section{GDP-NCB: Nash Confidence Bound in Global DP}
\label{section:ourAlgorithm}
In this section, we present our globally differentially private algorithm: GDP-NCB. To privatize the mean estimates, we adopt the hybrid mechanism proposed by \citet{chan2011private}. Notably, the overall privacy guarantee of GDP-NCB aligns with that of the mechanism used to compute the private mean for each arm. Moreover, the privacy budget is consumed only when a new private mean is released.

To preserve privacy more effectively, we release private means sparingly, which motivates the introduction of an episodic structure. In this setup, a new private mean is released only once per episode, rather than at every round. While this improves privacy, it introduces a trade-off: the released estimate may not reflect the most recent rewards, potentially affecting performance. Moreover, we compute each private mean using only the rewards collected during the current episode. This ensures that the sensitivity of the mean remains episode-local, preventing cumulative leakage over time and yielding stronger privacy guarantees.

Algorithm~\ref{algo:gdp-ncb} incorporates these design choices. Specifically, $N_{1,i}$ denotes the number of times arm $i$ is pulled during Phase I, while $N_{2,i}$ counts the number of pulls for arm $i$ within an episode during Phase II. The empirical and private means in Phase I are denoted by $\widehat{\mu}$ and $\widetilde{\mu}$, respectively, and the empirical mean in episode $\ell$ of Phase II is denoted by $\widehat{\mu}^\ell$. The following theorem provides a bound on the Nash regret.

\begin{algorithm}[t]
    \caption{GDP-NCB }
    \label{algo:gdp-ncb}
    \textbf{Input:} Number of arms $k$, horizon $T$, privacy budget $\epsilon$\\
    \vspace{-10pt}
    \begin{algorithmic}[1]
	\STATE \textbf{Initialization:} $\widehat{\mu}_i = 0$, $\widetilde{\mu}_i = 0$, $N_{1,i} =0$ $\ \forall \ i \in [k]$;  $\ t=1$, $c=3$, $\alpha = 3.1$ 
	\label{step:budget}
		\\ \textbf{Phase {\rm I} - Uniform Exploration}
        \WHILE{ $\max_{i}{n_i \ \widetilde{\mu}_i} \leq 1600\bigg(c^2\log T+\frac{(\log T)^2}{\epsilon}\bigg)$ and $t \leq T$} \label{step:PhaseOneAlgTwo}
		\STATE Pull arm $I_t \sim \Unif([k]), \;$ observe $X_t$
		  \STATE Update $N_{1,I_t} \leftarrow N_{1,I_t} + 1$, $\widehat{\mu}_{I_t} \leftarrow \frac{N_{1,I_t}-1}{N_{1,I_t}}\widehat{\mu}_{I_t} + \frac{X_t}{N_{1,I_t}}$
            \STATE Update $\widetilde{\mu}_{I_t} \gets \widehat{\mu}_{I_t}+ \Lap\left(\frac{ \log T}{\epsilon N_{1, I_t}}\right)$
            \STATE Update $t\gets t + 1$
        \ENDWHILE
		\\ \textbf{Phase {\rm II} - Private Adaptive Exploration}
        \STATE Set $N_{2,i}(t) = 1$ $\forall \ i \in [k]$
	\FOR{episode $\ell = 1, 2, \dots$}             
            \STATE Let $t_{\ell} = t + 1$
            \STATE Compute $A = \operatorname{argmax}_{i \in [k]} \NCB_\GDP(i,t_\ell-1)$ 
            \STATE Compute $n_s$ = $N_{2,A}(t_{\ell} - 1)$
            \STATE Update $\widehat{\mu}_{A}^\ell\leftarrow \widehat{\mu}_{A}, N_{2, A} \leftarrow 0$
		\FOR{$m=1$ to $2n_s$}
			\STATE Pull arm $A, \;$ observe $X_t$ 
            \STATE Update $N_{2,A} \leftarrow N_{2,A} + 1$
			\STATE Compute $n_{A} = N_{2,A}+ N_{1,A}$
			\STATE Update $\widehat{\mu}_{A}^\ell\leftarrow \frac{n_{A}-1}{n_{A}}\widehat{\mu}_{A}^\ell + \frac{X_t}{n_{A}}$
		\ENDFOR
        \STATE Update $\widetilde{\mu}_{A}\leftarrow \widehat{\mu}_{A}^\ell + \Lap\left(\frac{\log T}{\epsilon n_{A}}\right)$ \label{step:laplace}
        \STATE Clip private mean $\widetilde{\mu}_{A}=\min\{1, \max\{0, \widetilde{\mu}_{A}\}\}$
            
        \ENDFOR
    \end{algorithmic}
\end{algorithm}







\newtheorem{repthm}{Theorem}
\begin{repthm}[Nash regret in global model]
\label{theorem:improvedNashRegret}
Fix time horizon $T \in \mathbb{N}$ and privacy budget $\epsilon > 0$. Then, for the $k$-armed bandit problem, GDP-NCB enjoys the Nash regret
\[
\NRg_T \;=\;
O\Bigg(\sqrt{\frac{k\log T}{\,T\,}} \;+\;\underbrace{\frac{k(\log T)^2}{\epsilon\,T}}_{\text{Privacy Cost}}\Bigg).
\]
\end{repthm}

\begin{remark}[Privacy cost] 
The cost for privacy appears in the additive term $\frac{k (\log T)^2}{\epsilon T}$. Notably, it only becomes dominant  when $\epsilon \leq \widetilde{O}\left(\sqrt{\frac{k}{T}}\right)$. This indicates that in the low-privacy regime (i.e., when $\epsilon$ is relatively large), the impact of privacy on the Nash regret is minimal. Moreover, since the privacy cost goes down as $1/\epsilon T$, GDP-NCB maintains strong performance even in the high-privacy regime.
\end{remark}
\begin{remark}[Comparison with lower bound]
\citet{azize2022privacy} shows that the minimax average regret of stochastic bandits with $\epsilon$-global DP is $\ARg_T = \Omega \Big( \max \big\lbrace \sqrt{\frac{k}{T}}, \frac{k}{\epsilon\, T} \big\rbrace\Big)$. From Remark \ref{rem:comp-regret}, this lower bound applies to Nash regret $\NRg_T$ also. Therefore, our upper bound on Nash regret in Theorem~\ref{theorem:improvedNashRegret}
is order optimal up to polylogarithmic factors. 
\end{remark}

The next theorem states the privacy guarantee.

\begin{restatable}{repthm}{TheoremGlobalDP}
\label{theorem:epsilonGlobalDP}
(Privacy guarantee)
GDP-NCB satisfies $\epsilon$-global differential privacy for any $\epsilon >0$.
\end{restatable}

\begin{proof}[Proof Sketch]
The proof proceeds via a case analysis, a standard approach in $\epsilon$-differential privacy (DP) proofs, where we consider the perturbation of a single reward and analyze the effect based on the phase in which the perturbed reward appears.
In the first case, we assume the perturbed reward belongs to Phase II. Here, we leverage the fact that rewards are non-overlapping across episodes, which ensures that the Laplace noise added in Line~\eqref{step:laplace} of Algorithm~\ref{algo:gdp-ncb} maintains differential privacy within each episode. In the second case, we consider a reward perturbed in Phase I. While Phase I itself involves no privacy leakage (as no private outputs are released during this phase), the same reward may influence multiple episodes in Phase II. This could, in principle, lead to a composition of privacy loss. However, we exploit the doubling schedule used in the algorithm, which bounds the number of episodes by $O(\log T)$. Consequently, the overall privacy cost remains controlled and does not become prohibitive. A detailed proof is provided in Appendix \ref{sec:appendix_privacy}. 
\end{proof}

\section{LDP-NCB: Nash Confidence Bound in Local DP}
Local differential privacy (LDP) presents a compelling alternative for decentralized or federated environments where no trusted curator exists. In the LDP model, each user independently perturbs their reward using the Laplace mechanism, i.e., by adding noise drawn from $\mathrm{Lap}(1/\epsilon)$ before reporting, thereby ensuring $\epsilon$-DP at the individual level for rewards in the range $[0,1]$. Although LDP typically incurs higher regret than GDP due to increased noise, it is often more suitable for real-world scenarios involving privacy-sensitive or distributed data collection. Building on the work of \citet{ren2020localdifferentialprivacy}, we propose LDP-NCB, a locally private variant of the NCB algorithm. 
Analogous to our global model result, we also provide formal guarantees for Nash Regret in the LDP setting, as stated in the next theorem.




\makeatletter
    \renewcommand{\ALG@name}{Algorithm}
\makeatother

\begin{algorithm}[t]
    \caption{LDP-NCB }
    \label{algo:ldp-ncb}
    \textbf{Input:} Number of arms $k$, horizon $T$ and privacy budget $\epsilon$\\
    \vspace{-10pt}
    \begin{algorithmic}[1]
	\STATE \textbf{Initialization:} $\widetilde{\mu}_i = 0$, $N_{i} =0$ $\ \forall \ i \in [k]$;  $\ t=1$, $c=3$, $\alpha=3.1$  
	\label{step:budget_ldp}
		\\ \textbf{Phase {\rm I} - Uniform Exploration}
        \WHILE{$\widetilde{\mu}_{i}\leq\frac{1}{\epsilon}\sqrt{\frac{8\alpha\log T}{n_i}}$\\ or ${n_i \ \widetilde{\mu}_i} \leq 1600\left(c^2\log T+\frac{(\log T)^2}{\left(\widetilde{\mu}_{i}-\frac{1}{\epsilon}\sqrt{\frac{8\alpha\log T}{n_i}}\right)\epsilon^2}\right)+\frac{\sqrt{8n_i\alpha\log T}} {\epsilon} \; \forall \;i\in[k]$ and $t \leq T$ } \label{step:PhaseOneAlgTwo_ldp}
		\STATE Pull arm $I_t \sim \Unif([k])$
        \STATE Observe private reward $\widetilde X_t=X_t+\Lap(1/\epsilon)$ 
		  \STATE Update $N_{I_t} \leftarrow N_{I_t} + 1$, $\widetilde{\mu}_{I_t} \leftarrow \frac{N_{I_t}-1}{N_{I_t}}\widetilde{\mu}_{I_t} + \frac{\widetilde X_t}{N_{I_t}}$
            \STATE Update $t\gets t + 1$. 
        \ENDWHILE
		\\ \textbf{Phase {\rm II} - Private Adaptive Exploration}
	\WHILE{$t\leq T$}             
            \STATE Pull arm $I_t = \operatorname{argmax}_{i \in [k]} \NCB_{\LDP}(i, t)$ 
        \STATE Observe private reward $\widetilde X_t=X_t+\Lap(1/\epsilon)$ 
		  \STATE Update $N_{I_t} \leftarrow N_{I_t} + 1$, $\widetilde{\mu}_{I_t} \leftarrow \frac{N_{I_t}-1}{N_{I_t}}\widetilde{\mu}_{I_t} + \frac{\widetilde X_t}{N_{I_t}}$
          \STATE Clip private mean $\widetilde{\mu}_{I_t}=\min\{1, \max\{0, \widetilde{\mu}_{I_t}\}\}$
            \STATE Update $t\gets t + 1$. 
        \ENDWHILE
    \end{algorithmic}
\end{algorithm}

\begin{repthm}[Nash regret in local model]
\label{theorem:improvedNashRegretLDP}
Fix time horizon $T \in \mathbb{N}$ and privacy budget $\epsilon > 0$. Then, for the $k$-armed bandit problem, LDP-NCB enjoys the Nash regret
\[
\NRg_T \;=\;
O\Bigg(\sqrt{\frac{k\log T}{\,T\,}} \;+\;\underbrace{\frac{\sqrt{k}(\log T)^2}{\epsilon\sqrt{T}}}_{\text{Privacy Cost}}\Bigg).
\]
\end{repthm}

\begin{remark}[Privacy cost]
    In the local-DP setting, the privacy penalty is significantly more severe due to the multiplicative dependence on $1/\epsilon$. Unlike the global-DP setting—where the impact of privacy can be negligible for moderately high values of $\epsilon$—LDP does not admit a regime where performance is unaffected by privacy constraints. This is expected, as LDP imposes a stricter privacy requirement, with the algorithm observing only locally privatized data. As a result, the regret scales as $\widetilde{O}\left(\frac{1}{\epsilon}\sqrt{\frac{k}{T}}\right)$, deteriorating rapidly as stronger privacy (i.e., smaller $\epsilon$) is enforced.
\end{remark}

\begin{remark}[Lower bound comparison]
Our bound is order-optimal up to logarithmic factors. This follows from Remark~\ref{rem:comp-regret} and the $\Omega\left(\frac{1}{\epsilon}\sqrt{\frac{k}{T}} \right)$ minimax lower bound on average regret for $\epsilon$-LDP algorithms in high-privacy (sufficiently small $\epsilon$) regime \citep{basu2019differential}.
\end{remark}

\section{Analysis and Proof Sketch}
We now provide a proof sketch for Theorems~\ref{theorem:improvedNashRegret} and~\ref{theorem:improvedNashRegretLDP}. The sketch is presented in a unified manner, as the key ideas underlying both proofs are largely similar, differing only in a few crucial aspects. A complete and rigorous treatment is deferred to Appendix \ref{appendix:key-lemmas}.

Denote by $\widehat{\mu}_{i,s}$ the empirical mean reward of arm $i$ after its first $s$ pulls. For the global setup, define 
\begin{equation}
\label{eq:S-global}
S\coloneqq \frac{c^{2}\,\log T}{\mu^{*}} + \frac{(\log T)^{2}}{\mu^{*}\,\epsilon}~, 
\end{equation}
and for the local setup, define
\begin{equation}
\label{eq:S-local}
S=\frac{c^2\log T}{\mu^*}+\left(\frac{\log T}{\mu^*\epsilon}\right)^2~.
\end{equation}
Next, we define a \emph{good event} $\cE$, which holds with high probability. All subsequent arguments are conditioned on the event $\cE$ holding. Let $\cE$ be the intersection of four sub-events: $\cE_1$, $\cE_2$, $\cE_3$, and $\cE_4$, defined as follows:

\begin{itemize}
  \item[$\cE_{1}$:] During the initial $r$ rounds of uniform sampling, for every arm $i\in[k]$ and any $r\ge 512k\, S$, the number of pulls of arm $i$ lies between $r/(2k)$ and $3r/(2k)$.
  
  \item[$\cE_{2}$:] For each arm $i\in[k]$ with mean $\mu_{i}>\mu^{*}/256$, and for all sample sizes $s$ satisfying $256\,S\le s\le T$, the empirical estimate obeys $\bigl|\widehat{\mu}_{i,s}-\mu_{i}\bigr| \;\le\; c\sqrt{\frac{\mu_{i}\,\log T}{s}}$.
  
  \item[$\cE_{3}$:] For each arm $j\in[k]$ with $\mu_{j}\le \mu^{*}/256$, and for all $s$ in the range $256\,S\le s\le T$, we have $\widehat{\mu}_{j,s}<\frac{\mu^{*}}{128}$~.

  \item[$\cE_{4}$:] For each arm $i\in[k]$, the difference between private and non-private empirical mean $|\widetilde{\mu}_i-\widehat{\mu}_i|$ is $\delta$, where  $\delta  = \frac{\alpha (\log T)^2}{\epsilon\,n_{i}}$ for GDP and $\delta = \frac{1}{\epsilon}\sqrt\frac{8\alpha\log T}{n_i}$ for LDP.
\end{itemize}
We now state a lemma that provides a high-probability guarantee for the event $\cE$ with its proof deferred to Appendix \ref{appendix:proof-of-E}.

\begin{lemma}[Good event]
\label{lemma:modifiedgoodeventpr}
For any $T\in \mathbb{N}$, $\Pr\left[ \cE \right] \ge 1 - \frac{6}{T}$ for both GDP-NCB and LDP-NCB algorithms. 
\end{lemma}
Next, we state the following informal lemma, which helps us bound the total length of Phase I in our algorithms. Rigorous statements are given in lemmas \ref{tau:lower}-\ref{tau:anytime} (for GDP) and \ref{tau:lower_ldp}-\ref{tau:anytime_ldp} (for LDP) in the Appendix \ref{appendix:key-lemmas}.

\begin{lemma*}[Informal]  Under event $\cE$, Phase I in both algorithms runs for $\Theta(kS)$ rounds. 
\end{lemma*}

We now turn to three crucial properties of our algorithms that hold under the “good” event $\cE$.
\begin{lemma*}[Informal] 
    Throughout Phase II, the NCB of the optimal arm \(i^*\) never falls below its true mean \(\mu^*\).
\end{lemma*}

\begin{lemma*}[Informal]
    Any arm \(j\) with mean \(\mu_j \le \mu^*/256\) is never selected during Phase II.
\end{lemma*}

\begin{lemma*}[Informal]
    If an arm \(i\) is pulled in Phase II, then \(\mu_i\) must be close to \(\mu^*\), so its contribution to the Nash regret is negligible.
\end{lemma*}

More rigorous treatment for the above informal lemmata is provided in lemmas \ref{lem:emp:anytime}-\ref{lem:suboptimal_arms:anytime} (for GDP) and \ref{lem:emp:anytimeldp}-\ref{lem:suboptimal_arms:anytime_ldp} (for LDP) in the Appendix \ref{appendix:key-lemmas}. The above three lemmata help us derive a bound on the Nash Social Welfare (NSW) for our algorithms, which is captured below informally.
\begin{lemma*}[Informal]
    For $\mu^* = \widetilde{\Omega}(\frac{1}{\sqrt T}+\frac{1}{\epsilon T})$, the NSW for the GDP-NCB upto $T$ rounds satisfies $(\prod_{t=1}^{T}\E[\mu_{I_t}])^\frac{1}{T} \approx \mu^{*}-\sqrt{\frac{\mu^*}{T}}-\frac{1}{\epsilon T}$.
\end{lemma*}
This bound is stated in detail in Lemma \ref{lem:modified_ncb} (for GDP). Finally, we use this lemma to derive the Nash regret bound in Theorem \ref{theorem:improvedNashRegret}). Observe that the conclusion in Theorem~\ref{theorem:improvedNashRegret} holds trivially when $\mu^* = \widetilde{O}(\frac{1}{\sqrt T}+\frac{1}{\epsilon T})$ 
Thus, we need to prove the bound 
for $\mu^* = \widetilde{\Omega}(\frac{1}{\sqrt T}+\frac{1}{\epsilon T})$, which follows from the NSW identity stated above. Detailed proof of Theorem \ref{theorem:improvedNashRegret} is given in Appendix \ref{appendix:key-lemmas}.

A similar analysis for NSW works for the LDP-NCB algorithm (Lemma \ref{lem:modified_ncb_ldp}) and its Nash regret (Theorem \ref{theorem:improvedNashRegretLDP}). 
\begin{remark}[Clipping]
The concentration bounds remain valid even after clipping of the private mean to the interval $[0,1]$. Specifically, if $\tilde{\mu}$ lies within an interval $[L, U]$, then the clipped value $\tilde{\mu}_{\text{clip}} = \min\{1, \max\{0, \tilde{\mu}\}\}$ continues to satisfy $L \leq \tilde{\mu}_{\text{clip}} \leq U$. If $L \leq 0$, clipping only increases $\tilde{\mu}$ (if needed) up to $0$, thereby preserving the lower bound. If $L > 0$, then $\tilde{\mu} \geq L > 0$, so clipping does not alter the value. Similarly, the upper bound is preserved since $\tilde{\mu}_{\text{clip}} \leq \tilde{\mu} \leq U$. Hence, all upper and lower bounds used in regret and confidence interval analyses remain valid under clipping.
\end{remark}

\section{Anytime Private and Fair Algorithm}
\label{section:final-anytime}
We now present the \emph{anytime} version of our DP-NCB framework, described in Algorithm~\ref{algo:ncb:anytime}. An \emph{anytime algorithm} is designed to operate without prior knowledge of the total time horizon. Following the approach of \citet{barman2023fairness}, our algorithm employs the \emph{doubling trick}. It maintains a current window length \( W \in \mathbb{Z}_+ \) as a guess for the horizon. During each epoch of \( W \) rounds, the algorithm proceeds as follows:
\begin{enumerate}
  \item With probability \( 1 / W^2 \), it performs uniform exploration; 
  \item With probability \( 1 - 1 / W^2 \), it invokes either Algorithm~\ref{algo:gdp-ncb} or Algorithm~\ref{algo:ldp-ncb}, depending on the privacy setting, over the remaining rounds of the window.
\end{enumerate}
At the end of each epoch, the window length is doubled, and the process is repeated until a termination signal is received.

\begin{algorithm}[ht!]
    \caption{Anytime Algorithm for Nash Regret}
    \label{algo:ncb:anytime}
    \noindent
    \textbf{Input:} Number of arms $k$, privacy budget $\epsilon$

    \begin{algorithmic}[1]
        \STATE Initialize $ W = 1$.
        \WHILE{the MAB process continues}
        \STATE With probability $\frac{1}{W^2}$ set {\rm flag} = \textsc{Uniform} , otherwise, with probability $\left(1 - \frac{1}{W^2}\right)$, set {\rm flag }= \textsc{DP-NCB}
        \IF {{\rm flag} = \textsc{Uniform} }
        \FOR{$t=1$ to $W$}
        \STATE Select $I_t$ uniformly at random from $[k]$. Pull arm $I_t$ and observe reward $X_t$. \label{step:calluniform}
        \ENDFOR
        \ELSIF {{\rm flag }= \textsc{DP-NCB}}
        \STATE  Execute DP-NCB($k$, $W$, $\epsilon$). \label{step:callsubroutine}
        \ENDIF
        \STATE Update $W \gets 2 \times W$. 
        \ENDWHILE
    \end{algorithmic}
\end{algorithm}
Note that in Algorithm~\ref{algo:ncb:anytime}, DP-NCB can refer to either GDP-NCB or LDP-NCB, depending on the required privacy setting. Algorithm~\ref{algo:ncb:anytime} leads to Theorem~\ref{thm:anytime-nucb} and Theorem~\ref{thm:anytime-nucb_ldp}, which share similar proof techniques. A proof sketch is provided below, with complete proofs deferred to the Appendix \ref{appendix:key-lemmas}.

\begin{repthm}
\label{thm:anytime-nucb}
For any $\epsilon >0$ and a sufficiently large $T$, Algorithm \ref{algo:ncb:anytime}, instantiated with GDP-NCB, satisfies $\epsilon$-global DP and guarantees Nash regret
\begin{align*}
\NRg_T = O\bigg(\sqrt{\frac{k\log T}{T}}\log T+\frac{k\ (\log T)^3}{\epsilon T}\bigg)~.
\end{align*}
\end{repthm}
\begin{repthm}
\label{thm:anytime-nucb_ldp}
For any $\epsilon >0$ and a sufficiently large $T$, Algorithm \ref{algo:ncb:anytime}, instantiated with LDP-NCB, satisfies $\epsilon$-local DP and guarantees Nash regret
\begin{align*}
\NRg_T = O\bigg(\sqrt{\frac{k\log T}{T}}\log T + \frac{\sqrt k(\log T)^2}{\epsilon\sqrt{T}}\bigg)~.
\end{align*}
\end{repthm}

\begin{proof}[Proof Sketch]
The proof uses the doubling trick to create an anytime version of the DP-NCB algorithm. In each epoch of guessed length $W$, the algorithm either performs uniform exploration with probability $\frac{1}{W^2}$ or runs the base DP-NCB algorithm (GDP-NCB or LDP-NCB) on the epoch. Since the probability of choosing uniform exploration is low and decays with $W$, most epochs run DP-NCB, ensuring that the Nash regret remains close to that of the fixed-horizon case. The overall regret is then obtained by summing the regrets (calculated from Theorem \ref{theorem:improvedNashRegret} and Theorem \ref{theorem:improvedNashRegretLDP} for each epoch) over all epochs, using the fact that the number of epochs is $O(\log T)$. This introduces only an $O(\log T)$ multiplicative overhead in regret, yielding the final bound stated in Theorem \ref{thm:anytime-nucb}. Theorem \ref{thm:anytime-nucb_ldp} can be proved along similar lines. 
\end{proof}


\section{Experiments}

In this section, we present numerical results comparing our methods with NCB ~\citep{barman2023fairness} and AdaP-UCB~\citep{azize2022privacy} algorithms. All experiments were conducted on an Apple M4 CPU with 16GB RAM. All experiments report the average reward over 50 runs to estimate \( \mathbb{E}[\mu_{I_T}] \), using a privacy parameter of \( \epsilon = 0.2 \) wherever applicable.

First, we demonstrate how the Nash regret \( \NRg_T \) of AdaP-UCB can grow rapidly for certain bandit instances. Specifically, we consider a setting with \( k = 2 \) Bernoulli arms having means \( \mu_1 = (2e)^{-T} \) and \( \mu_2 = 1 \), for \( T \in (0, 1000) \), following \citet{barman2023fairness}. In this case, AdaP-UCB exhibits significantly inflated Nash regret, while our private algorithm maintains controlled regret, as shown in Figure (a).



Next, we compare the Nash regret of Non-Private NCB, AdaP-UCB, LDP-UCB~\citep{ren2020localdifferentialprivacy}, GDP-NCB, and LDP-NCB on a problem instance with \( k = 50 \) Bernoulli arms, where the means are sampled uniformly from the interval \( (0.005, 1) \). We use parameters \( c = 3 \) and \( \alpha = 3.1 \), following the experimental setup of \citet{krishna2025p}. As shown in Figure (b), GDP-NCB successfully reduces Nash regret compared to AdaP-UCB for larger values of \( T \). Similarly, in the LDP setting, LDP-NCB achieves significantly lower regret compared to LDP-UCB.

Further, we evaluate Nash Regret for GDP-NCB and LDP-NCB for varying \(\epsilon\) (Figure (c) and (d)). As predicted by our analysis, regret increases as \(\epsilon\) decreases, with non-private NCB consistently achieving the lowest Nash Regret.  \\
We then compare Non-Private NCB, GDP-NCB, and LDP-NCB on \(k=50\) Bernoulli arms with similar setup as the previous experiment. Figure (e) shows that GDP-NCB's regret decays more slowly than Non-Private NCB (due to Laplace noise), and LDP-NCB decays the slowest, following an \(O(1/\sqrt{T})\) rather than \(O(1/T)\) rate.

Finally, we evaluate performance with mixed arms, where the 50 arms have means sampled uniformly from \((0.005, 1)\), but differ in reward distributions: arms with $\mu_i \geq 0.75$ are Bernoulli; arms with $\mu_i \in [0.5, 0.75)$ follow $\mathrm{Beta}(4, 1)$; arms with $\mu_i \in [0.25, 0.5)$ follow a two-point distribution $\{0.4, 1\}$ and arms with $\mu_i < 0.25$ follow $\Unif(0, 1)$. As shown in Figure (f), Nash regret trends are similar to Figure (e) as $T$ increases, highlighting the algorithms' robustness to heterogeneous reward distributions.

\begin{figure*}[t]\centering
		\begin{subfigure}[b]{0.32\textwidth}
			\includegraphics[scale=0.25]{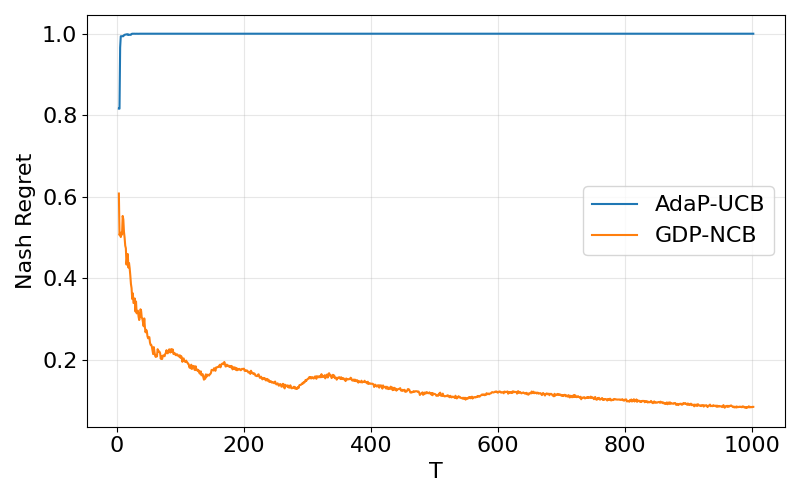}
			\caption{AdaP-UCB vs GDP-NCB}
		\end{subfigure} 
		\begin{subfigure}[b]{0.32\textwidth}
			\includegraphics[scale=0.25]{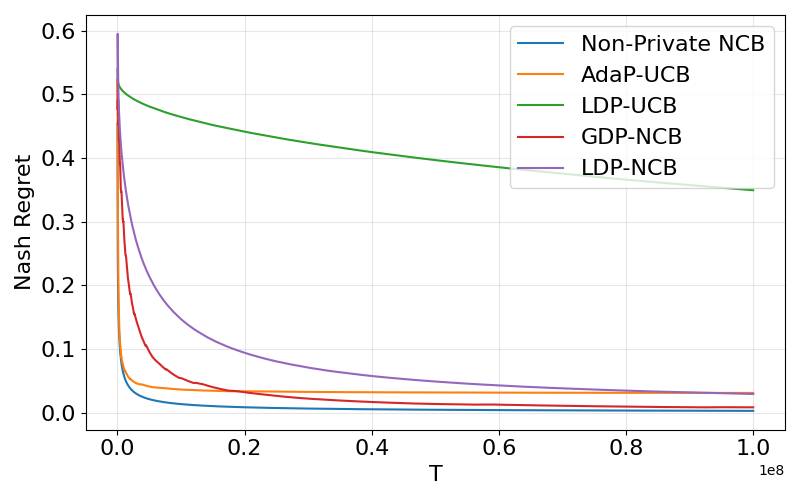}
			\caption{Different Algorithms, $\epsilon=0.2$}
		\end{subfigure}\ \ 
		\begin{subfigure}[b]{0.32\textwidth}
			\includegraphics[scale=0.25]{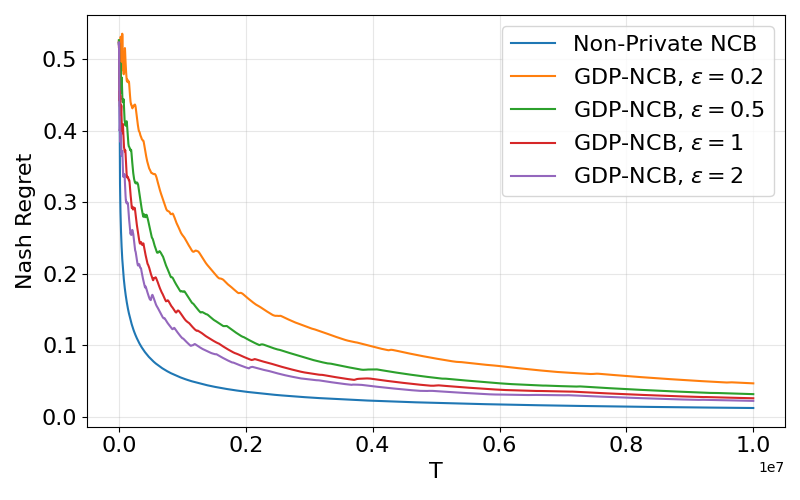}
			\caption{GDP-NCB, vary $\epsilon$}
		\end{subfigure}\\
		\begin{subfigure}[b]{0.32\textwidth}
			\includegraphics[scale=0.25]{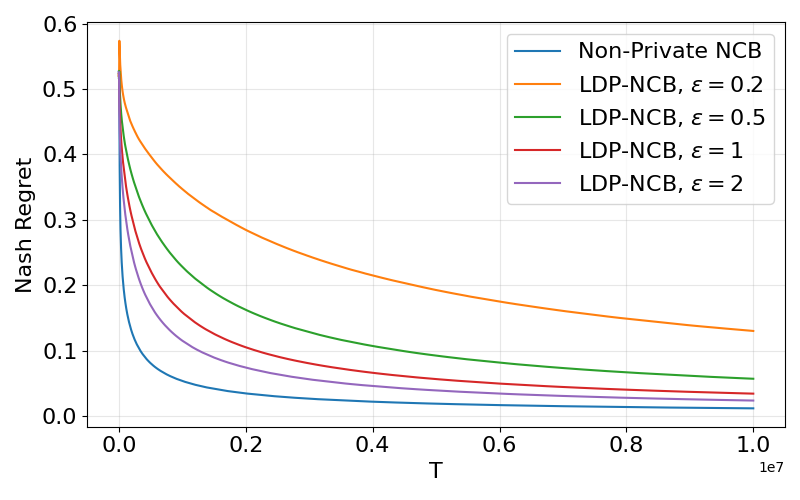}
			\caption{LDP-NCB, vary $\epsilon$}
		\end{subfigure}\ \ 
		\begin{subfigure}[b]{0.32\textwidth}
			\includegraphics[scale=0.25]{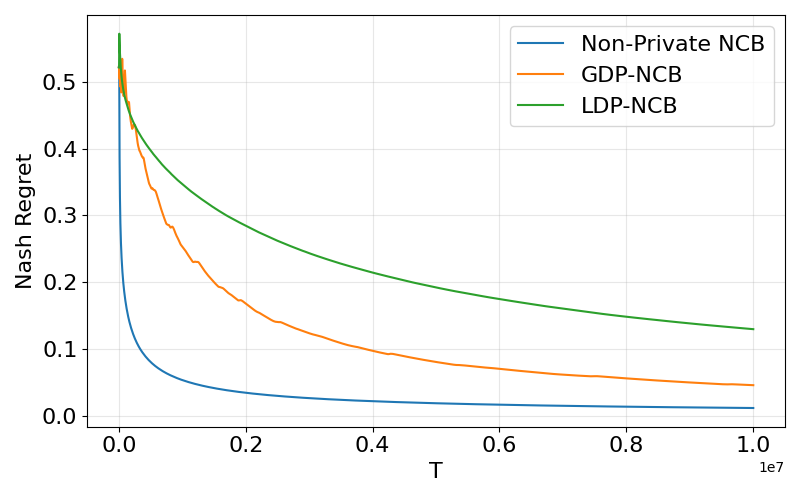}
			\caption{Bernoulli arms, $\epsilon = 0.2$}
		\end{subfigure}
		\begin{subfigure}[b]{0.32\textwidth}
			\includegraphics[scale=0.25]{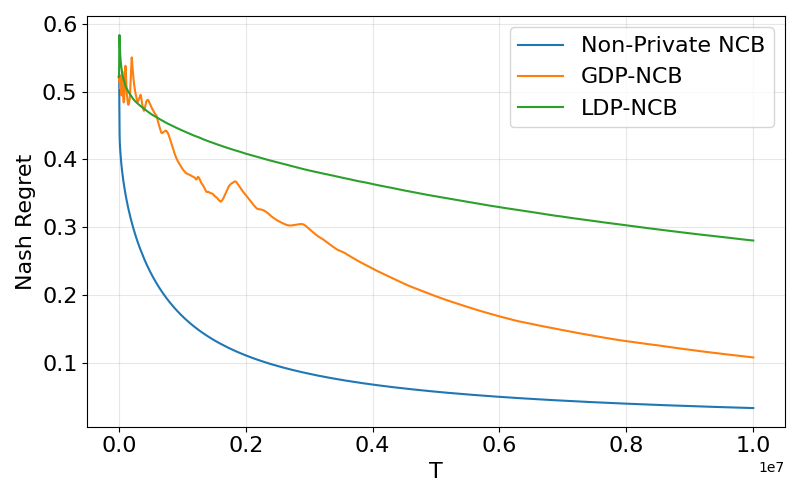}
			\caption{Mixed types of arms, $\epsilon = 0.2$.}
		\end{subfigure}
    \caption{Numerical results for GDP-NCB and LDP-NCB. (a) shows that GDP-NCB significantly outperforms AdaP-UCB by avoiding blown-up Nash regret in extreme instances. (b) illustrates the comparison between our algorithms and existing algorithms. (c) and (d) show that as $\epsilon$ decreases, Nash regret for DP-NCB increases, aligning with theoretical predictions. (e) shows that GDP-NCB’s regret decays slower than NCB, and LDP-NCB follows the slowest decay, under Bernoulli arms. (f) shows that GDP-NCB remains robust and maintains Nash regret trends even under heterogeneous reward distributions.}
\end{figure*}

\section{Conclusion}

This work bridges the gap between privacy-preserving algorithms and welfarist concerns by introducing a unified framework—DP-NCB. Our algorithms provide a versatile solution to the MAB problem with applications in sensitive domains such as personalized recommendations and healthcare. Future directions include extending the framework to other reinforcement learning settings and exploring alternative fairness metrics tailored to specific domains.


\appendix

\newpage
\section{\LARGE Appendix}
In Appendix \ref{app:rel_w}, we discuss some of the existing literature related to multi-arm bandits and differential privacy. We
provide proofs for the global differential privacy of Algorithm \ref{algo:gdp-ncb} in Appendix \ref{sec:appendix_privacy} step by step. Finally, Appendix \ref{appendix:key-lemmas} provides formal statements and proofs for all the informal lemmas and theorems mentioned during regret analysis.
\section{Related Work}
\label{app:rel_w}
\subsection{Differential Privacy in Bandits}
Differential Privacy is a framework which is used to model the privacy of datasets in the sense that one cannot discern the identity of individual data subjects from the statistics released. This is primarily achieved through noise injection. The main trade-off is between privacy and utility. That is, we want to ensure privacy while preserving the utility of the released statistics. Much of differential privacy research has focused on static databases. While this is useful, the complete picture can be formed only when add the time dimension to our model. 

\citet{dwork2009complexity} and \citet{chan2011private} dealt with the same problem. It was essentially a study of the problem where statistics are continually released.

\paragraph{Foundations of Private Continual Release.}  
The work of \citet{chan2011private} laid critical groundwork for privatizing streaming data through their \textit{binary tree-based aggregation scheme}. By hierarchically perturbing partial sums with Laplace noise, they achieved polylogarithmic error ($O(\log^{1.5} T/\epsilon)$) for continual counting, while guaranteeing $\epsilon$-DP. This mechanism enabled applications like private top-$k$ recommendations and range queries. Their framework also introduced \textit{pan-privacy}, ensuring robustness against intrusions into internal algorithm states. These techniques became foundational for later MAB algorithms requiring private reward aggregation.

\paragraph{Early Private Bandit Algorithms.}  
Building on \citet{chan2011private}, \citet{mishra2015nearly} adapted Upper Confidence Bound (UCB) and Thompson Sampling (TS) to the DP setting. Their UCB variant privatized per-arm reward sums using tree-based aggregation, splitting the privacy budget ($\epsilon/k$) across $k$ arms and adjusting confidence intervals to account for noise. While achieving $O\left(\sum \frac{\log^2 T}{\epsilon \Delta_a}\right)$ regret, this introduced poly-log multiplicative factors compared to non-private UCB. Similarly, their TS adaptation required modified exploration phases, but regret bounds remained suboptimal. These works demonstrated the feasibility of private MAB but highlighted the need for tighter regret-privacy trade-offs.

\paragraph{Near-Optimal Private Regret Bounds.}  
A breakthrough came with \citet{tossou2016algorithms}, who introduced novel mechanisms to minimize privacy-induced regret. Their \textit{DP-UCB-Int} algorithm employed interval-based updates, lazily refreshing arm means every $O(1/\epsilon)$ pulls. By decoupling noise magnitude from $\epsilon$ via adaptive composition, they achieved near-optimal $O(\log T + \epsilon^{-1})$ regret—adding only an \textit{additive constant} to non-private bounds. This contrasted sharply with prior multiplicative penalties (e.g., $O(\epsilon^{-2}\log^2 T)$ in \citet{mishra2015nearly}). Their hybrid mechanism (logarithmic + tree-based noise) also improved earlier DP-UCB variants, reducing regret to $O(\epsilon^{-1}\log \log T)$. These innovations demonstrated that privacy need not preclude near-optimal performance.

\paragraph{Refining Privacy-Partial Information Tradeoffs.}  
Recent work by \citet{azize2022privacy} significantly advances the theoretical understanding of DP bandits by unifying privacy constraints with the inherent partial-information structure of bandit environments. The authors derive the first \textit{problem-dependent regret lower bounds} for general reward distributions (Theorem 3), explicitly linking the hardness of private learning to the TV- and KL-indistinguishability gaps of the environment. This contrasts with prior bounds limited to Bernoulli rewards (e.g., \cite{shariff2018differentially}) and resolves an open question on KL-dependent regret under DP \citep{tenenbaum2021differentially}. Their analysis reveals two distinct regimes: in the \textit{high-privacy regime} ($\epsilon \ll \Delta_a$), regret depends on a coupled effect of $\epsilon$ and the environment’s distinguishability; in the \textit{low-privacy regime} ($\epsilon \gg \Delta_a$), DP adds no asymptotic regret over non-private algorithms—generalizing insights from \citet{tossou2016algorithms} to structured settings.  

Algorithmically, the authors propose a framework to convert standard index-based algorithms (e.g., UCB, KL-UCB) into $\epsilon$-global DP variants via three innovations: (1) Laplace noise calibrated to arm-specific sensitivity, (2) adaptive episodes with doubling to limit mean recomputations, and (3) forgetting rewards outside the last active episode. Instantiated as \textit{AdaP-KLUCB}, their algorithm achieves problem-dependent regret matching the lower bound (Theorem 8), improving upon prior additive gaps in \cite{hu2021optimal} and multiplicative factors in \cite{mishra2015nearly}. 

Technically, the work extends the \textit{Karwa-Vadhan lemma} to sequential settings (Lemma 2) and introduces a novel information-processing lemma (Theorem 10) to bound policy divergences under DP—tools enabling tighter analysis of structured bandits. These advances close gaps in prior linear bandit lower bounds \citep{basu2019differential} and establish a blueprint for privatizing optimistic algorithms beyond stochastic bandits.

A common thread in these works is that they use the DP-mechanism to compute the private means from the non-private means which the player then uses to compute its actions. This means often involves using modified concentrations inequalities to account for the noise injection done during computation of the private mean. Each arm corresponds to a stream of statistics and there are $K$ such streams on which we can independently apply the DP mechanism.

\section{Privacy Analysis of Algorithm~\ref{algo:gdp-ncb}}\label{sec:appendix_privacy}

Here, we show that any bandit algorithm built using Algorithm~\ref{algo:gdp-ncb}'s framework ensures $\epsilon$-global differential privacy (DP) over both Phases I and II. We first prove that the private indices calculated in Algorithm~\ref{algo:gdp-ncb} satisfy $\epsilon$-global DP. The final result then follows because DP maintains privacy under post-processing (as stated in Lemma 1 in \citet{azize2022privacy}), if rewards are changed in Phase II. For Phase I, we conduct rigorous analysis to show $\epsilon$-DP is satisfied. 

\begin{lemma}(Privacy of Episode‐Wise Means in Phase II)
\label{lem:privEpisode}
Denote by $T_1$ the total number of rounds in Phase~I, and let
\[
r_{T_1+1}, \dots, r_T \in [0,1]
\]
be the rewards gathered during Phase~II. We then partition the Phase~II horizon into a collection of non\-overlapping episodes:
\[
  E_1 = \{t_{T_1+1},\dots,t_1-1\},\;
  E_2 = \{t_{1},\dots,t_2-1\},\;
  E_3 = \{t_2,\dots,t_3-1\},\;\dots
\]
where episode \(\ell\) has length \(n_\ell = |E_\ell|\) and $T_1+1<t_1<t_2<\cdots<t_\ell<T$.  Fix constants \(N_{1,a}\) and \(\hat\mu_a\) from Phase I.  Define the private mean for episode \(\ell\) of arm \(a\) as
\begin{align}
  f^\epsilon_\ell(r)
  \;=\;
  \underbrace{
    \frac{N_{1,a}}{N_{1,a}+n_\ell}\,\hat\mu_a
    \;+\;
    \frac1{N_{1,a}+n_\ell}
    \sum_{t\in E_\ell} r_t
  }_{\displaystyle\text{unperturbed mean}}
  \;+\;
  \Lap\left(\frac{\log T}{\epsilon (N_{1,I_{\ell}}+n_\ell)}\right).
\end{align}
Then the vector \(\bigl(f^\epsilon_1,\dots,f^\epsilon_L\bigr)\) satisfies \(\epsilon\)-differential privacy.
\end{lemma}

\begin{proof}
Consider two reward streams \(\mathbf r\) and \(\mathbf r'\) that differ in exactly one entry, say at time \(j\). If \(j\notin E_\ell\), then \(f^\epsilon_\ell(\mathbf r)=f^\epsilon_\ell(\mathbf r')\) and we are done. If \(j\in E_\ell\), then the only term affected in the clean mean:
\begin{align} 
  \frac{N_{1,a}}{N_{1,a}+n_\ell}\,\hat\mu_a
  + \frac1{N_{1,a}+n_\ell}\sum_{t\in E_\ell}r_t
\end{align}
is one summand \(r_j\).  Hence its sensitivity is
\begin{align}
  |\Delta|
  \;\le\;
  \frac1{N_{1,a}+n_\ell}.
\end{align}
Adding Laplace noise with scale $\frac{\log T}{(N_{1,a}+n_\ell)\,\epsilon}$ guarantees (by properties of Laplace distribution and triangle inequality) that:
\begin{align}
  \Pr\bigl[f^\epsilon_\ell(\mathbf r)\in S\bigr]
  \;\le\;
  e^{\frac{\epsilon}{\log T}}\,
  \Pr\bigl[f^\epsilon_\ell(\mathbf r')\in S\bigr]\;\le\;e^{\epsilon}\,
  \Pr\bigl[f^\epsilon_\ell(\mathbf r')\in S\bigr]
\end{align}
for every measurable set \(S\).

Since each episode \(E_1,E_2,\dots\) is disjoint, a single changed reward can affect at most one output \(f^\epsilon_\ell\).  Therefore the entire vector \((f^\epsilon_1,\dots,f^\epsilon_L)\) satisfies \(\epsilon\)-DP by simple composition of disjoint releases.
\end{proof}

\TheoremGlobalDP*

\begin{proof}
We need to consider the following two cases:\\\\
\textit{1. Changing a reward in Phase II: }A change in any one Phase II reward only affects the private mean of the single episode in which it lives. By Theorem 6 from \citet{azize2022privacy}, each episode’s Laplace release is $\epsilon$‐DP (w.r.t. that one reward), and because episodes are disjoint in Phase II, the collection of all Phase II releases is $\epsilon$‐DP outright. Further, by the argument in Lemma \ref{lem:privEpisode}, a constant addition of means from Phase I (since no reward is changing in Phase I) still keeps individual episodic releases $\epsilon$-DP. Thus, the entire algorithm over both Phases I and II is $\epsilon$-DP.\\\\
\textit{2. Changing a reward in Phase I: }Now, each private-mean release also depends (linearly) on the fixed Phase I empirical mean $\widehat{\mu_i}$, which in turn depends on all $N_{1,a}$ rewards from uniform exploration. A change to one of those Phase I rewards will shift $\widehat{\mu_i}$ by at most $\frac{1}{N_{1,a}}$, and therefore shifts every private mean by at most

\begin{align}
    \Delta_I = \frac{1/N_{1,a}}{N_{1,a}+n_\ell}N_{1,a}=\frac{1}{N_{1,a}+n_\ell} 
\end{align}

Since we have set the scale of the Laplace noise as $b = \frac{\log T}{\epsilon(N_{1,a}+n_\ell)}$, for the $\ell^{th}$ episode, the privacy release is
\begin{align*}
    \epsilon_{\ell} = \frac{\Delta_I}{b} = \frac{\epsilon(N_{1,a}+n_\ell)}{(N_{1,a}+n_\ell)\log T}=\frac{\epsilon}{\log T}
\end{align*} 
So the total privacy cost over all $\ell$ episodes is
\begin{align*}
    \epsilon_{net} = \sum_{i=1}^{\ell}\epsilon_\ell=\frac{\epsilon\;\ell}{\log T}
\end{align*}

To find a bound on $\ell$, assume that the arm with the maximum $N_{1,a}$ is pulled for all $\ell$ episodes between $T_1$ and $T$. Since the number of pulls double every episode, we have
\begin{align*}
    2N_{1,a}+4N_{1,a}+\cdots+2^{\ell}N_{1,a} =T-T_1
    \implies (2^{\ell+1}-2)N_{1,a}=T-T_1
\end{align*}
Since $\ell$ has to be an integer, we have
\begin{align*}
    \ell \le \log \left(1+\frac{T-T_1}{2N_{1,a}}\right) \leq\log \left(1+\frac{T-T_1}{2}\right) 
\end{align*}

The last inequality holds as $N_{1,a}\geq1$ for the arm whose reward is changed. This inequality implies that the net privacy cost is 
\begin{align*}
    \epsilon_{net} \le \log\left(1+\frac{T-T_1}{2}\right)\frac{\epsilon}{\log T}\le\epsilon
\end{align*}

Hence, the total privacy cost is less than the privacy budget $\epsilon$, and our algorithm is $\epsilon$-DP, which completes the proof.

\end{proof}

\section{Missing Proofs}
\label{appendix:key-lemmas}

\subsection{Proof of Lemma \ref{lemma:modifiedgoodeventpr}}
\label{appendix:proof-of-E}

In this part, we show that the \emph{good} event $\cE$ occurs with probability at least $1 - \frac{6}{T}$. We will separately bound the probabilities of the complements of $\cE_1$, $\cE_2$, $\cE_3$ and $\cE_4$, and then apply the union bound. We begin by recalling standard concentration results that will be instrumental in our analysis. In particular, we define Lemmas \ref{lem:chernoff}, \ref{lemma:hoeffding}, \ref{lemma:hoeff_small_mean} which are mentioned in the book by \citet{dubhashi2009concentration}.\\

\begin{lemma}[Chernoff Bound]\label{lem:chernoff}
Let $Z_1, \ldots, Z_n$ be independent $\{0,1\}$-valued random variables. Define $S = \sum_{r=1}^n Z_r$ and denote $\nu = \E[S]$. Then, for any $\lambda \in [0,1]$,
\[
    \prob\bigl\{S \le (1-\lambda)\,\nu\bigr\} \;\le\; \exp\!\Bigl(-\tfrac{\nu\,\lambda^2}{2}\Bigr).
\]
\end{lemma}
\begin{lemma}[Hoeffding's Inequality]\label{lemma:hoeffding}
Let $Y_1,\dots,Y_n$ be independent random variables taking values in $[0,1]$. Define $\widehat Y = \tfrac1n\sum_{j=1}^n Y_j$ and let $\nu = \E[\widehat Y]$. Then for any $\delta\in[0,1]$,
\[
    \prob\{|\widehat Y - \nu|\ge\delta\,\nu\}
    \;\le\;
    2\,\exp\!\Bigl(-\tfrac{\delta^2}{3}\,n\,\nu\Bigr).
\]
\end{lemma}

\begin{lemma}[Hoeffding Extension for Small Means]\label{lemma:hoeff_small_mean}
Let $Y_1,\dots,Y_n$ be i.i.d.\ in $[0,1]$, and write $\widehat Y = \tfrac1n\sum_{j=1}^n Y_j$. If $\E[\widehat Y]\le\nu_H$, then for any $\delta\in[0,1]$,
\[
    \prob\{\widehat Y \ge (1+\delta)\,\nu_H\}
    \;\le\;
    \exp\!\Bigl(-\tfrac{\delta^2}{3}\,n\,\nu_H\Bigr).
\]
\end{lemma}
\hfill\\

We first bound $\prob\{\cE_1^c\}$. For each arm $i$ and round $t$ during uniform sampling, let the indicator $Z_{i,t}$ be $1$ if arm $i$ is chosen in that round, and $0$ otherwise. Setting $\lambda = \tfrac12$ and noting that $\E\bigl[\sum_{t=1}^r Z_{i,t}\bigr]=r/k\ge512\,\Sample$, Lemma \ref{lem:chernoff} together with a union bound yields the following\footnote{Recall that $\Sample = \tfrac{c^2 \log T}{\mu^*} + \tfrac{(\log T)^2}{\mu^*\epsilon}$ for Algorithm \ref{algo:gdp-ncb}, and $\Sample = \tfrac{c^2 \log T}{\mu^*} + \tfrac{(\log T)^2}{(\mu^*)^2\epsilon^2}$ for Algorithm \ref{algo:ldp-ncb}, with $\mu^*\le1$.}\\

For Algorithm \ref{algo:gdp-ncb}, we obtain:
\begin{align}
    \prob\{\cE_1^c\}
    &\le 2\,\exp\!\Bigl(-\tfrac{512\,c^2\log T}{12\,\mu^*} - \tfrac{512\,(\log T)^2}{12\,\mu^*\epsilon}\Bigr)\,\cdot kT 
    \;\le\; 2\,\exp\!\Bigl(-\tfrac{512\,c^2\log T}{12\,\mu^*}\Bigr)\,kT 
    \;\le\; \tfrac{1}{T}, 
    \label{ineq:probG1c:anytime}
\end{align}
and for Algorithm \ref{algo:ldp-ncb}:
\begin{align}
    \prob\{\cE_1^c\}
    &\le 2\,\exp\!\Bigl(-\tfrac{512\,c^2\log T}{12\,\mu^*} - \tfrac{512\,(\log T)^2}{12\,(\mu^*)^2\epsilon^2}\Bigr)\,\cdot kT 
    \;\le\; 2\,\exp\!\Bigl(-\tfrac{512\,c^2\log T}{12\,\mu^*}\Bigr)\,kT
    \;\le\; \tfrac{1}{T}.
    \label{ineq:probG1c:anytime_ldp}
\end{align}

We now turn to $\prob\{\cE_2^c\}$. Note that whenever $\mu_i > \tfrac{\mu^*}{256}$ and $s \ge 256\,\Sample$, one has $c\sqrt{\tfrac{\log T}{\mu_i s}}<1$. Applying Lemma \ref{lemma:hoeffding} with $n=s$, $\nu = \mu_i$, and $\delta = c\sqrt{\tfrac{\log T}{\mu_i s}}$ yields
\[
    \prob\bigl\{|\widehat\mu_{i,s} - \mu_i|\ge c\sqrt{\tfrac{\mu_i\log T}{s}}\bigr\}
    = \prob\bigl\{|\widehat\mu_{i,s} - \mu_i|\ge \delta\,\mu_i\bigr\}
    \;\le\;
    2\,\exp\!\Bigl(-\tfrac{c^2\log T}{3}\Bigr)
    \;=\;
    \tfrac{2}{T^3}.
\]
A union bound over all arms and time steps then gives
\begin{align}
    \prob\{\cE_2^c\}
    \;\le\;
    kT \,\cdot \tfrac{2}{T^3}
    \;\leq\;
    \tfrac{2}{T}.
    \label{ineq:probG2c:anytime}
\end{align}

Then, we bound $\prob\{\cE_3^c\}$. Fix any arm $j$ with $\mu_j \le \tfrac{\mu^*}{256}$ and any $s\ge256\,\Sample$. Applying Lemma \ref{lemma:hoeff_small_mean} with $\nu_H = \tfrac{\mu^*}{256}$ and $\delta=1$, we get, for Algorithm \ref{algo:gdp-ncb},
\[
    \prob\{\widehat\mu_{j,s} \ge \tfrac{\mu^*}{128}\}
    \;\le\;
    \exp\!\Bigl(-\tfrac{1}{3}\,\tfrac{\mu^*}{256}\,s\Bigr)
    = \exp\!\Bigl(-\tfrac{c^2\log T}{3} - \tfrac{(\log T)^2}{\epsilon}\Bigr)
    \;\le\;
    \exp\!\Bigl(-\tfrac{c^2\log T}{3}\Bigr)
    =
    \tfrac{1}{T^3}.
\]
Similarly, for Algorithm \ref{algo:ldp-ncb},
\[
    \prob\{\widehat\mu_{j,s} \ge \tfrac{\mu^*}{128}\}
    \;\le\;
    \exp\!\Bigl(-\tfrac{1}{3}\,\tfrac{\mu^*}{256}\,s\Bigr)
    = \exp\!\Bigl(-\tfrac{c^2\log T}{3} - \tfrac{(\log T)^2}{\mu^*\epsilon^2}\Bigr)
    \;\le\;
    \exp\!\Bigl(-\tfrac{c^2\log T}{3}\Bigr)
    =
    \tfrac{1}{T^3}.
\]

A union bound across all arms and rounds then yields
\begin{align}
    \prob\{\cE_3^c\}
    \;\le\;
    kT \,\cdot \tfrac{1}{T^3}
    \;\leq\;
    \tfrac{1}{T} 
    \label{ineq:probG3c:anytime}
\end{align}

Finally, we bound $\prob\{\cE_4^c\}$. To bound this probability for the GDP setup, we utilise the following lemma\\

\begin{lemma}[Concentration bound on Laplace Noise] 
\label{lem:concLap}
Let $X\sim\Lap({b})$. Then we have $\prob\{|X|\geq b\log 1/\delta\}\leq \delta$

\end{lemma}

It is clear that $|\widetilde{\mu}_{i,n}-\widehat{\mu}_{i,n}|$ is a Laplace random variable. Using Lemma \ref{lem:concLap} with  $\delta = T^{-\alpha}$ and $b=\frac{\log T}{\epsilon\  n}$, we get
\begin{align*}
    \prob\{|\widetilde{\mu}_{i,n}-\widehat{\mu}_{i,n}|\geq \frac{\alpha (\log T)^2}{\epsilon \ n}\}\leq T^{-\alpha}
\end{align*}

Again, via union bound across all arms and rounds, we get, under the GDP setup
\begin{align}
    \prob\{\cE_4^c\}
    \;\le\;
    kT \,\cdot \tfrac{1}{T^\alpha}
    \;\leq\;
    \tfrac{1}{T}
    \label{ineq:probG4c}
\end{align}

For the LDP case, we utilise the following lemma from \citet{ren2020localdifferentialprivacy}

\begin{lemma}[Bound on sum of Laplace Random Variables] \label{lem:concBoundLDP}
If $X_1, X_2\dots X_n \sim \Lap(b)$ and $Y_n=\sum_{i=1}^{n}X_i$, then for $v \geq b \sqrt{n}$ and $0 \leq \lambda \leq \frac{2\sqrt2 v^2}{b}$, we have $\prob\{Y_n \geq \lambda \} \leq \exp{\frac{-\lambda^2}{8v^2}}$
\end{lemma}

Now suppose we have real-valued observations \(X_1,\dots,X_n\) and i.i.d.\ noise variables
\[
N_1, N_2, \dots, N_n \;\sim\; \mathrm{Lap}(b).
\]
We form
\[
\hat\mu_{i,n} \;=\;\frac{1}{n}\sum_{i=1}^n X_i,
\qquad
\tilde\mu_{i,n} \;=\;\frac{1}{n}\sum_{i=1}^n \bigl(X_i + N_i\bigr).
\]
Then the \emph{error} due to the noise is
\[
\tilde\mu_{i,n} - \hat\mu_{i,n}
\;=\;\frac{1}{n}\sum_{i=1}^n N_i
\;=\;\frac{Y_n}{n},
\qquad
Y_n \;=\;\sum_{i=1}^n N_i.
\]

Note that by symmetry in Lemma \ref{lem:concBoundLDP}, the same bound holds for \(\Pr\{Y_n<-\lambda\}\).  Hence
\[
\Pr\bigl\{|Y_n| > \lambda\bigr\}
\;\le\;
2\exp\!\Bigl(-\frac{\lambda^2}{8b^2n}\Bigr).
\]

Taking $\lambda
=\;
b\sqrt{8\,n\,\alpha\ln T }$, we have

\begin{align*}
&\Pr\bigl\{|Y_n| > b\sqrt{8\,n\,\alpha\ln T}\bigr\}
\;\le\;
2\exp\!\Bigl(-\frac{{8b^2n\,\alpha\ln T }}{8b^2n}\Bigr)=\frac{2}{T^\alpha}\\
&\implies\Pr\!\Bigl\{\,|\tilde\mu_{i,n} - \hat\mu_{i,n}| \;>\;
b\sqrt{\frac{8\,\alpha\ln T }{n}}
\Bigr\}
\;\le\;\frac{2}{T^\alpha}.
\end{align*}
Lastly, via union bound, we get
\begin{align}
    \prob\{\cE_4^c\}
    \;\le\;
    kT \,\cdot \tfrac{2}{T^\alpha}
    \;\leq\;
    \tfrac{2}{T}
    \label{ineq:probG4cL}
\end{align}\\
So, from \eqref{ineq:probG4c} and \eqref{ineq:probG4cL}, we get
\begin{align}
    \prob\{\cE_4^c\}
    \;\le\;\max\{\frac{1}{T},\frac{2}{T}\}=\frac{2}{T} \label{ineq:probG4cFinal}
\end{align}\\

Combining \eqref{ineq:probG1c:anytime}, \eqref{ineq:probG1c:anytime_ldp}, \eqref{ineq:probG2c:anytime},  \eqref{ineq:probG3c:anytime}, and
\eqref{ineq:probG4cFinal}, we conclude
\[
    \prob\{\cE\}
    = 1 - \prob\{\cE^c\}
    \;\ge\;
    1 - \prob\{\cE_1^c\} - \prob\{\cE_2^c\} - \prob\{\cE_3^c\} - \prob\{\cE_4^c\}
    \;\ge\;
    1 - \tfrac{6}{T},
\]
as claimed.



\subsection{Proof of Theorem \ref{theorem:improvedNashRegret}}
\label{appendix:modifiedncb-supporting-lemmas}

We first present a useful numerical inequality from \citet{barman2023fairness}, which we will use several times in our analysis.
\begin{restatable}{claim}{LemmaNumeric}
\label{lem:binomial}
For all reals $x \in \left[0, \frac{1}{2}\right]$ and all $a \in [0,1]$, we have $(1-x)^{a} \geq  1- 2ax$. 
\end{restatable}

\begin{proof}
The binomial theorem gives us
\begin{equation}\label{eq:binomial}
(1 - x)^a
= 1 - a x
  + \frac{a(a - 1)}{2!}\,x^2
  - \frac{a(a - 1)(a - 2)}{3!}\,x^3
  + \cdots
\end{equation}
Re-grouping,
\[
  (1 - x)^a
  = 1 - a x
    - a x
      \Bigl(
        \underbrace{\frac{(1 - a)}{2!}\,x
        + \frac{(1 - a)(2 - a)}{3!}\,x^2
        + \frac{(1 - a)(2 - a)(3 - a)}{4!}\,x^3
        + \cdots}_{S}
      \Bigr).
\]
We can bound the parenthesized series \(S\) as follows (since \(a\in(0,1)\)):
\[
  S
  \;=\;
  \frac{(1-a)}{2!}\,x
  + \frac{(1-a)(2-a)}{3!}\,x^2
  + \frac{(1-a)(2-a)(3-a)}{4!}\,x^3
  + \cdots
  \;\le\;
  \frac{1}{2!}\,x
  + \frac{1\cdot2}{3!}\,x^2
  + \frac{1\cdot2\cdot3}{4!}\,x^3
  + \cdots
  = \frac{x}{1 - x},
\]
where in the last step we used the Taylor series for \(1/(1-x)\) (and \(x<1\)).  

Hence from \eqref{eq:binomial} we obtain
\[
  (1 - x)^a
  \;\ge\;
  1 - a x
  - a x \,\frac{x}{1 - x}.
\]
Finally, if \(x \le \tfrac12\) then \(\frac{x}{1 - x}\le1\), so
\[
  (1 - x)^a \;>\; 1 - 2 a x,
\]
which completes the proof.
\end{proof}

We now show that on the favourable event $\cE$, each arm's total reward remains bounded until it has been sampled a specified number of times.
This result is useful in controlling the duration of Phase~I in Algorithm~\ref{algo:gdp-ncb}. The following three lemmas help us establish this result.

\begin{restatable}{lemma}{LemmaTauLower}
\label{tau:lower}
Conditioned on $\cE$, for any arm $i$ and every sample count $n\le 768\,S$, we have
\[
  n\,\widetilde\mu_{i,n} 
  < 800 c^2\log T + (800+\alpha)\frac{(\log T)^2}{\epsilon}.
\]
\end{restatable}

\begin{proof}

Let us set 
\[
N \coloneqq 768 \Sample.
\]
Observe that for any fixed arm \(i\), the quantity \(n \,\widehat{\mu}_{i,n}\) exactly equals the cumulative reward obtained from arm \(i\) over its first \(n\) pulls. Hence, for every \(n \le N\), we deduce
\begin{align}
n \,\widehat{\mu}_{i, n} \leq N \,\widehat{\mu}_{i,N} \label{ineq:sumemp}
\end{align}

We begin by handling those arms \(j\) whose true means satisfy \(\mu_j \le \frac{\mu^*}{256}\). For any such arm \(j\), event \(\cE_3\) ensures that \(\widehat{\mu}_{j,N} \leq \frac{\mu^*}{128}\). Then, applying inequality (\ref{ineq:sumemp}), we obtain
\begin{align*}
&n \,\widehat{\mu}_{j, n}
\underset{\text{by (\ref{ineq:sumemp})}}{\leq}
N \,\widehat{\mu}_{j,N}
\leq
768 \Sample \,\frac{\mu^*}{128}
=
6 \Bigl(c^2 \log T + \tfrac{(\log T)^2}{\epsilon}\Bigr).\\
\implies 
& n \,\widetilde{\mu}_{j, n}  \leq n \,\widehat{\mu}_{j, n} + \frac{\alpha (\log T)^2}{\epsilon} \leq 6 c^2\log T + (6+\alpha)\frac{(\log T)^2}{\epsilon} \tag{via event $\cE_4$}
\end{align*}

This completes the proof for arms with means \(\mu_j \leq \frac{\mu^*}{256}\). For arms \(i\) with means satisfying \(\mu_i \ge \frac{\mu^*}{256}\), we have
\begin{align*}
\widehat{\mu}_{i,N}
&\leq \mu_i + c \,\sqrt{\frac{\mu_{i} \,\log T}{N}}\tag{by event \(\cE_2\)}\\
&\leq \mu^* + c \,\sqrt{\frac{\mu^* \,\log T}{N}}\tag{since  $\mu_i \le \mu^*$}\\
&< \mu^* + \frac{\mu^*}{\sqrt{768}}
\tag{as $N = 768 \Sample > \frac{768\,c^2\log T}{\mu^*}$}\\
&< \frac{800}{768}\,\mu^*.
\end{align*}
Consequently, for those arms with \(\mu_i \ge \frac{\mu^*}{256}\), we conclude
\begin{align*}
    &N \,\widehat{\mu}_{i,N}
< \frac{800}{768}\,\mu^* \,\cdot 768 \Sample
= 800 \Bigl(c^2\log T + \frac{(\log T)^2}{\epsilon}\Bigr).\\
&\implies N \,\widetilde{\mu}_{i,N} \leq N \,\widehat{\mu}_{i,N}+\frac{\alpha (\log T)^2}{\epsilon} \le 800 c^2\log T + (800+\alpha)\frac{(\log T)^2}{\epsilon} \tag{via event $\cE_4$}
\end{align*}

This completes the proof.
\end{proof}

\begin{restatable}{lemma}{LemmaTauUpper}
\label{tau:upper}
On event $\cE$, once the optimal arm $i^{*}$ has collected $n\ge 1936\,S$ samples, we have
\[
  n\,\widetilde\mu_{i^{*},n} 
  \ge 1892 c^2\log T + (1892-\alpha)\tfrac{(\log T)^2}{\epsilon}.
\]
\end{restatable}
\begin{proof}
Set 
\[
M \coloneqq 1936 \Sample.
\]
Note that for every \(n \ge M\), the relation
\[
n \,\widehat{\mu}_{j,n} \ge M \,\widehat{\mu}_{j,M}
\]
holds by the definition of empirical means. Next, we bound \(\widehat{\mu}_{i^*,M}\) from below. Invoking event \(\cE_2\), we get
\begin{align*}
\widehat{\mu}_{i^*,M}
&\ge \mu^* - c \,\sqrt{\frac{\mu^* \,\log T}{M}}
\quad\tag{by \(\cE_2\)}\\
&= \mu^* - \frac{\mu^*}{\sqrt{1936}}
\tag{since $M = 1936 \Sample > \tfrac{1936\,c^2\log T}{\mu^*}$}\\
&> \tfrac{43}{44}\,\mu^*.
\end{align*}
Therefore, for any \(n \ge M = 1936\,\Sample\), it follows that
\begin{align*}
& n \,\widehat{\mu}_{i^*,n}
\ge M \,\widehat{\mu}_{i^*,M}
> 1936\,\Sample \,\cdot \tfrac{43}{44}\,\mu^*
= 1892 \Bigl(c^2\log T + \tfrac{(\log T)^2}{\epsilon}\Bigr)\\
\implies & n \,\widetilde{\mu}_{i^*,n} \geq n \,\widehat{\mu}_{i^*,n}-\frac{\alpha (\log T)^2}{\epsilon} >1892 c^2\log T + (1892-\alpha)\tfrac{(\log T)^2}{\epsilon} \tag{via event $\cE_4$}
\end{align*}
Where the first inequality holds via event $\cE_4$. This completes the proof.
\end{proof}

\begin{restatable}{lemma}{LemmaTauAnytime}
\label{tau:anytime}
Let $\tau$ be the count of uniform‐sampling rounds until some arm's cumulative reward exceeds
\[
  1600\Bigl(c^{2}\log T + \tfrac{(\log T)^{2}}{\epsilon}\Bigr).
\]
Then, under $\cE$, $\tau$ satisfies
\[
  512\,k\,S \;\le\; \tau \;\le\; 3872\,k\,S.
\]
\end{restatable}
\begin{proof}
Let 
\[
T_1 = 512\,k\,\Sample.
\]
Under the event \(\cE_1\), no arm has been sampled more than \(768\,\Sample\) times in the first \(T_1\) uniform‐sampling rounds. Therefore, for each arm \(i\), Lemma~\ref{tau:lower} yields
\[
n_i\,\widetilde\mu_i 
< 800 c^2\log T + (800+\alpha)\tfrac{(\log T)^2}{\epsilon}
\le 1600\bigl(c^2\log T + \tfrac{(\log T)^2}{\epsilon}\bigr).
\]
Since \(\tau\) is the first time any arm’s cumulative estimate exceeds \(1600\bigl(c^2\log T + \tfrac{(\log T)^2}{\epsilon}\bigr)\), it follows that
\[
\tau \ge T_1 = 512\,k\,\Sample.
\]

Next, set
\[
T_2 = 3872\,k\,\Sample.
\]
On the event \(\cE\), each arm \(i\) is sampled at least \(1936\,\Sample\) times by round \(T_2\).  Applying Lemma~\ref{tau:upper} to the optimal arm \(i^*\) gives
\[
n_{i^*}\,\widetilde\mu_{i^*}
\ge 1892c^2\log T + (1892-\alpha)\tfrac{(\log T)^2}{\epsilon}
> 1600\bigl(c^2\log T + \tfrac{(\log T)^2}{\epsilon}\bigr).
\]
Hence the stopping condition is met by round \(T_2\), so
\[
\tau \le T_2 = 3872\,k\,\Sample.
\]
This completes the proof.
\end{proof}

Since the events $\cE_1$, $\cE_2$, $\cE_3$, and $\cE_4$ are formulated in the standard stochastic bandit framework, 
Lemmas~\ref{tau:lower}, \ref{tau:upper}, and \ref{tau:anytime} likewise apply in that setting.
\hfill\\



\begin{restatable}{lemma}{LemmaNCBoptAnytime}
\label{lem:emp:anytime}
Let $\NCB_{\GDP}(i^{*},t)$ be the Nash confidence bound for the optimal arm $i^{*}$ at time $t$ in Phase~II.
Under the event $\cE$, it holds that for every round $t$ in Phase~II,
\[
  \NCB_{\GDP}(i^{*},t) \ge \mu^{*}.
\]
\end{restatable}
\begin{proof}

Consider an arbitrary round $t$ in Phase II. Denote by $n_{i^*}$ the number of times the optimal arm $i^*$ has been selected up to (but not including) round $t$, and let $\widehat\mu_{i^*}$ be its corresponding empirical mean at round $t$. By definition, the Nash confidence bound at this round can be written as
\[
\NCB_{GDP}(i^*,t)
= \widetilde\mu_{i^*}
+ 2c \sqrt{\frac{2\,\widetilde\mu_{i^*}\log T}{n_{i^*}}}
+ \frac{\alpha\,(\log T)^2}{\epsilon\,n_{i^*}}
+ 4\sqrt{\frac{2\alpha}{\epsilon}}\;\frac{(\log T)^{3/2}}{n_{i^*}}.
\]

Since event $\cE$ holds, Lemma \ref{tau:anytime} guarantees that Algorithm \ref{algo:gdp-ncb} remains in Phase I for at least $512\,k\,\Sample$ rounds before transitioning to Phase II; indeed, the first while-loop terminates only once this many iterations have occurred. Combining this with event $\cE_1$ yields
\[
n_{i^*}\;\ge\;256\,\Sample.
\]
Hence,
\[
n_{i^*}\,\mu^* \;\ge\;256\,c^2\log T, \tag{since $S \geq \frac{c^2 \log T}{\mu^*}$}
\]
which in turn allows us to lower‐bound the empirical mean of the optimal arm as follows

\begin{align}
\widehat{\mu}^* & \geq \mu^* - c \sqrt{\frac{\mu^* \log T}{n_{i^*}}} \tag{since $n_{i^*} \geq 256 \Sample$ and event $\cE_2$ holds}    \\
                        & = \mu^* - c \mu^* \sqrt{\frac{\log T}{\mu^* n_{i^*}}} \nonumber \\
                        & \geq \mu^* - c \mu^* \sqrt{\frac{1}{256c^2}}                  \tag{since  $\mu^* \ n_{i^*} \geq 256c^2\log T$}   \\
                        & = \frac{15}{16} \mu^*. \nonumber
    \end{align}
    Therefore,
    \begin{align*}
        {\NCB_{GDP}(i^*,t)} & = \widetilde{\mu}^* + 2c \sqrt{\frac{2\widetilde{\mu}^* \log T}{n_{i^*}}} + \frac{\alpha (\log T)^2}{\epsilon n_{i^*}} +4\sqrt{\frac{2\alpha}{\epsilon}}\frac{(\log T)^\frac{3}{2}}{n_{i^*}}              \\
        & \geq \widehat{\mu}^*+2c\sqrt{\frac{2\widehat{\mu}^*\log T}{n_{i^*}}}-\vert Lap\left(\frac{\log T}{\epsilon {n_{i^*}}}\right)\vert -4c\sqrt{\frac{2\vert Lap\left(\frac{\log T}{\epsilon {n_i^*}}\right)\vert\log T}{n_{i^*}}}\\& +\frac{\alpha (\log T)^2}{\epsilon n_{i^*}} +  4\sqrt{\frac{2\alpha}{\epsilon}}\frac{(\log T)^\frac{3}{2}}{n_{i^*}}
        \\
	& \geq \widehat{\mu}^* + 2c \sqrt{\frac{2\widehat{\mu}^* \log T}{n_{i^*}}} \tag{via event $\cE_4$}\\
    & \geq \widehat{\mu}^* + 2c \sqrt{\frac{\widehat{\mu}^* \log T}{n_{i^*}}} \\
                       & \geq  \mu^* - c \sqrt{\frac{\mu^* \log T}{n_{i^*}}} + 2c \sqrt{\frac{\widehat{\mu}^* \log T}{n_{i^*}}} \tag{due to the event $\cE_{2}$}\\
                       & \geq  \mu^* - c \sqrt{\frac{\mu^* \log T}{n_{i^*}}}  + 2c \sqrt{\frac{15\mu^* \log T}{16n_{i^*}}}     \tag{since $ \widehat{\mu}^*\geq  \frac{15}{16} \mu^*$}   \\
                       & \geq \mu^*
    \end{align*}
    The lemma stands proved.
\end{proof}
\begin{restatable}{lemma}{LemmaBadArmsAnytime}
\label{lem:bad_arms:anytime}
Assuming $\cE$, any arm $j$ with mean $\mu_{j}\le \mu^{*}/256$ is never chosen in Phase~II.
\end{restatable}

\begin{proof}

Let $j$ be any arm with true mean satisfying $\mu_j \le \mu^*/256$. Define $r_j$ as the total number of times arm $j$ is selected during Phase I. We first establish that 
\[
r_j \;\ge\;256\,\Sample.
\]
Indeed, since event $\cE$ holds, Lemma \ref{tau:anytime} implies that Algorithm \ref{algo:gdp-ncb} remains in Phase I for no fewer than $512\,k\,\Sample$ iterations before entering Phase II. Combined with the uniform‐exploration guarantee from event $\cE_1$, this yields $r_j \ge 256\,\Sample$.

Next, under event $\cE_3$ and using $r_j \ge 256\,\Sample$, we deduce that throughout Phase II the empirical mean of arm $j$ obeys
\[
\widehat\mu_j \;\le\;\frac{\mu^*}{128}.
\]

Finally, for any round $t$ in Phase II, let $NCB_{GDP}(j,t)$ denote the Nash confidence bound for arm $j$. We now show
\[
\NCB_{GDP}(j,t)
\;<\;
\NCB_{GDP}(i^*,t),
\]
which implies that arm $j$ is never selected during Phase II. We have
\begingroup
\allowdisplaybreaks
    \begin{align*}
        {\NCB_{GDP}(j,t)} & \leq \widetilde{\mu}_j + 2c\sqrt{\frac{2\widetilde{\mu}_j \log T}{r_j}}+ \frac{\alpha (\log T)^2}{\epsilon r_j} +4\sqrt{\frac{2\alpha}{\epsilon}}\frac{(\log T)^\frac{3}{2}}{r_j}                   \\
                   & \leq \widehat{\mu}_j + 2c\sqrt{\frac{2\widehat{\mu}_j \log T}{r_j}}+ \frac{2\alpha (\log T)^2}{\epsilon r_j} +8\sqrt{\frac{2\alpha}{\epsilon}}\frac{(\log T)^\frac{3}{2}}{r_j}             \tag{due to event $\cE_4$}          \\
                   & \leq \frac{\mu^*}{128} + 2c\sqrt{\frac{\mu^* \log T}{64 r_j} } +\frac{2\alpha (\log T)^2}{\epsilon r_j} +8\sqrt{\frac{2\alpha}{\epsilon}}\frac{(\log T)^\frac{3}{2}}{r_j}     \tag{since $\widehat{\mu}_j \leq \frac{\mu^*}{128}$}                       \\
                   & \leq \frac{\mu^*}{128} + \frac{c}{4}\sqrt{\frac{\mu^* \log T}{256 S}}+\frac{2\alpha (\log T)^2}{256\epsilon S} +8\sqrt{\frac{2\alpha}{\epsilon}}\frac{(\log T)^\frac{3}{2}}{256S} \tag{since $r_j \geq 256 S$}\\
                   & \leq \frac{\mu^*}{128} + \frac{\mu^*}{64}+\frac{2\alpha (\log T)^2}{256\epsilon S} +8\sqrt{\frac{2\alpha}{\epsilon}}\frac{(\log T)^\frac{3}{2}}{256S}                       \tag{$S>\frac{c^2\log T}{\mu^*}$}                         \\
                   & \leq \frac{\mu^*}{128} + \frac{\mu^*}{64}+\frac{\mu^*\alpha}{128} +\frac{\mu^*\sqrt{2\alpha\epsilon}}{32\sqrt{\log T}}                         \tag{$S>\frac{(\log T)^2}{\mu^*\epsilon}$}                         \\
                   & = \mu^*\left(\frac{3}{128} +\frac{\alpha}{128} +\frac{\sqrt{2\alpha\epsilon}}{32\sqrt{\log T}}\right)\leq \mu^* \tag{For $\alpha=3.1$ and large enough $T$}\\
                   & < \NCB_{GDP}(i^*,t) \tag{via Lemma \ref{lem:emp:anytime}}
    \end{align*}
\endgroup

\end{proof}
\begin{restatable}{lemma}{LemmaHighMeanAnytime}
\label{lem:suboptimal_arms:anytime}
Under the event $\cE$, let $T_{i}$ be the count of pulls of arm $i$ by Algorithm~\ref{algo:gdp-ncb}.
For any arm $i$ that appears at least once in Phase~II, we have
\[
  \mu^{*} \le \mu_{i} + 4c\sqrt{\frac{\mu^{*}\,\log T}{T_{i}-1}} 
  + \frac{2\alpha\,(\log T)^{2}}{\epsilon\,(T_{i}-1)} 
  + 8\sqrt{\frac{2\alpha}{\epsilon}}\frac{(\log T)^{3/2}}{T_{i}-1}
\]
\end{restatable}

\begin{proof}

Let $i$ be any arm that is chosen at least once during Phase II, and suppose its $T_i$‑th selection in Phase II occurs at some round $t$.  By definition of the selection rule, at round $t$ the arm $i$ must satisfy
\[
\NCB_{GDP}(i,t)\;\ge\;\NCB_{GDP}(i^*,t)
\;\ge\;\mu^*,
\]
where the final inequality is Lemma \ref{lem:emp:anytime}.  Hence, denoting by $\widetilde\mu_i$ the corrected empirical mean of arm $i$ just before its $T_i$‑th pull, we have
\begin{equation}\label{eq:tilde-mu-bound}
\widetilde{\mu}_i
+2c\sqrt{\frac{2\,\widetilde{\mu}_i\log T}{T_i-1}}
+\frac{\alpha(\log T)^2}{\epsilon\,(T_i-1)}
+4\sqrt{\frac{2\alpha}{\epsilon}}\;\frac{(\log T)^{3/2}}{T_i-1}
\;\ge\;\mu^*.
\end{equation}

Moreover, under event $\cE$, any arm that appears in Phase II has been pulled at least $256\,\Sample$ times in Phase I (by Lemmas \ref{lem:emp:anytime} and \ref{lem:bad_arms:anytime}), so $T_i>256\,\Sample$.  Since arm $i$ survives into Phase II, Lemma \ref{lem:bad_arms:anytime} also ensures
\[
\mu_i \;>\;\frac{\mu^*}{256}.
\]

We now turn to an upper bound on the empirical mean $\widehat\mu_i$ in terms of $\mu^*$ to complement \eqref{eq:tilde-mu-bound}. We have

\begin{align}
\widehat{\mu_i} & \leq  \mu_i + c\sqrt{\frac{\mu_i \log T}{T_i -1}}                \tag{since $\mu_i >\frac{\mu^*}{256}$ and event $\cE_{2}$ holds} \nonumber                                                    \\
& \leq \mu^* +c\sqrt{\frac{\mu^* \log T}{ 256 \Sample}} \tag{since $T_i> 256 \Sample$ and $\mu_i \leq \mu^*$}        \\
& = \mu^* + \frac{\mu^*}{16}  \tag{since $\Sample > \frac{c^2 \log T}{\mu^*}$}\\
& = \frac{17}{16} \mu^*  \label{ineq:hatmuo:anytime}
\end{align}

    Inequalities (\ref{eq:tilde-mu-bound}) and (\ref{ineq:hatmuo:anytime}) give us
    \begin{align*}
        &  \mu^* \leq\widetilde{\mu}_i + 2c \sqrt{\frac{2\widetilde{\mu}_i \log T}{T_i-1}} + \frac{\alpha (\log T)^2}{\epsilon (T_i-1)} +4\sqrt{\frac{2\alpha}{\epsilon}}\frac{(\log T)^\frac{3}{2}}{T_i-1}\\
         & \leq\widehat{\mu}_i + 2c \sqrt{\frac{17\mu^* \log T}{8(T_i-1)}} + \frac{2\alpha (\log T)^2}{\epsilon (T_i-1)} +8\sqrt{\frac{2\alpha}{\epsilon}}\frac{(\log T)^\frac{3}{2}}{T_i-1}                        \tag{via event $\cE_4$}                                         \\
         & \leq \mu_i + c\sqrt{\frac{\mu_i \log T}{T_i -1 }} + 3c\sqrt{\frac{\mu^* \log T}{T_i -1}}+\frac{2\alpha (\log T)^2}{\epsilon (T_i-1)} +8\sqrt{\frac{2\alpha}{\epsilon}}\frac{(\log T)^\frac{3}{2}}{T_i-1}   \tag{via event $\cE_{2}$}                           \\
         & \leq \mu_i + c\sqrt{\frac{\mu^* \log T}{T_i  -1}} + 3c\sqrt{\frac{\mu^* \log T}{T_i -1}} +\frac{2\alpha (\log T)^2}{\epsilon (T_i-1)} +8\sqrt{\frac{2\alpha}{\epsilon}}\frac{(\log T)^\frac{3}{2}}{T_i-1}                   \tag{since $\mu_i \leq \mu^*$} \\
         & \leq \mu_i +  4c\sqrt{\frac{\mu^* \log T}{T_i -1 }}+\frac{2\alpha (\log T)^2}{\epsilon (T_i-1)} +8\sqrt{\frac{2\alpha}{\epsilon}}\frac{(\log T)^\frac{3}{2}}{T_i-1}
    \end{align*}
    This completes the proof of the lemma.
\end{proof}
\hfill\\
We now derive a bound on Nash social welfare of Algorithm~\ref{algo:gdp-ncb} through the following lemma:
\begin{restatable}{lemma}{LemmaModifiedNCB}
\label{lem:modified_ncb}
Assume the optimal mean $\mu^{*}$ obeys
\begin{align*}
  \mu^{*} \ge \frac{512\sqrt{k\,\log T}}{\sqrt{T}} 
  + \frac{10000\,\log(256k)\,(\log T)^{2}}{\epsilon\,T},
\end{align*}

Then for any $w\le T$, the geometric mean of the expected rewards up to time $w$ satisfies
\begin{multline}
  \biggl(\prod_{t=1}^{w} \E[\mu_{I_{t}}]\biggr)^{1/T}
  \ge (\mu^{*})^{w/T} \biggl(1 - 4000c\sqrt{\frac{k\log T}{\mu^{*}T}}- \frac{4000\log(256k)\,(\log T)^{2}}{\mu^{*}\,\epsilon\,T}
  - \frac{4k\,\alpha\,(\log T)^{2}}{\mu^{*}\,\epsilon\,T}- \frac{16k}{T}\sqrt{\frac{2\alpha}{\epsilon}}\frac{(\log T)^{3/2}}{\mu^{*}}\biggr).
\end{multline}

\end{restatable}

\begin{proof}
We begin by deriving a lower bound on the expected reward $\mathbb{E}[\mu_{I_t}]$ of Algorithm~\ref{algo:gdp-ncb}, which will hold uniformly for all rounds $t \leq T$.

In particular, let $p_t$ denote the probability that the algorithm is in Phase I at round $t$. Then, with probability $p_t$, the algorithm performs uniform exploration, selecting an arm uniformly at random. In this case, the expected reward is at least $\frac{\mu^*}{k}$.

On the other hand, with probability $(1 - p_t)$, the algorithm is in Phase II. By Lemma~\ref{lem:bad_arms:anytime}, in this phase the expected reward is guaranteed to be at least $\frac{\mu^*}{256}$. Combining both scenarios, we obtain:

\begin{align}
 \E[\mu_{I_t}]&\geq \E \left[\mu_{I_t}|\cE \right] \ \prob\{\cE\} \nonumber \geq  \E \left[\mu_{I_t}|\cE \right] \left(1-\frac{6}{T}\right) \tag{Lemma \ref{lemma:modifiedgoodeventpr}}\\
 &=  \left(1-\frac{6}{T}\right)\left(p_t  \frac{\mu^*}{k} + (1-p_t)  \frac{\mu^*}{256}\right) \geq  \left(1-\frac{6}{T}\right)\left(\frac{\mu^*}{256k}\right) \label{ineq:UniformExpLB}
\end{align}

To facilitate the subsequent case analysis, set the cutoff 
\[
\overline{T} \;=\; 3872\,k\,\Sample.
\]
By Lemma \ref{tau:anytime}, the algorithm is guaranteed to have exited Phase I by round~\(\overline{T}\); equivalently, the stopping condition of the first while-loop (Line \ref{step:PhaseOneAlgTwo}) is met no later than this round.  Furthermore, under the stated lower‐bound assumption on \(\mu^*\) and for \(T\) taken sufficiently large, we have
\begin{align}
&\frac{\overline{T} \ \log \left(256k \right)}{T} =  \frac{3872 \ k \Sample \ \log \left( 256k\right) }{T}\\
& = \frac{3872 \ k c^2 \log T \ \log \left( 256k\right) }{\mu^* T}+\frac{3872\log (256k)(\log T)^2}{\mu^* \epsilon T}\\
&\leq \frac{3872 c^2 \log (256k) \sqrt{k \log T}}{512 \sqrt{T}}+\frac{3872\log (256k)(\log T)^2}{\mu^* \epsilon T} \tag{$\mu^*\geq \frac{512\sqrt{k\log T}}{\sqrt T}$} \\
&\leq \frac{3872 c^2 \log (256k) \sqrt{k \log T}}{512 \sqrt{T}}+{\frac{3872}{10000}} \tag{$\mu^*\geq \frac{10000\log (256k)(\log T)^2}{T\epsilon}$} \\
&\leq \frac{1}{2} +\frac{3872}{10000}\leq 1\label{ineq:overlineTexp}
\end{align} 

To prove the lemma, we split the argument into two exhaustive scenarios based on the round $w$:
\[
\text{Case 1: } w \le \overline{T}, 
\qquad
\text{Case 2: } w > \overline{T}.
\]

{\it Case 1} ($w \leq \overline T$): Invoking inequality (\ref{ineq:UniformExpLB}) we have 
\begingroup
\allowdisplaybreaks
\begin{align}
    \left(\prod_{t=1}^{w} \E [\mu_{I_t} ] \right)^\frac{1}{T}
     & \geq \left(1-\frac{6}{T}\right)^{\frac{w}{T}}\left(\frac{\mu^*}{256k}\right)^{\frac{w}{T}}\nonumber                                 \\
     & \geq \left(1-\frac{6}{T}\right) \left(\mu^*\right)^{\frac{w}{T}} \left(\frac{1}{256k}\right)^{\frac{w}{T}} \nonumber 
     = \left(1-\frac{6}{T}\right)\left(\mu^*\right)^{\frac{w}{T}} \left(\frac{1}{2}\right)^{\frac{w \  \log (256k)}{T}} \nonumber      \\
     & \geq \left(1-\frac{6}{T}\right)\left(\mu^*\right)^{\frac{w}{T}} \left(\frac{1}{2}\right)^{\frac{\overline{T} \  \log (256k)}{T}}  \tag{since $w \leq \overline{T}$} \\
     & = \left(1-\frac{6}{T}\right)\left(\mu^*\right)^{\frac{w}{T}} \left(1- \frac{1}{2}\right)^{\frac{\overline{T} \ \log (256k)}{T}} \nonumber   \\
     & \geq  \left(\mu^*\right)^{\frac{w}{T}} \left(1-{\frac{\overline{T}  \log (256k)}{T}}\right)\left(1-\frac{6}{T}\right) \tag{via inequality (\ref{ineq:overlineTexp}) and Claim \ref{lem:binomial}}\\
   & 	\geq \left(\mu^*\right)^{\frac{w}{T}} \left(1-{\frac{\overline{T}  \log (256k)}{T}}-\frac{6}{T}\right)\nonumber\\ 
   & = \bigg(\mu^*\bigg)^{\frac{w}{T}} \bigg(
       1 - \frac{3872 \, k c^2 \log T \, \log (256k)}{\mu^* T} \nonumber
       - \frac{3872 \log (256k) (\log T)^2}{\mu^* \epsilon T}
       - \frac{6}{T}
   \bigg)\\
  & = \left(\mu^*\right)^{\frac{w}{T}} \Bigg(
      1
      - \frac{3872c \sqrt{k \log T}}{\sqrt{\mu^* T}} \cdot \frac{c \log{(256k)} \sqrt{k \log T}}{\sqrt{\mu^* T}} \nonumber
      - \frac{3872\log (256k)(\log T)^2}{\mu^* \epsilon T}
      - \frac{6}{T}
    \Bigg) \\
      & \geq (\mu^*)^{\frac{w}{T}}\left(1-4000c\sqrt{\frac{ k \log T }{\mu^*T} }
      -\frac{4000\log (256k)(\log T)^2}{\mu^* \epsilon T}\right)\label{ineq:CaseOneW} \\
    & \geq (\mu^*)^{\frac{w}{T}}\bigg(1-4000c\sqrt{\frac{ k \log T }{\mu^*T} }-\frac{4000\log (256k)(\log T)^2}{\mu^* \epsilon T} \nonumber -\frac{4k}{T}\frac{\alpha (\log T)^2}{\mu^*\epsilon } -\frac{16k}{T}\sqrt{\frac{2\alpha}{\epsilon}}\frac{(\log T)^\frac{3}{2}}{\mu^*}\bigg).\nonumber 
    \end{align}
\endgroup

Inequality \eqref{ineq:CaseOneW} immediately follows from the estimate
\[
\frac{c\,\log(256k)\,\sqrt{k\log T}}{\sqrt{\mu^*\,T}}\;\le\;1
\]
for all sufficiently large \(T\), using the fact that
\[
\mu^*\;\ge\;\frac{512\sqrt{k\log T}}{\sqrt{T}}.
\]

{\it Case 2} ($w>\overline{T}$): Now we separate the Nash social welfare into the following terms: 
\begin{align}
    \left(\prod_{t=1}^{w} \E \left[ \mu_{I_t} \right] \right)^\frac{1}{T} =  \left(\prod_{t=1}^{\overline{T}} \E \left[ \mu_{I_t} \right]\right)^\frac{1}{T} \left(\prod_{t=\overline{T} + 1}^{w}  \E \left[ \mu_{I_t} \right] \right)^\frac{1}{T} \label{eqn:splitW}
\end{align}

In the product decomposition, the first factor captures the total expected reward accumulated over rounds \(t \le \overline{T}\), while the second factor corresponds to rounds \(\overline{T} < t \le w\). We proceed by deriving separate lower bounds for each of these two factors.

We now derive an upper bound on the first summand appearing on the right‐hand side of \eqref{eqn:splitW}.

\begingroup
\allowdisplaybreaks
\begin{align}
    \left(\prod_{t=1}^{\overline{T}} \E [\mu_{I_t} ] \right)^\frac{1}{T} 
     & \geq \left(1-\frac{6}{T}\right)^{\frac{\overline{T}}{T}}\left(\frac{\mu^*}{256k}\right)^{\frac{\overline{T}}{T}} \tag{via inequality (\ref{ineq:UniformExpLB})}                                 \\
     & \geq \left(1-\frac{6}{T}\right) \left(\mu^*\right)^{\frac{\overline{T}}{T}} \left(\frac{1}{256k}\right)^{\frac{\overline{T}}{T}} = \left(1-\frac{6}{T}\right)\left(\mu^*\right)^{\frac{\overline{T}}{T}} \left(\frac{1}{2}\right)^{\frac{\overline{T}  \log (256k)}{T}} \nonumber  \\  
     & = \left(1-\frac{6}{T}\right)\left(\mu^*\right)^{\frac{\overline{T}}{T}} \left(1- \frac{1}{2}\right)^{\frac{\overline{T}  \log (256k)}{T}} \nonumber   \\
     & \geq  \left(\mu^*\right)^{\frac{\overline{T}}{T}} \left(1-{\frac{\overline{T}  \log (256k)}{T}}\right)\left(1-\frac{6}{T}\right)    \label{ineq:phaseone:anytime}
\end{align}
\endgroup
To justify the final inequality, observe that the exponent
\[
\frac{\overline{T}\,\log(256k)}{T}\;\le\;1
\]
by \eqref{ineq:overlineTexp}, and then invoke Claim \ref{lem:binomial}.

Next, we bound the second summand on the right‐hand side of \eqref{eqn:splitW}.

\begin{align}
    \left(\prod_{t=\overline{T}+ 1}^{w} \E\left[ \mu_{I_t} \right]\right)^\frac{1}{T}
     & \geq \E\left[\left( \prod_{t=\overline{T}+1}^{w} \mu_{I_t} \right)^\frac{1}{T}\right ] \tag{Multivariate Jensen's inequality} \nonumber               
     \\& \geq \E \left[\left( \prod_{t=\overline{T}+1}^{w} \mu_{I_t} \right)^\frac{1}{T} \;\middle|\; \cE \right]  \prob\{ \cE \} \label{ineq:interim:anytime}
\end{align}
By Lemma \ref{tau:anytime}, Phase I must have concluded by round \(\overline T\), so every \(t>\overline T\) lies in Phase II.  To bound the second term on the right‐hand side of \eqref{ineq:interim:anytime}, we restrict attention to the arms chosen after round \(\overline T\).  Reindex these arms as \(\{1,2,\dots,\ell\}\), and let \(m_i\ge1\) be the number of times arm \(i\) is pulled after \(\overline T\); note that
\[
\sum_{i=1}^\ell m_i \;=\; w - \overline T.
\]
Let \(T_i\) denote the total pulls of arm \(i\) over the entire algorithm, so that \(T_i - m_i\) is its number of pulls in the first \(\overline T\) rounds.  With this notation,
\[
\mathbb{E}\!\Bigl[\bigl(\prod_{t=\overline T+1}^T \mu_{I_t}\bigr)^{1/T}\Bigm|\cE\Bigr]
\;=\;
\mathbb{E}\!\Bigl[\prod_{i=1}^\ell\mu_i^{\,m_i/T}\Bigm|\cE\Bigr].
\]
Finally, since we condition on the good event \(\cE\), Lemma \ref{lem:suboptimal_arms:anytime} applies to each arm \(i\in[\ell]\), yielding the desired bound. Thus, we have

\begin{align}
    \E \left[\left( \prod_{t=\overline{T}+1}^{w} \ \mu_{I_t} \right)^\frac{1}{T} \;\middle|\; \cE \right] & = \E\left[\left( \prod_{i=1}^{\ell} \mu_{i}^\frac{m_i}{T}  \right)\;\middle|\; \cE \right] \nonumber \\                                                                                             
    &  \geq \E\left[\prod_{i=1}^{\ell}\left(\mu^* - 4c\sqrt{\frac{\mu^* \log T}{T_i -1 }}-\frac{2\alpha (\log T)^2}{\epsilon (T_i-1)} -8\sqrt{\frac{2\alpha}{\epsilon}}\frac{(\log T)^\frac{3}{2}}{T_i-1} \right)^\frac{m_i}{T} \;\middle|\; \cE \right]   \tag{Lemma \ref{lem:suboptimal_arms:anytime}} \nonumber                                                                                                      \\
   & = (\mu^*)^{\frac{w-\overline{T}}{T}} \E\left[\prod_{i=1}^{\ell}\left(1 - 4c\sqrt{\frac{\log T}{\mu^*(T_i -1 )}}-\frac{2\alpha (\log T)^2}{\mu^*\epsilon (T_i-1)} -8\sqrt{\frac{2\alpha}{\epsilon}}\frac{(\log T)^\frac{3}{2}}{\mu^*(T_i-1)} \right)^\frac{m_i}{T}	\;\middle|\; \cE \right] \label{ineq:dunzo:anytime}
\end{align}

In the equality above we have used the identity $\sum_{i=1}^\ell m_i = w - \overline{T}$.  Moreover, under the good event $\cE$, each arm is guaranteed to be sampled at least $256\,\Sample$ times within the first $\overline T$ rounds.  Consequently, for every $i\in[\ell]$,
\[
T_i \;=\;(T_i - m_i) + m_i
\;>\;256\,\Sample + 1
\;\ge\;256\,\Sample,
\]
and hence the lower‐bound conditions of Lemma \ref{lem:suboptimal_arms:anytime} apply to each such arm. Using this, we get
\begin{align*}
& 4c\sqrt{\frac{\log T}{\mu^*(T_i -1 )}}+\frac{2\alpha (\log T)^2}{\mu^*\epsilon (T_i-1)} + 8\sqrt{\frac{2\alpha}{\epsilon}}\frac{(\log T)^\frac{3}{2}}{\mu^*(T_i-1)} \leq 4c\sqrt{\frac{\log T}{256c^2\log T}}+\frac{2\alpha (\log T)^2}{\mu^*\epsilon (T_i-1)} + 8\sqrt{\frac{2\alpha}{\epsilon}}\frac{(\log T)^\frac{3}{2}}{\mu^*(T_i-1)} &\tag{$ T_i > 256S > 256\frac{c^2\log T}{\mu^*}$} \\
&\qquad\qquad\qquad \leq \frac{1}{4}+\frac{2\alpha (\log T)^2}{256 (\log T)^2} + 8\sqrt{\frac{2\alpha}{\epsilon}}\frac{\epsilon (\log T)^\frac{3}{2}}{256 (\log T)^2}&\tag{$T_i > 256S > 256\frac{(\log T)^2}{\mu^*\epsilon}$} \\
&\qquad\qquad\qquad = \frac{1}{4}+\frac{\alpha}{128} +\frac{\sqrt{2\alpha\epsilon}}{32\sqrt{\log T}} \leq \frac{1}{2} &\tag{For large enough T}
\end{align*}

Since the above bound holds for every $i \in [\ell]$, applying Claim~\ref{lem:binomial} allows us to simplify the expectation in inequality~\eqref{ineq:dunzo:anytime} as follows:

\begin{align*}
    &\E \left[\prod_{i=1}^{\ell}\left(1 - 4c\sqrt{\frac{\log T}{\mu^*(T_i -1)}}-\frac{2\alpha (\log T)^2}{\mu^*\epsilon (T_i-1)} -8\sqrt{\frac{2\alpha}{\epsilon}}\frac{(\log T)^\frac{3}{2}}{\mu^*(T_i-1)} \right)^\frac{m_i}{T}\;\middle|\; \cE \right] \\
    & \geq \E \left[\prod_{i=1}^{\ell}\left(1 - \frac{8c\ m_i}{T}\sqrt{\frac{\log T}{\mu^*(T_i -1)}}-\frac{4m_i}{T}\frac{\alpha (\log T)^2}{\mu^*\epsilon (T_i-1)} -\frac{16m_i}{T}\sqrt{\frac{2\alpha}{\epsilon}}\frac{(\log T)^\frac{3}{2}}{\mu^*(T_i-1)} \right)\;\middle|\; \cE \right]                               \\
 & \geq \E \left[\prod_{i=1}^{\ell}\left(1 - \frac{8c}{T}\sqrt{\frac{m_i\log T}{\mu^*}}-\frac{4}{T}\frac{\alpha (\log T)^2}{\mu^*\epsilon } -\frac{16}{T}\sqrt{\frac{2\alpha}{\epsilon}}\frac{(\log T)^\frac{3}{2}}{\mu^*} \right)\;\middle|\; \cE \right]             & \text{(since $T_i \geq m_i + 1$)}
\end{align*}
In particular, because for any $x,y \ge 0$ we have 
$
(1 - x)(1 - y) \;\ge\; 1 - x - y,
$
we can replace the multiplicative term in the preceding bound by its corresponding linear lower bound. 

\begingroup
\allowdisplaybreaks
\begin{align*}
    &\E \left[\prod_{i=1}^{\ell}\left(1 - \frac{8c}{T}\sqrt{\frac{m_i\log T}{\mu^*}}-\frac{4}{T}\frac{\alpha (\log T)^2}{\mu^*\epsilon } -\frac{16}{T}\sqrt{\frac{2\alpha}{\epsilon}}\frac{(\log T)^\frac{3}{2}}{\mu^*} \right)\;\middle|\; \cE \right] \\ & \geq \E\left[1 -\sum_{i=1}^{\ell}\left(\frac{8c}{T}\sqrt{\frac{ m_i \log T }{\mu^*}} \right)\;\middle|\; \cE \right] -\frac{4k}{T}\frac{\alpha (\log T)^2}{\mu^*\epsilon }-\frac{16k}{T}\sqrt{\frac{2\alpha}{\epsilon}}\frac{(\log T)^\frac{3}{2}}{\mu^*}                                          \\
     & = 1 -\left(\frac{8c }{T}\sqrt{\frac{ \log T }{\mu^*}} \right) \E\left[ \sum_{i=1}^{\ell} \sqrt{m_i}\;\middle|\; \cE \right]  -\frac{4k}{T}\frac{\alpha (\log T)^2}{\mu^*\epsilon } -\frac{16k}{T}\sqrt{\frac{2\alpha}{\epsilon}}\frac{(\log T)^\frac{3}{2}}{\mu^*}      \\
     & \geq 1 -\left(\frac{8c}{T}\sqrt{\frac{ \log T }{\mu^*}} \right) \E\left[ \sqrt{\ell} \ \sqrt{\sum_{i=1}^\ell m_i} \;\middle|\; \cE \right] -\frac{4k}{T}\frac{\alpha (\log T)^2}{\mu^*\epsilon } -\frac{16k}{T}\sqrt{\frac{2\alpha}{\epsilon}}\frac{(\log T)^\frac{3}{2}}{\mu^*}\tag{Cauchy-Schwarz inequality} \\
     & \geq 1 -\left(\frac{8c }{T}\sqrt{\frac{ \log T }{\mu^*}} \right) \E\left[ \sqrt{\ell \ T} \;\middle|\; \cE \right]-\frac{4k}{T}\frac{\alpha (\log T)^2}{\mu^*\epsilon } -\frac{16k}{T}\sqrt{\frac{2\alpha}{\epsilon}}\frac{(\log T)^\frac{3}{2}}{\mu^*} \tag{since $\sum_i m_i \leq T$}                         \\
     & = 1 -\left( 8c \sqrt{\frac{ \log T }{\mu^* T }} \right) \E\left[ \sqrt{\ell} \;\middle|\; \cE \right] -\frac{4k}{T}\frac{\alpha (\log T)^2}{\mu^*\epsilon } -\frac{16k}{T}\sqrt{\frac{2\alpha}{\epsilon}}\frac{(\log T)^\frac{3}{2}}{\mu^*}                                                                     \\
     & \geq 1 - \left( 8c\sqrt{\frac{ k \log T }{\mu^* T }} \right)-\frac{4k}{T}\frac{\alpha (\log T)^2}{\mu^*\epsilon } -\frac{16k}{T}\sqrt{\frac{2\alpha}{\epsilon}}\frac{(\log T)^\frac{3}{2}}{\mu^*} \tag{since $\ell \leq k$}
\end{align*}
\endgroup

By substituting this estimate into inequalities~\eqref{ineq:interim:anytime} and~\eqref{ineq:dunzo:anytime}, we arrive at

\begin{align}
    \left(\prod_{t=\overline{T} + 1}^{w} \E\left[ \mu_{I_t} \right]\right)^\frac{1}{T} \geq (\mu^*)^{\frac{w-\overline{T}}{T}} \bigg ( 1 -8c\sqrt{\frac{ k \log T }{\mu^*T}} -\frac{4k}{T}\frac{\alpha (\log T)^2}{\mu^*\epsilon } -\frac{16k}{T}\sqrt{\frac{2\alpha}{\epsilon}}\frac{(\log T)^\frac{3}{2}}{\mu^*}\bigg) \prob \{\cE\} \label{ineq:toomany:anytime}
\end{align}
Utilizing the bounds from inequalities~\eqref{ineq:toomany:anytime} and~\eqref{ineq:phaseone:anytime} for the two terms in equation~\eqref{eqn:splitW}, we obtain the following lower bound on the algorithm’s Nash social welfare:

\begin{align*}
    &\left(\prod_{t=1}^{w} \E \left[ \mu_{I_t} \right] \right)^\frac{1}{T}
     \geq (\mu^*)^{\frac{w}{T}}\bigg(1-{\frac{\overline{T}\cdot \log (256k)}{T}}-\frac{6}{T}\bigg) \bigg( 1 -8c\sqrt{\frac{ k \log T }{\mu^*T} }-\frac{4k}{T}\frac{\alpha (\log T)^2}{\mu^*\epsilon } 
      -\frac{16k}{T}\sqrt{\frac{2\alpha}{\epsilon}}\frac{(\log T)^\frac{3}{2}}{\mu^*}\bigg) \prob\{ \cE \}                 \\
      & \geq (\mu^*)^{\frac{w}{T}} \left(1 - \frac{\overline{T} \cdot \log (256k)}{T} - \frac{6}{T} \right) \left(1 - 8c\sqrt{\frac{k \log T}{\mu^* T}} \right. 
\left. - \frac{4k}{T} \cdot \frac{\alpha (\log T)^2}{\mu^* \epsilon} - \frac{16k}{T} \sqrt{\frac{2\alpha}{\epsilon}} \cdot \frac{(\log T)^{3/2}}{\mu^*} \right) \left(1 - \frac{6}{T} \right) \tag{via Lemma \ref{lemma:modifiedgoodeventpr}}\\
     & \geq (\mu^*)^{\frac{w}{T}} \bigg(1-{\frac{\overline{T}\cdot \log (256k)}{T}}-8c\sqrt{\frac{ k \log T }{\mu^*T} }-\frac{4k}{T}\frac{\alpha (\log T)^2}{\mu^*\epsilon } -\frac{16k}{T}\sqrt{\frac{2\alpha}{\epsilon}}\frac{(\log T)^\frac{3}{2}}{\mu^*} -\frac{12}{T}\bigg)   \\                                              
     &= (\mu^*)^{\frac{w}{T}} \bigg(1 - \frac{3872kc^2 \log T \log (256k)}{\mu^* T} - \frac{3872 \log (256k) (\log T)^2}{\mu^* \epsilon T}  - 8c\sqrt{\frac{k \log T}{\mu^* T}} \\
     &\hspace{28em}- \frac{4k\alpha (\log T)^2}{T\mu^* \epsilon} - \frac{16k}{T} \sqrt{\frac{2\alpha}{\epsilon}} \cdot \frac{(\log T)^{3/2}}{\mu^*} - \frac{12}{T} \bigg)\\
& \geq (\mu^*)^{\frac{w}{T}}\bigg(1-4000c\sqrt{\frac{ k \log T }{\mu^*T} }-\frac{4000\log (256k)(\log T)^2}{\mu^* \epsilon T} -\frac{4k}{T}\frac{\alpha (\log T)^2}{\mu^*\epsilon }-\frac{16k}{T}\sqrt{\frac{2\alpha}{\epsilon}}\frac{(\log T)^\frac{3}{2}}{\mu^*}\bigg).
    \end{align*}
The last inequality is derived in the same manner as the final step of the proof of inequality~\eqref{ineq:CaseOneW}, thereby completing the argument and establishing the lemma.

\end{proof}

Finally, we use the above lemmas to prove Theorem \ref{theorem:improvedNashRegret}

\begin{proof}[Proof of Theorem \ref{theorem:improvedNashRegret}]
\hfill\\
 For $\mu^* = O\left(\sqrt {\frac{k\log T}{T}}+\frac{(\log T)^2 \log k}{\epsilon T}\right)$, the theorem holds trivially. We thus, consider the case when $\mu^*=\Omega\left(\sqrt {\frac{k\log T}{T}}+\frac{(\log T)^2 \log k}{\epsilon T}\right)$.
 Under this case, we can apply Lemma \ref{lem:modified_ncb} with $w = T$. Specifically, 
 \begin{align*}
     \NRg_T  &\leq \mu^* - (\mu^*)^{\frac{T}{T}}\bigg(1-4000c\sqrt{\frac{ k \log T }{\mu^*T} }-\frac{4000\log (256k)(\log T)^2}{\mu^* \epsilon T} -\frac{4k}{T}\frac{\alpha (\log T)^2}{\mu^*\epsilon } -\frac{16k}{T}\sqrt{\frac{2\alpha}{\epsilon}}\frac{(\log T)^\frac{3}{2}}{\mu^*}\bigg) \\
     &\leq(\mu^*)\bigg(4000c\sqrt{\frac{ k \log T }{\mu^*T} }+\frac{4000\log (256k)(\log T)^2}{\mu^* \epsilon T} +\frac{4k}{T}\frac{\alpha (\log T)^2}{\mu^*\epsilon } +\frac{16k}{T}\sqrt{\frac{2\alpha}{\epsilon}}\frac{(\log T)^\frac{3}{2}}{\mu^*}\bigg)\\
     &=\bigg(4000c\sqrt{\frac{ \mu^*k \log T }{T} }+\frac{4000\log (256k)(\log T)^2}{\epsilon T} +\frac{4k}{T}\frac{\alpha (\log T)^2}{\epsilon } +\frac{16k}{T}\sqrt{\frac{2\alpha}{\epsilon}}{(\log T)^\frac{3}{2}}\bigg)\\
    & = O\bigg(\sqrt{\frac {k\log T}{T}}+\frac{k(\log T)^2}{\epsilon T}\bigg) & \tag{$\mu^* \leq1$}
 \end{align*}
 This completes the proof of the theorem. 
\end{proof}


\subsection{Proof of Theorem \ref{theorem:improvedNashRegretLDP}}

We first reformulate Lemmas~\ref{tau:lower}-\ref{lem:suboptimal_arms:anytime} into the following lemmas to account for the Laplace noise introduced by Local Differential Privacy.
All statements continue to hold on the “good” event \(\cE\).

\begin{restatable}{lemma}{LemmaTauLowerLDP}
\label{tau:lower_ldp}
On event $\cE$, for any arm $i$ and every sample count $n\le 768\,S$, we have
\[
  n\,\widetilde{\mu}_{i,n} < 800\Bigl(c^{2}\log T + \tfrac{(\log T)^{2}}{\max\left(0,\ \widetilde\mu_{i,n} - \tfrac{1}{\epsilon}\sqrt{\tfrac{8\alpha\log T}{n}}\right)\,\epsilon^{2}}\Bigr)
  + \tfrac{\sqrt{8\,n\,\alpha\log T}}{\epsilon}.
\]
\end{restatable}
\begin{proof}
Define 
\[
N \;\coloneqq\; 768\,\Sample.
\]
Observe that, for any arm \(i\) and any sample size \(n\), the quantity \(n\,\widehat\mu_{i,n}\) equals the total reward accumulated by arm \(i\) in its first \(n\) pulls.  Hence, for all \(n\le N\),
\begin{equation}\label{ineq:sumempldp}
n\,\widehat\mu_{i,n}
\;\le\;
N\,\widehat\mu_{i,N}.
\end{equation}

We now specialize to any arm \(j\) whose true mean satisfies \(\mu_j\le\mu^*/256\).  By event \(\cE\), arm \(j\) is pulled at least \(256\,\Sample\) times in Phase I, and since \(256\,\Sample\le N\), inequality \eqref{ineq:sumempldp} applied at \(n=r_j\) gives
\[
r_j\,\widehat\mu_{j,r_j}
\;\le\;
N\,\widehat\mu_{j,N}.
\]
In particular, the bound \(\widehat\mu_{j,N}\le\mu_j + \sqrt{\tfrac{2\mu_j\log T}{N}}\) (by the concentration guarantee in event \(\cE_3\)) then yields
\[
r_j\,\widehat\mu_{j,r_j}
\;\le\;
N\Bigl(\mu_j + \sqrt{\tfrac{2\mu_j\log T}{N}}\Bigr)
\;\le\;
N\Bigl(\tfrac{\mu^*}{256} + \sqrt{\tfrac{2(\mu^*/256)\log T}{N}}\Bigr),
\]
which suffices to establish the lemma for all arms with \(\mu_j\le\mu^*/256\).


%

We first consider the case where $n<=N_1$ where $N_1 = 768 \frac{c^2\log T}{\mu^*}$.
\begin{align}
    n\ \widetilde{\mu}_{j,n}\leq n\left(\widehat{\mu}_{j,n} + \frac{1}{\epsilon}\sqrt{\frac{8\alpha\log T}{n}}\right) \leq N_1 \mu_{j,N_1} + \frac{\sqrt{8 n \alpha\log T}}{\epsilon}
\end{align}
By event $\cE_3$, $\mu_{j,N_1} \leq \frac{\mu^*}{128}$, therefore we have,
\begin{align}
    n\ \widetilde{\mu}_{j,n}\leq 6c^2\log T + \frac{\sqrt{8 n \alpha\log T}}{\epsilon} \leq 6\left(c^2\log T+\frac{(\log T)^2}{\max\left(0,\ \widetilde{\mu}_{j,n}-\frac{1}{\epsilon}\sqrt{\frac{8\alpha\log T}{n}}\right)\epsilon^2}\right)+\frac{\sqrt{8n\alpha\log T}}{\epsilon}
\end{align}

So we can be sure that any arm is pulled at least $N_1$ times. Note that for any such arm $j$, and $n>N_1$, event $\cE_3$ gives us $\widehat{\mu}_{j,n} \leq \frac{\mu^*}{128} \implies \widetilde{\mu}_{j,n}  \leq \frac{\mu^*}{128} + \frac{1}{\epsilon}\sqrt{\frac{8\alpha \log T}{n}}$ with high probability. Therefore we have, 
\begin{align*}
    n\ \widetilde{\mu}_{j,n}\leq n\left(\widehat{\mu}_{j,n} + \frac{1}{\epsilon}\sqrt{\frac{8\alpha\log T}{n}}\right) &\leq 768S\frac{\mu^*}{128}+\frac{\sqrt{8 n \alpha\log T}}{\epsilon}=6\left(c^2\log T+\frac{(\log T)^2}{\mu^*\epsilon^2}\right)+ \frac{\sqrt{8n\alpha\log T}}{\epsilon}\\
    &\leq 6\left(c^2\log T+\frac{(\log T)^2}{128\max\left(0, \ \widetilde{\mu}_{i,n}-\frac{1}{\epsilon}\sqrt{\frac{8\alpha\log T}{n}}\right)\epsilon^2}\right)+\frac{\sqrt{8n\alpha\log T}}{\epsilon}\\
    &\leq 6\left(c^2\log T+\frac{(\log T)^2}{\max\left(0,\ \widetilde{\mu}_{i,n}-\frac{1}{\epsilon}\sqrt{\frac{8\alpha\log T}{n}}\right)\epsilon^2}\right)+\frac{\sqrt{8n\alpha\log T}}{\epsilon}
\end{align*}
This completes the proof for arms with means \(\mu_j \leq \frac{\mu^*}{256}\). For arms \(i\) with means satisfying \(\mu_i \ge \frac{\mu^*}{256}\). For any such arm \(i\), we have
\begin{align*}
\widehat{\mu}_{i,N} & \leq \mu_i + c \sqrt{\frac{\mu_{i} \log T}{N}} \tag{via event $\cE_2$}\\
& \leq \mu^*+c \sqrt{\frac{\mu^* \log T}{N}}   \tag{since $\mu_i\leq \mu^*$}\\
& <\mu^*+\frac{\mu^*}{\sqrt{768}}     \tag{since $N=768 \Sample > \frac{768 c^2 \log T}{\mu^*}$}\\
& < \frac{800}{768} \mu^*
\end{align*}
Consequently, for those arms with \(\mu_i \ge \frac{\mu^*}{256}\), we conclude
\begin{align*}
n \ \widetilde{\mu}_{i,n}  \leq n\left(\widehat{\mu}_{i,n} + \frac{1}{\epsilon}\sqrt{\frac{8\alpha\log T}{n}}\right) &\leq \frac{800}{768} \mu^* \ 768 \Sample+\frac{\sqrt{8n\alpha\log T}}{\epsilon}=800\left(c^2\log T+\frac{(\log T)^2}{\mu^*\epsilon^2}\right)+\frac{\sqrt{8n\alpha\log T}}{\epsilon}\\
&\leq 800\left(c^2\log T+\frac{(\log T)^2}{128\max\left(0,\,\widetilde{\mu}_{i,n}-\frac{1}{\epsilon}\sqrt{\frac{8\alpha\log T}{n}}\right)\epsilon^2}\right)+\frac{\sqrt{8n\alpha\log T}} {\epsilon} \\
&\leq 800\left(c^2\log T+\frac{(\log T)^2}{\max\left(0,\,\widetilde{\mu}_{i,n}-\frac{1}{\epsilon}\sqrt{\frac{8\alpha\log T}{n}}\right)\epsilon^2}\right)+\frac{\sqrt{8n\alpha\log T}} {\epsilon} .
\end{align*}
\end{proof}

\begin{restatable}{lemma}{LemmaTauUpperLDP}
\label{tau:upper_ldp}
On event $\cE$, once the optimal arm $i^{*}$ has collected $n\ge 1936\,S$ samples, we have
\[
  n\,\widetilde{\mu}_{i^{*},n} \ge 1848\Bigl(c^{2}\log T + \tfrac{(\log T)^{2}}{\max\left(0,\ \widetilde\mu_{i,n} - \tfrac{1}{\epsilon}\sqrt{\tfrac{8\alpha\log T}{n}}\right)\,\epsilon^{2}}\Bigr)
  + \tfrac{\sqrt{8\,n\,\alpha\log T}}{\epsilon}.
\]
\end{restatable}
\begin{proof}
Set 
\[
M \coloneqq 1936 \Sample.
\]
Note that for every \(n \ge M\), the relation
\[
n \,\widehat{\mu}_{j,n} \ge M \,\widehat{\mu}_{j,M}
\]
holds by the definition of empirical means. Since this is a terminating condition, it is superflous to prove the lemma for $n \geq 1936 \Sample$. Thus, we will prove this lemma specifically for $M=1936\Sample$. Furthermore, 
\begin{align*}
\widehat{\mu}_{i^*,M} & \geq \mu^*-c \sqrt{\frac{\mu^* \log T}{M}} \tag{via event $\cE_2$} \\
& =\mu^*-\frac{\mu^*}{\sqrt{1936}}      \tag{since $M = 1936 \Sample > \frac{1936 c^2 \log T}{\mu^*}$}\\
& > \frac{43}{44} \mu^*
\end{align*}
Thus, the private means satisfy
\begin{align}
& \widetilde{\mu}_{i^*, M} \geq \widehat{\mu}_{i^*, M} - \frac{1}{\epsilon}\sqrt{\frac{8\alpha\log T}{M}}\\
&\implies \widetilde{\mu}_{i^*, M} -\frac{1}{\epsilon}\sqrt{\frac{8\alpha\log T}{M}}\geq \widehat{\mu}_{i^*, M} - \frac{2}{\epsilon}\sqrt{\frac{8\alpha\log T}{M}}\\
& \geq \widehat{\mu}_{i^*, M} - \frac{2}{\epsilon}\sqrt{\frac {8\alpha \log T (\mu^*\epsilon)^2}{(1936\log T)^2}} \tag{since $M = 1936S > 1936\frac{(\log T)^2}{(\mu^*\epsilon)^2}$}\\
& \geq \mu^* \left(\frac{43}{44}-\frac{2}{44}\sqrt{\frac{8\alpha}{\log T}} \right) \geq \frac{41\mu^*}{44} \tag{For sufficiently large $T$}
\end{align}
Therefore, for any \(n \ge M = 1936\,\Sample\), it follows that

\begin{align}
& M \ \widetilde{\mu}_{i,M} \geq \frac{42}{44} \mu^* \ 1936 \Sample=1848\left(c^2\log T+\frac{(\log T)^2}{\mu^*\epsilon^2}\right)\label{orig}
\end{align}
Also we have for some $\lambda>0$
\begin{align}
\frac{\lambda(\log T)^2}{\left(\widehat{\mu}_{i^*,M}-\frac{1}{\epsilon}\frac{8\alpha \log T}{M}\right)\epsilon^2} \leq \frac{44 \lambda(\log T)^2}{41 \mu^*\epsilon^2}\label{partone}
\end{align}
and since $M = 1936S \geq 1936 \frac{(\log T)^2}{(\mu^*)^2 \epsilon^2}$,
\begin{align}
    \frac{\sqrt{8M\alpha\log T}}{\epsilon} = \frac{M \sqrt{8\alpha \log T}}{\epsilon\sqrt M} \leq \frac{M\sqrt{8\alpha\log T}\mu^*\epsilon}{44\epsilon \log T} =  \frac{M\mu^*\sqrt{8\alpha}}{44 \sqrt{\log T}} = 44 \sqrt{\frac{8\alpha}{\log T}}\left(c^2\log T+\frac{(\log T)^2}{\mu^*\epsilon^2}\right) \label{parttwo}
\end{align}
Adding (\ref{partone}) and (\ref{parttwo}) and $\lambda c^2 \log T$, we have
\begin{align}
&\lambda\left(\frac{(\log T)^2}{\max\left(0,\ \widehat{\mu}_{i^*,M}-\frac{1}{\epsilon}\frac{8\alpha \log T}{M}\right)\epsilon^2} + c^2 \log T \right)+ \frac{\sqrt{8M\alpha\log T}}{\epsilon}\nonumber\\
&\leq \lambda\left(\frac{44  (\log T)^2}{41 \mu^*\epsilon^2} + c^2\log T \right)+  44 \sqrt{\frac{8\alpha}{\log T}}\left(c^2\log T+\frac{(\log T)^2}{\mu^*\epsilon^2}\right) \nonumber \\
& \leq \lambda\left(\frac{44  (\log T)^2}{41 \mu^*\epsilon^2} + c^2\log T \right)+  44 \left(c^2\log T+\frac{(\log T)^2}{\mu^*\epsilon^2}\right) \leq 44\left(1+\frac{\lambda}{41}\right)\left(c^2\log T+ \frac{(\log T)^2}{\mu^*\epsilon^2}\right)\label{lambda}
\end{align}
By (\ref{orig}) and (\ref{lambda}), we can say that
\begin{align}
    M\ \widetilde{\mu}_{i^*, M} \geq \lambda\left(c^2\log T+\frac{(\log T)^2}{\max \left(0,\ \widehat{\mu}_{i^*,M}-\frac{1}{\epsilon}\frac{8\alpha \log T}{M}\right)\epsilon^2}\right)+ \frac{\sqrt{8M\alpha\log T}}{\epsilon}
\end{align}
For $\lambda = 1848$, the lemma stands proved.
\end{proof}

\begin{restatable}{lemma}{LemmaTauAnytimeLDP}
\label{tau:anytime_ldp}
Let $\tau$ be the count of uniform‐sampling rounds until any arm's cumulative reward exceeds
\[
  1600\Bigl(c^{2}\log T + \tfrac{(\log T)^{2}}{\max\left(0,\  \widetilde\mu_{i,n} - \tfrac{1}{\epsilon}\sqrt{\tfrac{8\alpha\log T}{n}}\right)\,\epsilon^{2}}\Bigr)
  + \tfrac{\sqrt{8\,n\,\alpha\log T}}{\epsilon}.
\]
Then under $\cE$, we have
\[
  512\,k\,S \le \tau \le 3872\,k\,S.
\]
\end{restatable}

\begin{proof}
Let 
\[
T_1 = 512\,k\,\Sample.
\]
Under the event \(\cE_1\), no arm has been sampled more than \(768\,\Sample\) times in the first \(T_1\) uniform‐sampling rounds. Therefore, for each arm \(i\), Lemma~\ref{tau:lower_ldp} yields
\begin{align*}
n_i\,\widehat\mu_i 
&< 800\left(c^2\log T+\frac{(\log T)^2}{\max\left(0,\ \widetilde{\mu}_{i,n}-\frac{1}{\epsilon}\sqrt{\frac{8\alpha\log T}{n}}\right)\epsilon^2}\right)+\frac{\sqrt{8n\alpha\log T}} {\epsilon}\\
& \leq 1600\left(c^2\log T+\frac{(\log T)^2}{\max\left(0,\ \widetilde{\mu}_{i,n}-\frac{1}{\epsilon}\sqrt{\frac{8\alpha\log T}{n}}\right)\epsilon^2}\right)+\frac{\sqrt{8n\alpha\log T}} {\epsilon}.
\end{align*}
Since \(\tau\) is the first time any arm’s cumulative estimate exceeds \(1600\bigl(c^2\log T + \tfrac{(\log T)^2}{\epsilon}\bigr)\), it follows that
\[
\tau \ge T_1 = 512\,k\,\Sample.
\]

Next, set
\[
T_2 = 3872\,k\,\Sample.
\]
On the event \(\cE\), each arm \(i\) is sampled at least \(1936\,\Sample\) times by round \(T_2\).  Applying Lemma~\ref{tau:upper_ldp} to the optimal arm \(i^*\) gives
\begin{align*}
n_{i^*}\,\widehat\mu_{i^*}
&\ge 1848\left(c^2\log T+\frac{(\log T)^2}{\max\left(0,\ \widetilde{\mu}_{i,n}-\frac{1}{\epsilon}\sqrt{\frac{8\alpha\log T}{n}}\right)\epsilon^2}\right)+\frac{\sqrt{8n\alpha\log T}} {\epsilon} \\&> 1600\left(c^2\log T+\frac{(\log T)^2}{\max\left(0,\ \widetilde{\mu}_{i,n}-\frac{1}{\epsilon}\sqrt{\frac{8\alpha\log T}{n}}\right)\epsilon^2}\right)+\frac{\sqrt{8n\alpha\log T}} {\epsilon}.
\end{align*}
Hence the stopping condition is met by round \(T_2\), so
\[
\tau \le T_2 = 3872\,k\,\Sample.
\]
This completes the proof.

\end{proof}

\begin{restatable}{lemma}{LemmaNCBoptAnytimeLDP}
\label{lem:emp:anytimeldp}
Let $\NCB_{\LDP}(i^{*},t)$ be the Nash confidence bound of the optimal arm $i^{*}$ at round $t$ in Phase~II.
Under $\cE$, for every Phase~II round $t$,
\[
  \NCB_{\LDP}(i^{*},t) \ge \mu^{*}.
\]
\end{restatable}

\begin{proof}

Consider an arbitrary round $t$ in Phase II. Denote by $n_{i^*}$ the number of times the optimal arm $i^*$ has been selected up to (but not including) round $t$, and let $\widehat\mu_{i^*}$ be its corresponding empirical mean at round $t$. By definition, the Nash confidence bound at this round can be written as
\[
{{\NCB}_{LDP}}(i^*,t) =\; \widetilde{\mu}^* 
        \;+\; 2c \sqrt{\frac{2 \,\widetilde{\mu}^*\,\log T}{n_{i^*}}} 
        \;+\; \frac{1}{\epsilon}\sqrt\frac{8\alpha\log T}{n_i^*}
        \;\nonumber+ 4c\frac{(2\alpha)^{\frac{1}{4}}(\log T)^{\frac{3}{4}}}{\sqrt{\epsilon}(n_i^*)^{\frac{3}{4}}}
\]

Since event $\cE$ holds, Lemma \ref{tau:anytime_ldp} guarantees that Algorithm \ref{algo:ldp-ncb} remains in Phase I for at least $128\,k\,\Sample$ rounds before transitioning to Phase II; indeed, the first while-loop terminates only once this many iterations have occurred. Combining this with event $\cE_1$ yields
\[
n_{i^*}\;\ge\;256\,\Sample.
\]
Hence,
\[
n_{i^*}\,\mu^* \;\ge\;256\,c^2\log T,
\]
which in turn allows us to lower‐bound the empirical mean of the optimal arm as follows

%

%
\begin{align}
\widehat{\mu}^* & \geq \mu^* - c \sqrt{\frac{\mu^* \log T}{n_{i^*}}} \tag{since $n_{i^*} \geq 256 \Sample$ and event $\cE_2$ holds}    \\
                        & = \mu^* - c \mu^* \sqrt{\frac{\log T}{\mu^* n_{i^*}}} \nonumber \\
                        & \geq \mu^* - c \mu^* \sqrt{\frac{1}{256c^2}}                  \tag{since  $\mu^* \ n_{i^*} \geq 256c^2\log T$}   \\
                        & = \frac{15}{16} \mu^*. \nonumber
    \end{align}
    Therefore,
    \begin{align*}
{{\NCB}_{LDP}}(i^*,t) & =\; \widetilde{\mu}^* 
        \;+\; 2c \sqrt{\frac{2 \,\widetilde{\mu}^*\,\log T}{n_{i^*}}} 
        \;+\; \frac{1}{\epsilon}\sqrt\frac{8\alpha\log T}{n_i^*}
        \;\nonumber+ 4c\frac{(2\alpha)^{\frac{1}{4}}(\log T)^{\frac{3}{4}}}{\sqrt{\epsilon}(n_i^*)^{\frac{3}{4}}} \\
	& \geq \widehat{\mu}^* + 2c \sqrt{\frac{2\widehat{\mu}^* \log T}{n_{i^*}}} \tag{via event $\cE_4$}\\
        & \geq \widehat{\mu}^* + 2c \sqrt{\frac{\widehat{\mu}^* \log T}{n_{i^*}}} \\
                       & \geq  \mu^* - c \sqrt{\frac{\mu^* \log T}{n_{i^*}}} + 2c \sqrt{\frac{\widehat{\mu}^* \log T}{n_{i^*}}} \tag{due to the event $\cE_{2}$}\\
                       & \geq  \mu^* - c \sqrt{\frac{\mu^* \log T}{n_{i^*}}}  + 2c \sqrt{\frac{15\mu^* \log T}{16n_{i^*}}}     \tag{since $ \widehat{\mu}^*\geq  \frac{15}{16} \mu^*$}   \\
                       & \geq \mu^*
    \end{align*}
    The lemma stands proved.
\end{proof}

\begin{restatable}{lemma}{LemmaBadArmsAnytimeLDP}
\label{lem:bad_arms:anytime_ldp}
Assuming $\cE$, any arm $j$ with $\mu_{j}\le \mu^{*}/256$ is never selected in Phase~II.
\end{restatable}

\begin{proof}
Let $j$ be any arm with true mean satisfying $\mu_j \le \mu^*/256$. Define $r_j$ as the total number of times arm $j$ is selected during Phase I. We first establish that 
\[
r_j \;\ge\;256\,\Sample.
\]
Indeed, since event $\cE$ holds, Lemma \ref{tau:anytime_ldp} implies that Algorithm \ref{algo:ldp-ncb} remains in Phase I for no fewer than $512\,k\,\Sample$ iterations before entering Phase II. Combined with the uniform‐exploration guarantee from event $\cE_1$, this yields $r_j \ge 256\,\Sample$.

Next, under event $\cE_3$ and using $r_j \ge 256\,\Sample$, we deduce that throughout Phase II the empirical mean of arm $j$ obeys
\[
\widehat\mu_j \;\le\;\frac{\mu^*}{128}.
\]

Finally, for any round $t$ in Phase II, let $\NCB_{LDP}(j,t)$ denote the Nash confidence bound for arm $j$. We now show
\[
\NCB_{LDP}(j,t)
\;<\;
\NCB_{LDP}(i^*,t),
\]
which implies that arm $j$ is never selected during Phase II. We have
\begingroup
\allowdisplaybreaks
    \begin{align*}
{{\NCB}_{LDP}}(j,t) &=\; \widetilde{\mu}_j 
        \;+\; 2c \sqrt{\frac{2 \,\widetilde{\mu}_j\,\log T}{r_j}} 
        \;+\; \frac{1}{\epsilon}\sqrt\frac{8\alpha\log T}{r_j}
        \;\nonumber+ 4c\frac{(2\alpha)^{\frac{1}{4}}(\log T)^{\frac{3}{4}}}{\sqrt{\epsilon}(r_j)^{\frac{3}{4}}}\\
                   & \leq \; \widehat{\mu}_j 
        \;+\; 2c \sqrt{\frac{2 \,\widehat{\mu}_j\,\log T}{r_j}} 
        \;+\; \frac{2}{\epsilon}\sqrt\frac{8\alpha\log T}{r_j}
        \;\nonumber+ 8c\frac{(2\alpha)^{\frac{1}{4}}(\log T)^{\frac{3}{4}}}{\sqrt{\epsilon}(r_j)^{\frac{3}{4}}}           \tag{via event $\cE_4$}          \\
                   & \leq \; \frac{\mu^*}{128}
        \;+\; 2c \sqrt{\frac{\,{\mu}^*\,\log T}{64r_j}} 
        \;+\; \frac{2}{\epsilon}\sqrt\frac{8\alpha\log T}{r_j}
        \;\nonumber+ 8c\frac{(2\alpha)^{\frac{1}{4}}(\log T)^{\frac{3}{4}}}{\sqrt{\epsilon}(r_j)^{\frac{3}{4}}}      \tag{since $\widehat{\mu}_j \leq \frac{\mu^*}{128}$}                       \\
                   & \leq \frac{\mu^*}{128} + \frac{c}{4}\sqrt{\frac{\mu^* \log T}{256 S}}+\;+\; \frac{2}{\epsilon}\sqrt\frac{8\alpha\log T}{256S}
        \;\nonumber+ 8c\frac{(2\alpha)^{\frac{1}{4}}(\log T)^{\frac{3}{4}}}{\sqrt{\epsilon}(256S)^{\frac{3}{4}}} \tag{since $r_j \geq 256 S$}\\
                   & \leq \frac{\mu^*}{128} + \frac{\mu^*}{16}+\frac{2}{\epsilon}\sqrt\frac{8\alpha\log T}{256S}
        \;\nonumber+ 8c\frac{(2\alpha)^{\frac{1}{4}}(\log T)^{\frac{3}{4}}}{\sqrt{\epsilon}(256S)^{\frac{3}{4}}}                   \tag{$S>\frac{c^2\log T}{\mu^*}$}                         \\
                   & \leq \frac{\mu^*}{128} + \frac{\mu^*}{64}+\frac{2\mu^*\sqrt\alpha}{\sqrt32} +\frac{c(\mu^*)^\frac{3}{2}(2\alpha)^\frac{1}{4}\epsilon}{8}                         \tag{$S>\frac{(\log T)^2}{(\mu^*)^2\epsilon^2}>\frac{(\log T)}{(\mu^*)^2\epsilon^2}$}                         \\
                   & < \mu^*\left(\frac{3}{128} +\sqrt{\frac{\alpha}{8}} +\frac{(2\alpha)^\frac{1}{4} c \ \epsilon }{8}\right)\leq \mu^* \tag{For $\alpha=3$ and small enough $\epsilon$ }\\
                   & < \NCB_{LDP}(i^*,t) \tag{via Lemma \ref{lem:emp:anytimeldp}}
    \end{align*}
\endgroup

\end{proof}

\begin{restatable}{lemma}{LemmaHighMeanAnytimeLDP}
\label{lem:suboptimal_arms:anytime_ldp}
Under $\cE$, let $T_{i}$ be the number of pulls of arm $i$ by Algorithm~\ref{algo:ldp-ncb}.
For any arm $i$ pulled at least once in Phase~II,
\[
  \mu^{*} \le \mu_{i} + 4c\sqrt{\frac{\mu^{*}\,\log T}{T_{i}-1}}
  + \frac{2}{\epsilon}\sqrt{\frac{8\alpha\log T}{T_{i}-1}}
  + 8c\,\frac{(2\alpha)^{1/4}(\log T)^{3/4}}{\sqrt{\epsilon}\,(T_{i}-1)^{3/4}}.
\]
\end{restatable}
\begin{proof}
Let $i$ be any arm that is chosen at least once during Phase II, and suppose its $T_i$‑th selection in Phase II occurs at some round $t$.  By definition of the selection rule, at round $t$ the arm $i$ must satisfy
\[
\NCB_{LDP}(i,t)\geq\NCB_{LDP}(i^*,t)\;\ge\;\mu^*,
\]
where the final inequality is Lemma \ref{lem:emp:anytimeldp}.  Hence, denoting by $\widetilde\mu_i$ the corrected empirical mean of arm $i$ just before its $T_i$‑th pull, we have

\begin{align} 
\widetilde{\mu}_i
        \;+\; 2c \sqrt{\frac{2 \,\widetilde{\mu}_i\,\log T}{T_{i}-1}} 
        \;+\; \frac{1}{\epsilon}\sqrt\frac{8\alpha\log T}{T_i-1}
        \;+ 4c\frac{(2\alpha)^{\frac{1}{4}}(\log T)^{\frac{3}{4}}}{\sqrt{\epsilon}(T_i-1)^{\frac{3}{4}}}\geq \mu^* \label{ineq:hatmu:anytime_ldp}
\end{align}

Moreover, under event $\cE$, any arm that appears in Phase II has been pulled at least $256\,\Sample$ times in Phase I (by Lemmas \ref{lem:emp:anytimeldp} and \ref{lem:bad_arms:anytime_ldp}), so $T_i>256\,\Sample$.  Since arm $i$ survives into Phase II, Lemma \ref{lem:bad_arms:anytime_ldp} also ensures
\[
\mu_i \;>\;\frac{\mu^*}{256}.
\]

We now turn to an upper bound on the empirical mean $\widehat\mu_i$ in terms of $\mu^*$ to complement \eqref{ineq:hatmu:anytime_ldp}. We have



\begin{align}
\widehat{\mu_i} & \leq  \mu_i + c\sqrt{\frac{\mu_i \log T}{T_i -1}}                \tag{since $\mu_i >\frac{\mu^*}{256}$ and event $\cE_{2}$ holds} \nonumber                                                    \\
& \leq \mu^* +c\sqrt{\frac{\mu^* \log T}{ 256 \Sample}} \tag{since $T_i> 256 \Sample$ and $\mu_i \leq \mu^*$}        \\
& = \mu^* + \frac{\mu^*}{16}  \tag{since $\Sample > \frac{c^2 \log T}{\mu^*}$}\\
& = \frac{17}{16} \mu^*  \label{ineq:hatmuo:anytime_ldp}
\end{align}

    Inequalities (\ref{ineq:hatmu:anytime_ldp}) and (\ref{ineq:hatmuo:anytime_ldp}) give us
    \begin{align*}
        &  \mu^* \leq  \widetilde{\mu}_i
        \;+\; 2c \sqrt{\frac{2 \,\widetilde{\mu}_i\,\log T}{T_{i}-1}} 
        \;+\; \frac{1}{\epsilon}\sqrt\frac{8\alpha\log T}{T_i-1}
        \;\nonumber+ 4c\frac{(2\alpha)^{\frac{1}{4}}(\log T)^{\frac{3}{4}}}{\sqrt{\epsilon}(T_i-1)^{\frac{3}{4}}}\\
         & \leq  \widehat{\mu}_i
        \;+\; 2c \sqrt{\frac{17 \,\mu^*\log T}{8(T_{i}-1)}} 
        \;+\; \frac{2}{\epsilon}\sqrt\frac{8\alpha\log T}{T_i-1}
        \;\nonumber+ 8c\frac{(2\alpha)^{\frac{1}{4}}(\log T)^{\frac{3}{4}}}{\sqrt{\epsilon}(T_i-1)^{\frac{3}{4}}}                      \tag{due to event $\cE_4$}                                         \\
         & \leq \mu_i + c\sqrt{\frac{\mu_i \log T}{T_i -1 }} + 3c\sqrt{\frac{\mu^* \log T}{T_i -1}}+\frac{2}{\epsilon}\sqrt\frac{8\alpha\log T}{T_i-1}
        \;\nonumber+ 8c\frac{(2\alpha)^{\frac{1}{4}}(\log T)^{\frac{3}{4}}}{\sqrt{\epsilon}(T_i-1)^{\frac{3}{4}}}   \tag{via event $\cE_{2}$}                           \\
         & \leq \mu_i + c\sqrt{\frac{\mu^* \log T}{T_i  -1}} + 3c\sqrt{\frac{\mu^* \log T}{T_i -1}} +\frac{2}{\epsilon}\sqrt\frac{8\alpha\log T}{T_i-1}
        \;\nonumber+ 8c\frac{(2\alpha)^{\frac{1}{4}}(\log T)^{\frac{3}{4}}}{\sqrt{\epsilon}(T_i-1)^{\frac{3}{4}}}                   \tag{since $\mu_i \leq \mu^*$} \\
         & \leq \mu_i +  4c\sqrt{\frac{\mu^* \log T}{T_i -1 }}+\frac{2}{\epsilon}\sqrt\frac{8\alpha\log T}{T_i-1}
        \;\nonumber+ 8c\frac{(2\alpha)^{\frac{1}{4}}(\log T)^{\frac{3}{4}}}{\sqrt{\epsilon}(T_i-1)^{\frac{3}{4}}}
    \end{align*}
    This completes the proof of the lemma.
\end{proof}


\begin{restatable}{lemma}{LemmaModifiedNCBLDP}
\label{lem:modified_ncb_ldp}
Suppose the optimal mean $\mu^*$ satisfies
\[
  \mu^{*} \ge \frac{512\sqrt{k\log T}}{\sqrt{T}} + \frac{10000\,\log(256k)(\log T)^2}{\epsilon\,\sqrt{T}},
\]
Then for any $w\le T$,
\begin{multline*}
  \bigl(\prod_{t=1}^{w} \E[\mu_{I_{t}}]\bigr)^{1/T}
  \geq (\mu^*)^{\frac{w}{T}}\bigg(1-4000c\sqrt{\frac{ k \log T }{\mu^*T} }-\frac{4000\log (256k)(\log T)^2}{(\mu^*)^2 \epsilon^2 T} -\frac{4\sqrt{8k\alpha\log T}}{\sqrt T\mu^*\epsilon} -\frac{(2\alpha)^{\frac{1}{4}}(k\log T)^{\frac{3}{4}}}{T^\frac{3}{4}\mu^*\sqrt{\epsilon}}\bigg).
\end{multline*}
\end{restatable}
For the locally differentially private setup, we bound the Nash social welfare of Algorithm \ref{algo:ldp-ncb} using the following Lemma:
\begin{proof}
We begin by deriving a lower bound on the expected reward $\mathbb{E}[\mu_{I_t}]$ of Algorithm~\ref{algo:ldp-ncb}, which will hold uniformly for all rounds $t \leq T$.

In particular, let $p_t$ denote the probability that the algorithm is in Phase~\textcal{1} at round $t$. Then, with probability $p_t$, the algorithm performs uniform exploration, selecting an arm uniformly at random. In this case, the expected reward is at least $\frac{\mu^*}{k}$.

On the other hand, with probability $(1 - p_t)$, the algorithm is in Phase~\textcal{2}. By Lemma~\ref{lem:bad_arms:anytime_ldp}, in this phase the expected reward is guaranteed to be at least $\frac{\mu^*}{256}$. Combining both scenarios, we obtain:\\

\begin{align}
 \E[\mu_{I_t}]&\geq \E \left[\mu_{I_t}|\cE \right] \ \prob\{\cE\} \nonumber \geq  \E \left[\mu_{I_t}|\cE \right] \left(1-\frac{6}{T}\right) \tag{Lemma \ref{lemma:modifiedgoodeventpr}}\\
 &=  \left(1-\frac{6}{T}\right)\left(p_t  \frac{\mu^*}{k} + (1-p_t)  \frac{\mu^*}{256}\right) \geq  \left(1-\frac{6}{T}\right)\left(\frac{\mu^*}{256k}\right) \label{ineq:UniformExpLB_ldp}
\end{align}

To facilitate the subsequent case analysis, set the cutoff 
\[
\overline{T} \;=\; 3872\,k\,\Sample.
\]
By Lemma \ref{tau:anytime_ldp}, the algorithm is guaranteed to have exited Phase I by round~\(\overline{T}\); equivalently, the stopping condition of the first while-loop (Line \ref{step:PhaseOneAlgTwo_ldp}) is met no later than this round.  Furthermore, under the stated lower‐bound assumption on \(\mu^*\) and for \(T\) taken sufficiently large, one also shows that
\begin{align}
&\frac{\overline{T} \ \log \left(256k \right)}{T} =  \frac{3872 \ k \Sample \ \log \left( 256k\right) }{T}\\
& = \frac{3872 \ k c^2 \log T \ \log \left( 256k\right) }{\mu^* T}+\frac{3872\log (256k) (\log T)^2}{(\mu^*)^2 \epsilon^2 T}\\
&\leq \frac{3872 c^2 \log (256k) \sqrt{k \log T}}{512 \sqrt{T}}+\frac{3872\log (256k) (\log T)^2}{(\mu^*)^2 \epsilon^2 T} \tag{$\mu^*\geq \frac{512\sqrt{k\log T}}{\sqrt T}$} \\
&\leq \frac{3872 c^2 \log (256k) \sqrt{k \log T}}{512 \sqrt{T}}+{\frac{3872}{10000}} \tag{$\mu^*\geq \frac{10000\log (256k) (\log T)^2}{\epsilon\sqrt T}$} \\
&\leq \frac{1}{2} +\frac{3872}{10000}\leq 1\label{ineq:overlineTexp_ldp}
\end{align} 

To prove the lemma, we split the argument into two exhaustive scenarios based on the round $w$:
\[
\text{Case 1: } w \le \overline{T}, 
\qquad
\text{Case 2: } w > \overline{T}.
\]

{\it Case 1} ($w \leq \overline T$): Invoking inequality (\ref{ineq:UniformExpLB_ldp}) we have 
\begingroup
\allowdisplaybreaks
\begin{align}
    \left(\prod_{t=1}^{w} \E [\mu_{I_t} ] \right)^\frac{1}{T}
     & \geq \left(1-\frac{6}{T}\right)^{\frac{w}{T}}\left(\frac{\mu^*}{256k}\right)^{\frac{w}{T}}\nonumber                                 \\
     & \geq \left(1-\frac{6}{T}\right) \left(\mu^*\right)^{\frac{w}{T}} \left(\frac{1}{256k}\right)^{\frac{w}{T}} \nonumber 
     = \left(1-\frac{6}{T}\right)\left(\mu^*\right)^{\frac{w}{T}} \left(\frac{1}{2}\right)^{\frac{w \  \log (256k)}{T}} \nonumber      \\
     & \geq \left(1-\frac{6}{T}\right)\left(\mu^*\right)^{\frac{w}{T}} \left(\frac{1}{2}\right)^{\frac{\overline{T} \  \log (256k)}{T}}  \tag{since $w \leq \overline{T}$} \\
     & = \left(1-\frac{6}{T}\right)\left(\mu^*\right)^{\frac{w}{T}} \left(1- \frac{1}{2}\right)^{\frac{\overline{T} \ \log (256k)}{T}} \nonumber   \\
     & \geq  \left(\mu^*\right)^{\frac{w}{T}} \left(1-{\frac{\overline{T}  \log (256k)}{T}}\right)\left(1-\frac{6}{T}\right) \tag{via inequality (\ref{ineq:overlineTexp_ldp}) and Claim \ref{lem:binomial}}\\
   & 	\geq \left(\mu^*\right)^{\frac{w}{T}} \left(1-{\frac{\overline{T}  \log (256k)}{T}}-\frac{6}{T}\right)\nonumber\\ 
   & = \bigg(\mu^*\bigg)^{\frac{w}{T}} \bigg(
       1 - \frac{3872 \, k c^2 \log T \, \log (256k)}{\mu^* T} \nonumber
       - \frac{3872 \log (256k) (\log T)^2}{(\mu^*)^2 \epsilon^2 T}
       - \frac{6}{T}
   \bigg)\\
  & = \left(\mu^*\right)^{\frac{w}{T}} \Bigg(
      1
      - \frac{3872c \sqrt{k \log T}}{\sqrt{\mu^* T}} \cdot \frac{c \log{(256k) (\log T)^2} \sqrt{k \log T}}{\sqrt{\mu^* T}} \nonumber
      - \frac{3872\log (256k) (\log T)^2}{(\mu^*)^2 \epsilon^2 T}
      - \frac{6}{T}
    \Bigg) \\
      & \geq (\mu^*)^{\frac{w}{T}}\left(1-4000c\sqrt{\frac{ k \log T }{\mu^*T} }
      -\frac{4000\log (256k) (\log T)^2}{(\mu^*)^2 \epsilon^2 T}\right)\label{ineq:CaseOneW_ldp} \\
    &\geq (\mu^*)^{\frac{w}{T}}\bigg(1-4000c\sqrt{\frac{ k \log T }{\mu^*T} }-\frac{4000\log (256k)(\log T)^2}{(\mu^*)^2 \epsilon^2 T} -\frac{4\sqrt{8k\alpha\log T}}{\sqrt T\mu^*\epsilon} -\frac{(2\alpha)^{\frac{1}{4}}(k\log T)^{\frac{3}{4}}}{T^\frac{3}{4}\mu^*\sqrt{\epsilon}}\bigg) \nonumber
    \end{align}
\endgroup

Inequality \eqref{ineq:CaseOneW_ldp} immediately follows from the estimate
\[
\frac{c\,\log(256k)\,\sqrt{k\log T}}{\sqrt{\mu^*\,T}}\;\le\;1
\]
for all sufficiently large \(T\), using the fact that
\[
\mu^*\;\ge\;\frac{512\sqrt{k\log T}}{\sqrt{T}}.
\]

{\it Case 2} ($w>\overline{T}$): Now we separate the Nash social welfare into the following terms: 
\begin{align}
    \left(\prod_{t=1}^{w} \E \left[ \mu_{I_t} \right] \right)^\frac{1}{T} =  \left(\prod_{t=1}^{\overline{T}} \E \left[ \mu_{I_t} \right]\right)^\frac{1}{T} \left(\prod_{t=\overline{T} + 1}^{w}  \E \left[ \mu_{I_t} \right] \right)^\frac{1}{T} \label{eqn:splitW_ldp}
\end{align}

In the product decomposition, the first factor captures the total expected reward accumulated over rounds \(t \le \overline{T}\), while the second factor corresponds to rounds \(\overline{T} < t \le w\). We proceed by deriving separate lower bounds for each of these two factors.

The first term in the right-hand side of equation (\ref{eqn:splitW_ldp}) can be bounded as follows
\begingroup
\allowdisplaybreaks
\begin{align}
    \left(\prod_{t=1}^{\overline{T}} \E [\mu_{I_t} ] \right)^\frac{1}{T} 
     & \geq \left(1-\frac{6}{T}\right)^{\frac{\overline{T}}{T}}\left(\frac{\mu^*}{256k}\right)^{\frac{\overline{T}}{T}} \tag{via inequality (\ref{ineq:UniformExpLB_ldp})}                                 \\
     & \geq \left(1-\frac{6}{T}\right) \left(\mu^*\right)^{\frac{\overline{T}}{T}} \left(\frac{1}{256k}\right)^{\frac{\overline{T}}{T}} = \left(1-\frac{6}{T}\right)\left(\mu^*\right)^{\frac{\overline{T}}{T}} \left(\frac{1}{2}\right)^{\frac{\overline{T}  \log (256k)}{T}} \nonumber  \\  
     & = \left(1-\frac{6}{T}\right)\left(\mu^*\right)^{\frac{\overline{T}}{T}} \left(1- \frac{1}{2}\right)^{\frac{\overline{T}  \log (256k)}{T}} \nonumber   \\
     & \geq  \left(\mu^*\right)^{\frac{\overline{T}}{T}} \left(1-{\frac{\overline{T}  \log (256k)}{T}}\right)\left(1-\frac{6}{T}\right)    \label{ineq:phaseone:anytime_ldp}
\end{align}
\endgroup
To justify the final inequality, observe that the exponent
\[
\frac{\overline{T}\,\log(256k)}{T}\;\le\;1
\]
by \eqref{ineq:overlineTexp_ldp}, and then invoke Claim \ref{lem:binomial}.

Next, we bound the second summand on the right‐hand side of \eqref{eqn:splitW_ldp}.
\begin{align}
    \left(\prod_{t=\overline{T}+ 1}^{w} \E\left[ \mu_{I_t} \right]\right)^\frac{1}{T}
     & \geq \E\left[\left( \prod_{t=\overline{T}+1}^{w} \mu_{I_t} \right)^\frac{1}{T}\right ] \tag{Multivariate Jensen's inequality} \nonumber               \\
     & \geq \E \left[\left( \prod_{t=\overline{T}+1}^{w} \mu_{I_t} \right)^\frac{1}{T} \;\middle|\; \cE \right]  \prob\{ \cE \} \label{ineq:interim:anytime_ldp}
\end{align}

By Lemma \ref{tau:anytime_ldp}, Phase I must have concluded by round \(\overline T\), so every \(t>\overline T\) lies in Phase II.  To bound the second term on the right‐hand side of \eqref{ineq:interim:anytime_ldp}, we restrict attention to the arms chosen after round \(\overline T\).  Reindex these arms as \(\{1,2,\dots,\ell\}\), and let \(m_i\ge1\) be the number of times arm \(i\) is pulled after \(\overline T\); note that
\[
\sum_{i=1}^\ell m_i \;=\; w - \overline T.
\]
Let \(T_i\) denote the total pulls of arm \(i\) over the entire algorithm, so that \(T_i - m_i\) is its number of pulls in the first \(\overline T\) rounds.  With this notation,
\[
\mathbb{E}\!\Bigl[\bigl(\prod_{t=\overline T+1}^T \mu_{I_t}\bigr)^{1/T}\Bigm|\cE\Bigr]
\;=\;
\mathbb{E}\!\Bigl[\prod_{i=1}^\ell\mu_i^{\,m_i/T}\Bigm|\cE\Bigr].
\]
Finally, since we condition on the good event \(\cE\), Lemma \ref{lem:suboptimal_arms:anytime_ldp} applies to each arm \(i\in[\ell]\), yielding the desired bound. Thus, we have

\begin{align}
    \E \left[\left( \prod_{t=\overline{T}+1}^{w} \ \mu_{I_t} \right)^\frac{1}{T} \;\middle|\; E \right] & = \E\left[\left( \prod_{i=1}^{\ell} \mu_{i}^\frac{m_i}{T}  \right)\;\middle|\; \cE \right] \nonumber \\                                                                                             
    &  \geq \E\left[\prod_{i=1}^{\ell}\left(\mu^* -  4c\sqrt{\frac{\mu^* \log T}{T_i -1 }}-\frac{2}{\epsilon}\sqrt\frac{8\alpha\log T}{T_i-1}
        \;\nonumber - 8c\frac{(2\alpha)^{\frac{1}{4}}(\log T)^{\frac{3}{4}}}{\sqrt{\epsilon}(T_i-1)^{\frac{3}{4}}} \right)^\frac{m_i}{T} \;\middle|\; \cE \right]   \tag{Lemma \ref{lem:suboptimal_arms:anytime_ldp}} \nonumber                                                                                                      \\
   & = (\mu^*)^{\frac{w-\overline{T}}{T}} \E\left[\prod_{i=1}^{\ell}\left(1 - 4c\sqrt{\frac{\log T}{\mu^*(T_i -1) }}-\frac{2}{\mu^*\epsilon}\sqrt\frac{8\alpha\log T}{T_i-1}
        \; - \frac{8c}{\mu^*}\frac{(2\alpha)^{\frac{1}{4}}(\log T)^{\frac{3}{4}}}{\sqrt{\epsilon}(T_i-1)^{\frac{3}{4}}} \right)^\frac{m_i}{T}	\;\middle|\; \cE \right] \label{ineq:dunzo:anytime_ldp}
\end{align}

In the equality above we have used the identity $\sum_{i=1}^\ell m_i = w - \overline{T}$.  Moreover, under the good event $\cE$, each arm is guaranteed to be sampled at least $64\,\Sample$ times within the first $\overline T$ rounds.  Consequently, for every $i\in[\ell]$,
\[
T_i \;=\;(T_i - m_i) + m_i
\;>\;256\,\Sample + 1
\;\ge\;256\,\Sample,
\]
and hence the lower‐bound conditions of Lemma \ref{lem:suboptimal_arms:anytime_ldp} apply to each such arm. Using this, we get
\begin{align*}
& 4c\sqrt{\frac{\log T}{\mu^*(T_i -1) }}+\frac{2}{\mu^*\epsilon}\sqrt\frac{8\alpha\log T}{T_i-1}
        \;\nonumber + \frac{8c}{\mu^*}\frac{(2\alpha)^{\frac{1}{4}}(\log T)^{\frac{3}{4}}}{\sqrt{\epsilon}(T_i-1)^{\frac{3}{4}}} \leq 4c\sqrt{\frac{\log T}{256c^2\log T}}+\frac{2}{\mu^*\epsilon}\sqrt\frac{8\alpha\log T}{T_i-1}
        \;\nonumber + \frac{8c}{\mu^*}\frac{(2\alpha)^{\frac{1}{4}}(\log T)^{\frac{3}{4}}}{\sqrt{\epsilon}(T_i-1)^{\frac{3}{4}}} &\tag{$ T_i > 256S > 256\frac{c^2\log T}{\mu^*}$} \\
&\qquad\qquad\qquad \leq \frac{1}{4}+\frac{2\sqrt{8\alpha \log T}}{\sqrt{256 (\log T)^2}} + \frac{8\sqrt{\mu^*}\epsilon(2\alpha)^{\frac{1}{4}}(\log T)^{\frac{3}{4}}}{(256(\log T)^2)^{\frac{3}{4}}}&\tag{$T_i > 256S > 256\frac{(\log T)^2}{(\mu^*)^2\epsilon^2}$} \\
&\qquad\qquad\qquad = \frac{1}{4}+\frac{\sqrt{8\alpha}}{8\sqrt{\log T}} +\frac{\sqrt{\mu^*}{(2\alpha)^\frac{1}{4}}\epsilon}{8(\log T)^\frac{3}{2}} \leq \frac{1}{2} &\tag{For large enough T}
\end{align*}

Since the above bound holds for every $i \in [\ell]$, applying Claim~\ref{lem:binomial} allows us to simplify the expectation in inequality~\eqref{ineq:dunzo:anytime_ldp} as follows:
\begin{align*}
    &\E\left[\prod_{i=1}^{\ell}\left(1 - 4c\sqrt{\frac{\log T}{\mu^*(T_i -1) }}-\frac{2}{\mu^*\epsilon}\sqrt\frac{8\alpha\log T}{T_i-1}
        \;\nonumber - \frac{8c}{\mu^*}\frac{(2\alpha)^{\frac{1}{4}}(\log T)^{\frac{3}{4}}}{\sqrt{\epsilon}(T_i-1)^{\frac{3}{4}}} \right)^\frac{m_i}{T}	\;\middle|\; \cE \right] \\
        & \geq \E \left[\prod_{i=1}^{\ell}\left(1 - \frac{8cm_i}{T}\sqrt{\frac{\log T}{\mu^*(T_i -1) }}-\frac{4m_i}{T\mu^*\epsilon}\sqrt\frac{8\alpha\log T}{T_i-1}
        \;\nonumber - \frac{8c\ m_i}{T\mu^*}\frac{(2\alpha)^{\frac{1}{4}}(\log T)^{\frac{3}{4}}}{\sqrt{\epsilon}(T_i-1)^{\frac{3}{4}}} \right)\;\middle|\; \cE \right]                               \\
 & \geq \E \left[\prod_{i=1}^{\ell}\left(1 - \frac{8c}{T}\sqrt{\frac{m_i\log T}{\mu^*}}-\frac{4\sqrt{8m_i\alpha\log T}}{T\mu^*\epsilon}
        \;\nonumber - \frac{8c(m_i)^\frac{1}{4}(2\alpha)^{\frac{1}{4}}(\log T)^{\frac{3}{4}}}{T\mu^*\sqrt{\epsilon}} \right)\;\middle|\; \cE \right]             & \text{(since $T_i \geq m_i + 1$)}
\end{align*}
In particular, because for any $x,y \ge 0$ we have 
$
(1 - x)(1 - y) \;\ge\; 1 - x - y,
$
we can replace the multiplicative term in the preceding bound by its corresponding linear lower bound. 

\begingroup
\allowdisplaybreaks
\begin{align*}
    &\E \left[\prod_{i=1}^{\ell}\left(1 - \frac{8c}{T}\sqrt{\frac{m_i\log T}{\mu^* }}-\frac{4\sqrt{8m_i\alpha\log T}}{T\mu^*\epsilon}
        \;\nonumber - \frac{16c(m_i)^\frac{1}{4}(2\alpha)^{\frac{1}{4}}(\log T)^{\frac{3}{4}}}{T\mu^*\sqrt{\epsilon}} \right)\;\middle|\; \cE \right] \\ 
        & \geq \E\left[1 -\sum_{i=1}^{\ell}\left( \frac{8c}{T}\sqrt{\frac{m_i\log T}{\mu^*}} + \frac{4\sqrt{8m_i\alpha\log T}}{T\mu^*\epsilon}
        \;\nonumber + \frac{16c(m_i)^\frac{1}{4}(2\alpha)^{\frac{1}{4}}(\log T)^{\frac{3}{4}}}{T\mu^*\sqrt{\epsilon}} \right)\;\middle|\; \cE \right]\\
     & = 1 -\left(\frac{8c }{T}\sqrt{\frac{ \log T }{\mu^*}} \right) \E\left[ \sum_{i=1}^{\ell} \sqrt{m_i}\;\middle|\; \cE \right]  - \frac{4\sqrt{8\alpha\log T}}{T\mu^*\epsilon}\E\left[\sum_{i=1}^{\ell}\sqrt{m_i}\middle|\; \cE\right] -\frac{(2\alpha)^{\frac{1}{4}}(\log T)^{\frac{3}{4}}}{T\mu^*\sqrt{\epsilon}}\E\left[\sum_{i=1}^{\ell}m_i^\frac{1}{4}\middle|\; \cE\right]      \\
     & \geq 1 -\left(\frac{8c }{T}\sqrt{\frac{ \log T }{\mu^*}} \right) \E\left[\sqrt{\ell}\sqrt{ \sum_{i=1}^{\ell} m_i}\;\middle|\; \cE \right]  - \frac{4\sqrt{8\alpha\log T}}{T\mu^*\epsilon}\E\left[\sqrt{\ell}\sqrt{ \sum_{i=1}^{\ell} m_i}\;\middle|\; \cE \right] -\frac{(2\alpha)^{\frac{1}{4}}(\log T)^{\frac{3}{4}}}{T\mu^*\sqrt{\epsilon}}\E\left[\ell^\frac{3}{4} \left(\sum_{i=1}^{\ell} m_i\right)^\frac{1}{4}\;\middle|\; \cE \right]\tag{Holder's inequality} \\
     & \geq 1 -\left(\frac{8c }{T}\sqrt{\frac{ \log T }{\mu^*}} \right) \E\left[\sqrt{\ell \ T}\;\middle|\; \cE \right]  - \frac{4\sqrt{8\alpha\log T}}{T\mu^*\epsilon}\E\left[\sqrt{\ell \ T}\;\middle|\; \cE \right] -\frac{(2\alpha)^{\frac{1}{4}}(\log T)^{\frac{3}{4}}}{T\mu^*\sqrt{\epsilon}}\E\left[\ell^\frac{3}{4} T^\frac{1}{4}\;\middle|\; \cE \right]\tag{since $\sum_i m_i \leq T$}                         \\
     & = 1 -\left({8c }\sqrt{\frac{ \log T }{\mu^*T}} \right) \E\left[\sqrt{\ell}\;\middle|\; \cE \right]  - \frac{4\sqrt{8\alpha\log T}}{\sqrt T\mu^*\epsilon}\E\left[\sqrt{\ell}\;\middle|\; \cE \right] -\frac{(2\alpha)^{\frac{1}{4}}(\log T)^{\frac{3}{4}}}{T^\frac{3}{4}\mu^*\sqrt{\epsilon}}\E\left[\ell^\frac{3}{4} \;\middle|\; \cE \right]                                                                     \\
     & \geq 1 -\left({8c }\sqrt{\frac{ k\log T }{\mu^*T}} \right)  - \frac{4\sqrt{8k\alpha\log T}}{\sqrt T\mu^*\epsilon} -\frac{(2\alpha)^{\frac{1}{4}}(k\log T)^{\frac{3}{4}}}{T^\frac{3}{4}\mu^*\sqrt{\epsilon}}   \tag{since $\ell \leq k$}
\end{align*}
\endgroup

By substituting this estimate into inequalities~\eqref{ineq:interim:anytime_ldp} and~\eqref{ineq:dunzo:anytime_ldp}, we arrive at
\begin{align}
    \left(\prod_{t=\overline{T} + 1}^{w} \E\left[ \mu_{I_t} \right]\right)^\frac{1}{T} \geq (\mu^*)^{\frac{w-\overline{T}}{T}} \bigg ( 1 -{8c }\sqrt{\frac{ k\log T }{\mu^*T}} - \frac{4\sqrt{8k\alpha\log T}}{\sqrt T\mu^*\epsilon} -\frac{(2\alpha)^{\frac{1}{4}}(k\log T)^{\frac{3}{4}}}{T^\frac{3}{4}\mu^*\sqrt{\epsilon}}\bigg) \prob \{\cE\} \label{ineq:toomany:anytime_ldp}
\end{align}
Utilizing the bounds from inequalities~\eqref{ineq:toomany:anytime_ldp} and~\eqref{ineq:phaseone:anytime_ldp} for the two terms in equation~\eqref{eqn:splitW_ldp}, we obtain the following lower bound on the algorithm’s Nash social welfare:
\begin{align*}
    &\left(\prod_{t=1}^{w} \E \left[ \mu_{I_t} \right] \right)^\frac{1}{T}
     \geq (\mu^*)^{\frac{w}{T}}\bigg(1-{\frac{\overline{T}\cdot \log (256k)}{T}}-\frac{6}{T}\bigg) \bigg( 1 -{8c }\sqrt{\frac{ k\log T }{\mu^*T}}   - \frac{4\sqrt{8k\alpha\log T}}{\sqrt T\mu^*\epsilon} -\frac{(2\alpha)^{\frac{1}{4}}(k\log T)^{\frac{3}{4}}}{T^\frac{3}{4}\mu^*\sqrt{\epsilon}}\bigg) \prob\{ \cE \}                 \\
      & \geq (\mu^*)^{\frac{w}{T}} \left(1 - \frac{\overline{T} \cdot \log (256k)}{T} - \frac{6}{T} \right) \left(1 -{8c }\sqrt{\frac{ k\log T }{\mu^*T}}   - \frac{4\sqrt{8k\alpha\log T}}{\sqrt T\mu^*\epsilon} -\frac{(2\alpha)^{\frac{1}{4}}(k\log T)^{\frac{3}{4}}}{T^\frac{3}{4}\mu^*\sqrt{\epsilon}} \right) \left(1 - \frac{6}{T} \right) \tag{via Lemma \ref{lemma:modifiedgoodeventpr}}\\
     & \geq (\mu^*)^{\frac{w}{T}} \bigg(1-{\frac{\overline{T}\cdot \log (256k)}{T}}-{8c }\sqrt{\frac{ k\log T }{\mu^*T}}   - \frac{4\sqrt{8k\alpha\log T}}{\sqrt T\mu^*\epsilon} -\frac{(2\alpha)^{\frac{1}{4}}(k\log T)^{\frac{3}{4}}}{T^\frac{3}{4}\mu^*\sqrt{\epsilon}} -\frac{12}{T}\bigg)   \\                                              
     &= (\mu^*)^{\frac{w}{T}} \bigg(1 - \frac{3872kc^2 \log T \log (256k)}{\mu^* T} - \frac{3872 \log (256k) (\log T)^2}{(\mu^*)^2 \epsilon^2 T}  -{8c }\sqrt{\frac{ k\log T }{\mu^*T}}  \\
     &\hspace{28em}- \frac{4\sqrt{8k\alpha\log T}}{\sqrt T\mu^*\epsilon} -\frac{(2\alpha)^{\frac{1}{4}}(k\log T)^{\frac{3}{4}}}{T^\frac{3}{4}\mu^*\sqrt{\epsilon}} - \frac{12}{T} \bigg)\\
    & \geq (\mu^*)^{\frac{w}{T}}\bigg(1-4000c\sqrt{\frac{ k \log T }{\mu^*T} }-\frac{4000\log (256k)(\log T)^2}{(\mu^*)^2 \epsilon^2 T} -\frac{4\sqrt{8k\alpha\log T}}{\sqrt T\mu^*\epsilon} -\frac{(2\alpha)^{\frac{1}{4}}(k\log T)^{\frac{3}{4}}}{T^\frac{3}{4}\mu^*\sqrt{\epsilon}}\bigg)
    \end{align*}
The last inequality is derived in the same manner as the final step of the proof of inequality~\eqref{ineq:CaseOneW_ldp}, thereby completing the argument and establishing the lemma.
\end{proof}

Based on the above lemmas, we are now in a position to prove Theorem \ref{theorem:improvedNashRegretLDP}

\begin{proof}[Proof of Theorem \ref{theorem:improvedNashRegretLDP}]
\hfill\\
 For $\mu^* = O\left(\sqrt {\frac{k\log T}{T}}+\frac{(\log T)^2 \log k}{\sqrt T\epsilon}\right)$, the theorem holds trivially. We thus, consider the case when $\mu^*=\Omega\left(\sqrt\frac{k \log T}{T}+\frac{(\log T)^2\log k}{\sqrt T \epsilon}\right)$
 The stated Nash regret guarantee follows directly by applying Lemma \ref{lem:modified_ncb_ldp} with $w = T$. Thus we have, 
 \begin{align*}
     \NRg_T  &\leq \mu^* - (\mu^*)^{\frac{T}{T}}\bigg(1-4000c\sqrt{\frac{ k \log T }{\mu^*T} }-\frac{4000\log (256k)(\log T)^2}{(\mu^*)^2 \epsilon^2 T} -\frac{4\sqrt{8k\alpha\log T}}{\sqrt T\mu^*\epsilon} -\frac{(2\alpha)^{\frac{1}{4}}(k\log T)^{\frac{3}{4}}}{T^\frac{3}{4}\mu^*\sqrt{\epsilon}}\bigg)\\ &\leq(\mu^*)\bigg(4000c\sqrt{\frac{ k \log T }{\mu^*T} }+\frac{4000\log (256k)(\log T)^2}{(\mu^*)^2 \epsilon^2 T} +\frac{4\sqrt{8k\alpha\log T}}{\sqrt T\mu^*\epsilon} +\frac{(2\alpha)^{\frac{1}{4}}(k\log T)^{\frac{3}{4}}}{T^\frac{3}{4}\mu^*\sqrt{\epsilon}}\bigg)\\
     &=\bigg(4000c\sqrt{\frac{ \mu^*k \log T }{T} }+\frac{4000\log (256k)(\log T)^2}{(\mu^*) \epsilon^2 T} +\frac{4\sqrt{8k\alpha\log T}}{\sqrt T\epsilon} +\frac{(2\alpha)^{\frac{1}{4}}(k\log T)^{\frac{3}{4}}}{T^\frac{3}{4}\sqrt{\epsilon}}\bigg)\\
     &\leq\bigg(4000c\sqrt{\frac{ \mu^*k \log T }{T} }+\frac{4000\log (256k)(\log T)^2}{\epsilon \sqrt T} +\frac{4\sqrt{8k\alpha\log T}}{\sqrt T\epsilon} +\frac{(2\alpha)^{\frac{1}{4}}(k\log T)^{\frac{3}{4}}}{T^\frac{3}{4}\sqrt{\epsilon}}\bigg)\tag{$\mu^*\geq\frac{1}{\sqrt T\epsilon}$}\\
    & = O\bigg(\sqrt{\frac {k\log T}{T}}+\frac{\sqrt k(\log T)^2}{\sqrt T\epsilon }\bigg) & \tag{$\mu^* \leq1$}
 \end{align*}
 This completes the proof of the theorem. Similar to Algorithm \ref{algo:gdp-ncb}, the result holds for clipped private means as well.

\end{proof}


\subsection{Anytime Algorithms}
We will first prove Theorem \ref{thm:anytime-nucb}. To do this, we establish the following lemma:
\begin{restatable}{lemma}{LemmaDPNCBAnytime}
\label{conditional:NCB:anytime}
In any multi-armed bandit instance where the optimal mean satisfies
\[
\mu^* \;\geq\; \frac{512\sqrt{k\log T}}{\sqrt{T}} + \frac{10000\,\log(256k)\,(\log T)^2}{\epsilon T},
\]
the following inequality holds for every epoch \(h \geq h^*\) and all rounds 
\[
r \in \{R_h + 1, R_h + 2, \ldots, R_{h+1}\}:
\]
    \begin{align*}
        \left(\prod_{t= R_h + 1}^{r}\E\left[\mu_{I_t}\right]\right)^\frac{1}{T} & \geq (\mu^*)^\frac{r-R_h}{T} \bigg(1-4001c\sqrt{\frac{ k \log T }{\mu^*T} }\nonumber-\frac{4001\log (256k)(\log T)^2}{\mu^* \epsilon T} -\frac{4k}{T}\frac{\alpha (\log T)^2}{\mu^*\epsilon } -\frac{16k}{T}\sqrt{\frac{2\alpha}{\epsilon}}\frac{(\log T)^\frac{3}{2}}{\mu^*}\bigg).
    \end{align*}    
\end{restatable}

\begin{proof}
Consider any epoch \(h \ge h^*\), and let \(F_h\) be the event that Algorithm~\ref{algo:ncb:anytime} invokes the GDP-NCB procedure (Algorithm~\ref{algo:gdp-ncb}) during epoch \(h\) (see Line~\ref{step:callsubroutine}).  Note that
\[
\prob(F_h) \;=\; 1 - \frac{1}{W_h^2}.
\]
Furthermore, by the definition of \(h^*\) and Claim~\ref{claim:we}, the epoch length satisfies
\[
\sqrt{T} \;\le\; W_h \;\le\; T.
\]
Therefore, applying Lemma~\ref{lem:modified_ncb} conditioned on \(F_h\) yields:
\begingroup
\allowdisplaybreaks
    \begin{align*}
    &\left(\prod_{t= R_h+1}^{r}\E\left[\mu_{I_t}\right]\right)^\frac{1}{T} 
    \geq \left(\prod_{t= R_h+1}^{r}\E\left[\mu_{I_t} | F_h \right] \ \prob \{ F_h \} \right) ^\frac{1}{T}
    \\
    &\geq \left(\prod_{t= R_h+1}^{r}\E\left[\mu_{I_t} | F_h \right] \left( 1- \frac{1}{W_h^2}\right) \right) ^\frac{1}{T} \geq \left( 1- \frac{1}{W_h^2}\right)  \left(\prod_{t= R_h+1}^{r}\E\left[\mu_{I_t} | F_h \right]  \right) ^\frac{1}{T}\\
    &\geq (\mu^*)^\frac{r-R_h}{T}\left( 1- \frac{1}{W_h^2}\right) \bigg(1-4000c\sqrt{\frac{ k \log T }{\mu^*T} }-\frac{4000\log (256k)(\log T)^2}{\mu^* \epsilon T}-\frac{4k}{T}\frac{\alpha (\log T)^2}{\mu^*\epsilon } -\frac{16k}{T}\sqrt{\frac{2\alpha}{\epsilon}}\frac{(\log T)^\frac{3}{2}}{\mu^*}\bigg)  \tag{Lemma \ref{lem:modified_ncb}}\\
    &\geq (\mu^*)^\frac{r-R_h}{T}\left( 1- \frac{1}{T}\right)\bigg(1-4000c\sqrt{\frac{ k \log T }{\mu^*T} }-\frac{4000\log (256k)(\log T)^2}{\mu^* \epsilon T}-\frac{4k}{T}\frac{\alpha (\log T)^2}{\mu^*\epsilon } -\frac{16k}{T}\sqrt{\frac{2\alpha}{\epsilon}}\frac{(\log T)^\frac{3}{2}}{\mu^*}\bigg) \tag{since $W_h \geq \sqrt{T}$}\\
   &\geq (\mu^*)^\frac{r-R_h}{T}\bigg(1-4000c\sqrt{\frac{ k \log T }{\mu^*T} }-\frac{4000\log (256k)(\log T)^2}{\mu^* \epsilon T}-\frac{4k}{T}\frac{\alpha (\log T)^2}{\mu^*\epsilon } -\frac{16k}{T}\sqrt{\frac{2\alpha}{\epsilon}}\frac{(\log T)^\frac{3}{2}}{\mu^*}-\frac{1}{T}\bigg)\\ 
   &\geq (\mu^*)^\frac{r-R_h}{T}\bigg(1-4001c\sqrt{\frac{ k \log T }{\mu^*T} }-\frac{4001\log (256k)(\log T)^2}{\mu^* \epsilon T}-\frac{4k}{T}\frac{\alpha (\log T)^2}{\mu^*\epsilon } -\frac{16k}{T}\sqrt{\frac{2\alpha}{\epsilon}}\frac{(\log T)^\frac{3}{2}}{\mu^*}\bigg).  	
    \end{align*}
    
\endgroup
The lemma stands proved 
\end{proof}

\textbf{Proof of Theorem \ref{thm:anytime-nucb}}\\
Throughout our analysis, let \(h\) index the epochs of Algorithm~\ref{algo:ncb:anytime}\,, and denote by \(W_h\) the length of the \(h\)‑th epoch (so that \(h = \lfloor\log_2 W_h\rfloor + 1\)).  Let \(e\) be the total number of epochs over the \(T\) rounds, noting that \(e \le \lfloor\log_2 T\rfloor + 1\).  Define
\[
R_h \;=\;\sum_{z=1}^{h-1} W_z
\]
as the cumulative count of rounds preceding the start of epoch \(h\) (hence \(R_1=0\) and epoch \(h\) begins at round \(R_h+1\)).  Finally, let \(h^*\) be the smallest epoch index for which \(W_{h^*}\ge\sqrt{T}\).  





We begin by relating epoch indices, window lengths, and cumulative round counts with the following three claims.

\begin{claim}\label{lem:window}
For every epoch \(h\), the window size satisfies
\[
W_h \;=\; R_h + 1.
\]
\end{claim}
\begin{proof}
By construction, the \(z\)‑th epoch has length \(W_z = 2^{z-1}\).  Hence
\[
R_h \;=\;\sum_{z=1}^{h-1}W_z
=\sum_{z=1}^{h-1}2^{z-1}
=2^{h-1}-1
=W_h - 1,
\]
whence \(W_h = R_h + 1\).
\end{proof}

\begin{claim}\label{claim:we}
The window length in the final epoch \(e\) obeys
\[
W_e \;\le\; T.
\]
\end{claim}
\begin{proof}
Since epoch \(e\) begins at round \(R_e+1\le T\) and ends after \(W_e\) rounds, it must hold that \(W_e \le T\).  Equivalently, by Claim~\ref{lem:window}, \(W_e = R_e+1 \le T\).
\end{proof}

\begin{claim}\label{claim3:anytime}
Let \(h^*\) be the first epoch with \(W_{h^*}\ge\sqrt{T}\).  Then
\[
R_{h^*} \;<\; 2\,\sqrt{T}.
\]
\end{claim}
\begin{proof}
By minimality of \(h^*\), we have \(W_{h^*-1}<\sqrt{T}\).  Using Claim~\ref{lem:window},
\[
R_{h^*}
=R_{h^*-1}+W_{h^*-1}
=2\,W_{h^*-1}-1
<2\,\sqrt{T}.
\]
\end{proof}

Finally, to prove Theorem \ref{thm:anytime-nucb}, restrict attention to instances satisfying
\[
\mu^* \;\ge\; \frac{512\sqrt{k\log T}}{\sqrt T}
\;+\;\frac{10000\,\log(256k)\,(\log T)^2}{\epsilon\,T},
\]
and assume \(\sqrt T \le W \le T\).  Let \(h^*\) be the first epoch with \(W_{h^*}\ge\sqrt T\), and split the horizon into the first \(R_{h^*}\) rounds and the remaining \(T - R_{h^*}\) rounds.

On any epoch \(g\le h^*\), we have \(W_g<\sqrt T\), so Algorithm \ref{algo:ncb:anytime} invokes uniform sampling with probability
\[
\frac{1}{W_g^2}\;\ge\;\frac{1}{T}.
\]
Hence, for every round \(t\le R_{h^*}\),
\[
p_t \;=\;\prob\{\text{Phase I at }t\}
\;\ge\;\frac{1}{T}.
\]
Therefore, for all rounds \(t\le R_{h^*}\), $$\E \left[\mu_{I_t} \right] \geq \frac{\mu^*}{k} \frac{1}{T}$$ The above bound yields
    \begin{align}
        &\left(\prod_{t=1}^{R_{h^*}} \E\left[\mu_{I_t}\right]\right)^{\frac{1}{T}} \geq \left( \frac{\mu^*} {kT}\right)^\frac{R_{h^*}}{T}\nonumber = (\mu^*)^{\frac{R_{h^*}}{T}} \left( \frac{1}{2}\right)^{\frac{R_{h^*}\log{(kT)}}{T}}\nonumber \tag{via Claim \ref{lem:binomial}}\\
&\geq (\mu^*)^{\frac{R_{h^*}}{T}}  \left( 1 - \frac{R_{h^*}\log{(kT)}}{T} \right)  \geq   (\mu^*)^{\frac{R_{h^*}}{T}}  \left( 1 - \frac{2\log{(kT)}}{\sqrt{T}} \right) \label{ineq:anytimeFirstPart}
    \end{align}
Here, the final inequality is justified by Claim~\ref{claim3:anytime}.

To handle the remaining \(T - R_{h^*}\) rounds, we conduct an epoch-wise analysis and apply Lemma~\ref{conditional:NCB:anytime}. Specifically,

\allowdisplaybreaks
    \begin{align}
        &\left(\prod_{t=R_{h^*}+1}^{T} \E\left[\mu_{I_t}\right]\right)^{\frac{1}{T}}
        =  \left(\prod_{h=h^*}^{e-1} \ \prod_{t= R_h + 1}^{R_{(h+1)}}\E\left[\mu_{I_t}\right]\right)^\frac{1}{T} \nonumber \cdot    \left(\prod_{t= R_{e}+1}^{T}\E\left[\mu_{I_t}\right]\right)^\frac{1}{T} \nonumber  =  \prod_{h=h^*}^{e-1} \left( \prod_{t= R_h+1}^{R_h+W_h}\E\left[\mu_{I_t}\right]\right)^\frac{1}{T} \cdot   \left(\prod_{t= R_{e}+1}^{T}\E\left[\mu_{I_t}\right]\right)^\frac{1}{T} \nonumber \\
        &\geq \prod_{h=h^*}^{e-1} (\mu^*)^{\frac{W_h}{T}}\bigg(1-4001c\sqrt{\frac{ k \log T }{\mu^*T} }\nonumber -\frac{4001\log (256k)(\log T)^2}{\mu^* \epsilon T}\nonumber -\frac{4k}{T}\frac{\alpha (\log T)^2}{\mu^*\epsilon } -\frac{16k}{T}\sqrt{\frac{2\alpha}{\epsilon}}\frac{(\log T)^\frac{3}{2}}{\mu^*}\bigg)\nonumber\\
        & \qquad \qquad \cdot (\mu^*)^{\frac{T-R_e}{T}} \bigg(1-4001c\sqrt{\frac{ k \log T }{\mu^*T} }\nonumber - \frac{4001\log (256k)(\log T)^2}{\mu^* \epsilon T}\nonumber -\frac{4k}{T}\frac{\alpha (\log T)^2}{\mu^*\epsilon } -\frac{16k}{T}\sqrt{\frac{2\alpha}{\epsilon}}\frac{(\log T)^\frac{3}{2}}{\mu^*}\bigg) \tag{via Lemma \ref{conditional:NCB:anytime}} \nonumber\\
        &\geq  (\mu^*)^{1 - \frac{R_{h^*}}{T}} \prod_{j=1}^{e} \bigg(1-4001c\sqrt{\frac{ k \log T }{\mu^*T} }\nonumber - \frac{4001\log (256k)(\log T)^2}{\mu^* \epsilon T}\nonumber -\frac{4k}{T}\frac{\alpha (\log T)^2}{\mu^*\epsilon } -\frac{16k}{T}\sqrt{\frac{2\alpha}{\epsilon}}\frac{(\log T)^\frac{3}{2}}{\mu^*}\bigg) \nonumber \\ 
        &\geq (\mu^*)^{1 - \frac{R_{h^*}}{T}}  \bigg(1-4001c\sqrt{\frac{ k \log T }{\mu^*T} }\nonumber -\frac{4001\log (256k)(\log T)^2}{\mu^* \epsilon T}\nonumber-\frac{4k}{T}\frac{\alpha (\log T)^2}{\mu^*\epsilon } -\frac{16k}{T}\sqrt{\frac{2\alpha}{\epsilon}}\frac{(\log T)^\frac{3}{2}}{\mu^*}\bigg) ^ {\log (2T)} \tag{since $e\leq \log T+1$}\nonumber\\
        &\geq   (\mu^*)^{1 - \frac{R_{h^*}}{T}}  \bigg(1-4001c\log 2T\sqrt{\frac{ k \log T }{\mu^*T} } -\frac{4001\log (256k)(\log T)^2\log 2T}{\mu^* \epsilon T} -\frac{4k}{T}\frac{\alpha (\log T)^2\log 2T}{\mu^*\epsilon }\nonumber\\ 
        & \hspace{28em}-\frac{16k}{T}\sqrt{\frac{2\alpha}{\epsilon}}\frac{(\log T)^\frac{3}{2}\log 2T}{\mu^*}\bigg)\label{ineq:anytimeSecond}
    \end{align}
By applying the elementary inequality
\[
(1 - x)(1 - y) \;\ge\; 1 \;-\; x \;-\; y
\quad\text{for all }x,y \ge 0,
\]
we immediately recover the last inequality.  Next, we substitute the bound from inequality~\eqref{ineq:anytimeFirstPart}, which governs the algorithm’s performance up to round $R_{h^*}$, together with the bound from inequality~\eqref{ineq:anytimeSecond}, which covers the remaining rounds $R_{h^*}+1$ through $T$.  Merging these two estimates yields a single, unified lower bound on the Nash social welfare achieved by Algorithm~\ref{algo:ncb:anytime}, when GDP-NCB is executed inside the epochs:  
 
    \begin{align*}
        &\left(\prod_{t=1}^{T} \E\left[\mu_{I_t}\right] \right)^{\frac{1}{T}}  = \left(\prod_{t=1}^{R_{h^*}} \E\left[\mu_{I_t}\right]\right)^{\frac{1}{T}} \left(\prod_{t=R_{h^*}+1}^{T} \E\left[\mu_{I_t}\right]\right)^{\frac{1}{T}} \\
        &\geq \mu^*  \left( 1 - \frac{2\log{(kT)}}{\sqrt{T}} \right) \bigg(1-4001c\log 2T\sqrt{\frac{ k \log T }{\mu^*T} }\\&-\frac{4001\log (256k)(\log T)^2\log 2T}{\mu^* \epsilon T}-\frac{4k}{T}\frac{\alpha (\log T)^2\log 2T}{\mu^*\epsilon } -\frac{16k}{T}\sqrt{\frac{2\alpha}{\epsilon}}\frac{(\log T)^\frac{3}{2}\log 2T}{\mu^*}\bigg) \\ 
        &\geq \mu^* \bigg(1-\frac{2\log (kT)}{\sqrt T}-4001c\log 2T\sqrt{\frac{ k \log T }{\mu^*T} }-\frac{4001\log (256k)(\log T)^2\log 2T}{\mu^* \epsilon T}-\frac{4k}{T}\frac{\alpha (\log T)^2\log 2T}{\mu^*\epsilon } \\
        &\qquad\qquad-\frac{16k}{T}\sqrt{\frac{2\alpha}{\epsilon}}\frac{(\log T)^\frac{3}{2}\log 2T}{\mu^*}\bigg). \\ 
    \end{align*}
Thus, Algorithm~\ref{algo:ncb:anytime} achieves the following upper bound on its Nash regret, when executing GDP-NCB:
\begin{align*}
\NRg_T &= \mu^*- \left( \prod_{t=1}^{T} \E\left[\mu_{I_t}\right] \right)^{\frac{1}{T}} \\
&\leq \frac{2\mu^*\log (kT)}{\sqrt T}+4001c\log 2T\sqrt{\frac{ \mu^*k \log T }{T} }+\frac{4001\log (256k)(\log T)^2\log 2T}{\epsilon T}+\frac{4k}{T}\frac{\alpha (\log T)^2\log 2T}{\epsilon }\\
& \qquad \qquad+\frac{16k(\log T)^\frac{3}{2}\log 2T}{T}\sqrt{\frac{2\alpha}{\epsilon}}\\
    & = O\bigg(\sqrt{\frac{k\log T}{T}}\log T+\frac{k(\log T)^3}{T\epsilon}\bigg)
\end{align*}
The theorem stands proved.


\hfill\\
Next, we establish the proof of Theorem \ref{thm:anytime-nucb_ldp} while using the exact same notation. Note that claims \ref{lem:window}, \ref{claim:we}, \ref{claim3:anytime} still hold with similar notation. Futher, recall that $h^*$ denotes the smallest value of $h$ for which $W_h\geq \sqrt{T}$. The following lemma provides a bound on the Nash social welfare accumulated by Algorithm \ref{algo:ncb:anytime} (when executing LDP-NCB) in an epoch $h \geq h^*$. 
\begin{restatable}{lemma}{LemmaDPNCBAnytimeLDP}
\label{conditional:NCB:anytime_ldp}
In any multi-armed bandit instance where the optimal mean satisfies
\[
\mu^* \;\geq\; \frac{512\sqrt{k\log T}}{\sqrt{T}} + \frac{10000\,\log(256k)\,(\log T)^2}{\epsilon \sqrt T},
\]
the following inequality holds for every epoch \(h \geq h^*\) and all rounds 
\[
r \in \{R_h + 1, R_h + 2, \ldots, R_{h+1}\}:
\]
    \begin{align*}
        \left(\prod_{t= R_h + 1}^{r}\E\left[\mu_{I_t}\right]\right)^\frac{1}{T} & \geq (\mu^*)^\frac{r-R_h}{T}\bigg(1-4001c\sqrt{\frac{ k \log T }{\mu^*T} }-\frac{4000\log (256k)(\log T)^2}{(\mu^*)^2 \epsilon^2 T} -\frac{4\sqrt{8k\alpha\log T}}{\sqrt T\mu^*\epsilon} -\frac{(2\alpha)^{\frac{1}{4}}(k\log T)^{\frac{3}{4}}}{T^\frac{3}{4}\mu^*\sqrt{\epsilon}}\bigg).  
    \end{align*}    
\end{restatable}
\begin{proof}

Consider any epoch \(h \ge h^*\), and let \(F_h\) be the event that Algorithm~\ref{algo:ncb:anytime} invokes the LDP-NCB procedure (Algorithm~\ref{algo:ldp-ncb}) during epoch \(h\) (see Line~\ref{step:callsubroutine}).  Note that
\[
\prob(F_h) \;=\; 1 - \frac{1}{W_h^2}.
\]
Furthermore, by the definition of \(h^*\) and Claim~\ref{claim:we}, the epoch length satisfies
\[
\sqrt{T} \;\le\; W_h \;\le\; T.
\]
Therefore, applying Lemma~\ref{lem:modified_ncb} conditioned on \(F_h\) yields:

\begingroup
\allowdisplaybreaks
    \begin{align*}
    &\left(\prod_{t= R_h+1}^{r}\E\left[\mu_{I_t}\right]\right)^\frac{1}{T} 
    \geq \left(\prod_{t= R_h+1}^{r}\E\left[\mu_{I_t} | F_h \right] \ \prob \{ F_h \} \right) ^\frac{1}{T}
    \\&\geq \left(\prod_{t= R_h+1}^{r}\E\left[\mu_{I_t} | F_h \right] \left( 1- \frac{1}{W_h^2}\right) \right) ^\frac{1}{T} \geq \left( 1- \frac{1}{W_h^2}\right)  \left(\prod_{t= R_h+1}^{r}\E\left[\mu_{I_t} | F_h \right]  \right) ^\frac{1}{T}\\
    &\geq (\mu^*)^\frac{r-R_h}{T}\left( 1- \frac{1}{W_h^2}\right) \geq (\mu^*)^{\frac{w}{T}}\bigg(1-4000c\sqrt{\frac{ k \log T }{\mu^*T} }-\frac{4000\log (256k)(\log T)^2}{(\mu^*)^2 \epsilon^2 T} -\frac{4\sqrt{8k\alpha\log T}}{\sqrt T\mu^*\epsilon} -\frac{(2\alpha)^{\frac{1}{4}}(k\log T)^{\frac{3}{4}}}{T^\frac{3}{4}\mu^*\sqrt{\epsilon}}\bigg)  \tag{Lemma \ref{lem:modified_ncb_ldp}}\\
    &\geq (\mu^*)^\frac{r-R_h}{T}\left( 1- \frac{1}{T}\right)\bigg(1-4000c\sqrt{\frac{ k \log T }{\mu^*T} }-\frac{4000\log (256k)(\log T)^2}{(\mu^*)^2 \epsilon^2 T} -\frac{4\sqrt{8k\alpha\log T}}{\sqrt T\mu^*\epsilon} -\frac{(2\alpha)^{\frac{1}{4}}(k\log T)^{\frac{3}{4}}}{T^\frac{3}{4}\mu^*\sqrt{\epsilon}}\bigg) \tag{since $W_h \geq \sqrt{T}$}\\
   &\geq (\mu^*)^\frac{r-R_h}{T}\bigg(1-4000c\sqrt{\frac{ k \log T }{\mu^*T} }-\frac{4000\log (256k)(\log T)^2}{(\mu^*)^2 \epsilon^2 T} -\frac{4\sqrt{8k\alpha\log T}}{\sqrt T\mu^*\epsilon} -\frac{(2\alpha)^{\frac{1}{4}}(k\log T)^{\frac{3}{4}}}{T^\frac{3}{4}\mu^*\sqrt{\epsilon}}-\frac{1}{T}\bigg)\\ 
   &\geq (\mu^*)^\frac{r-R_h}{T}\bigg(1-4001c\sqrt{\frac{ k \log T }{\mu^*T} }-\frac{4000\log (256k)(\log T)^2}{(\mu^*)^2 \epsilon^2 T} -\frac{4\sqrt{8k\alpha\log T}}{\sqrt T\mu^*\epsilon} -\frac{(2\alpha)^{\frac{1}{4}}(k\log T)^{\frac{3}{4}}}{T^\frac{3}{4}\mu^*\sqrt{\epsilon}}\bigg).  	
    \end{align*}
\endgroup
The lemma stands proved 
\end{proof}

\subsubsection{Proof of Theorem \ref{thm:anytime-nucb_ldp}}
\label{section:proof-of-anytime_ldp}
To prove Theorem \ref{thm:anytime-nucb_ldp}, we restrict attention to instances where
\[
\mu^* \;\ge\; \frac{512\sqrt{k\log T}}{\sqrt T}
\;+\;\frac{10000\,\log(256k)\,(\log T)^2}{\epsilon\,\sqrt T},
\]
since otherwise the claimed Nash‐regret bounds hold immediately.  Let \(h^*\) be the smallest epoch index for which \(W_{h^*}\ge\sqrt{T}\), and split the time horizon into the first \(R_{h^*}\) rounds and the remainder.

Observe that for every epoch \(g\le h^*\), we have \(W_g<\sqrt T\).  Hence, in each such epoch the algorithm invokes uniform sampling with probability
\[
\frac1{W_g^2}\;\ge\;\frac1T,
\]
so that for all \(t\le R_{h^*}\),
\[
p_t \;=\;\prob\{\text{Phase I at round }t\}
\;\ge\;\frac1T.
\]
Therefore, for all rounds \(t\le R_{h^*}\), $$\E \left[\mu_{I_t} \right] \geq \frac{\mu^*}{k} \frac{1}{T}$$ The above bound yields

    \begin{align}
        \left(\prod_{t=1}^{R_{h^*}} \E\left[\mu_{I_t}\right]\right)^{\frac{1}{T}} &\geq \left( \frac{\mu^*} {kT}\right)^\frac{R_{h^*}}{T}\nonumber 
= (\mu^*)^{\frac{R_{h^*}}{T}} \left( \frac{1}{2}\right)^{\frac{R_{h^*}\log{(kT)}}{T}}\nonumber \geq (\mu^*)^{\frac{R_{h^*}}{T}}  \left( 1 - \frac{R_{h^*}\log{(kT)}}{T} \right) \tag{via Claim \ref{lem:binomial}}\\
& \geq   (\mu^*)^{\frac{R_{h^*}}{T}}  \left( 1 - \frac{2\log{(kT)}}{\sqrt{T}} \right) \label{ineq:anytimeFirstPart_ldp}
    \end{align}
Here, the final inequality is justified by Claim~\ref{claim3:anytime}.

To handle the remaining \(T - R_{h^*}\) rounds, we conduct an epoch-wise analysis and apply Lemma~\ref{conditional:NCB:anytime_ldp}. Specifically,
\begingroup
\allowdisplaybreaks
    \begin{align}
        &\left(\prod_{t=R_{h^*}+1}^{T} \E\left[\mu_{I_t}\right]\right)^{\frac{1}{T}}
        =  \left(\prod_{h=h^*}^{e-1} \ \prod_{t= R_h + 1}^{R_{(h+1)}}\E\left[\mu_{I_t}\right]\right)^\frac{1}{T} \cdot    \left(\prod_{t= R_{e}+1}^{T}\E\left[\mu_{I_t}\right]\right)^\frac{1}{T} \nonumber \\
       & =  \prod_{h=h^*}^{e-1} \left( \prod_{t= R_h+1}^{R_h+W_h}\E\left[\mu_{I_t}\right]\right)^\frac{1}{T} \cdot   \left(\prod_{t= R_{e}+1}^{T}\E\left[\mu_{I_t}\right]\right)^\frac{1}{T} \nonumber \\
        &\geq \prod_{h=h^*}^{e-1} (\mu^*)^{\frac{W_h}{T}} \geq (\mu^*)^\frac{r-R_h}{T}\bigg(1-4001c\sqrt{\frac{ k \log T }{\mu^*T} }-\frac{4000\log (256k)(\log T)^2}{(\mu^*)^2 \epsilon^2 T} -\frac{4\sqrt{8k\alpha\log T}}{\sqrt T\mu^*\epsilon} -\frac{(2\alpha)^{\frac{1}{4}}(k\log T)^{\frac{3}{4}}}{T^\frac{3}{4}\mu^*\sqrt{\epsilon}}\bigg)\nonumber\\
&\qquad\qquad \cdot (\mu^*)^{\frac{T-R_e}{T}} \geq (\mu^*)^\frac{r-R_h}{T}\bigg(1-4001c\sqrt{\frac{ k \log T }{\mu^*T} }-\frac{4000\log (256k)(\log T)^2}{(\mu^*)^2 \epsilon^2 T} -\frac{4\sqrt{8k\alpha\log T}}{\sqrt T\mu^*\epsilon} -\frac{(2\alpha)^{\frac{1}{4}}(k\log T)^{\frac{3}{4}}}{T^\frac{3}{4}\mu^*\sqrt{\epsilon}}\bigg) \tag{via Lemma \ref{conditional:NCB:anytime_ldp}} \\
        &\geq  (\mu^*)^{1 - \frac{R_{h^*}}{T}} \prod_{j=1}^{e} \geq (\mu^*)^\frac{r-R_h}{T}\bigg(1-4001c\sqrt{\frac{ k \log T }{\mu^*T} }-\frac{4000\log (256k)(\log T)^2}{(\mu^*)^2 \epsilon^2 T} -\frac{4\sqrt{8k\alpha\log T}}{\sqrt T\mu^*\epsilon} -\frac{(2\alpha)^{\frac{1}{4}}(k\log T)^{\frac{3}{4}}}{T^\frac{3}{4}\mu^*\sqrt{\epsilon}}\bigg) \nonumber \\ 
        &\geq (\mu^*)^{1 - \frac{R_{h^*}}{T}}  \geq (\mu^*)^\frac{r-R_h}{T}\bigg(1-4001c\sqrt{\frac{ k \log T }{\mu^*T} }-\frac{4000\log (256k)(\log T)^2}{(\mu^*)^2 \epsilon^2 T} -\frac{4\sqrt{8k\alpha\log T}}{\sqrt T\mu^*\epsilon} -\frac{(2\alpha)^{\frac{1}{4}}(k\log T)^{\frac{3}{4}}}{T^\frac{3}{4}\mu^*\sqrt{\epsilon}}\bigg) ^ {\log (2T)} \tag{since $e\leq \log T+1$}\\
        &\geq   (\mu^*)^{1 - \frac{R_{h^*}}{T}}  \bigg(1-4001c\log 2T\sqrt{\frac{ k \log T }{\mu^*T} }-\frac{4000\log (256k)(\log T)^2\log 2T}{(\mu^*)^2 \epsilon^2 T} -\frac{4\log 2T\sqrt{8k\alpha\log T}}{\sqrt T\mu^*\epsilon}\nonumber\\
        &\hspace{28em}-\frac{(2\alpha)^{\frac{1}{4}}(k\log T)^{\frac{3}{4}}\log 2T}{T^\frac{3}{4}\mu^*\sqrt{\epsilon}}\bigg) \label{ineq:anytimeSecondPart_ldp}
    \end{align}
\endgroup
By applying the elementary inequality
\[
(1 - x)(1 - y) \;\ge\; 1 \;-\; x \;-\; y
\quad\text{for all }x,y \ge 0,
\]
we immediately recover the last inequality.  Next, we substitute the bound from inequality~\eqref{ineq:anytimeFirstPart_ldp}, which governs the algorithm’s performance up to round $R_{h^*}$, together with the bound from inequality~\eqref{ineq:anytimeSecondPart_ldp}, which covers the remaining rounds $R_{h^*}+1$ through $T$.  Merging these two estimates yields a single, unified lower bound on the Nash social welfare achieved by Algorithm~\ref{algo:ncb:anytime}, when executing LDP-NCB inside the epochs:  
    \begin{align*}
        &\left(\prod_{t=1}^{T} \E\left[\mu_{I_t}\right] \right)^{\frac{1}{T}} = \left(\prod_{t=1}^{R_{h^*}} \E\left[\mu_{I_t}\right]\right)^{\frac{1}{T}} \left(\prod_{t=R_{h^*}+1}^{T} \E\left[\mu_{I_t}\right]\right)^{\frac{1}{T}} \\
        &\geq \mu^*  \left( 1 - \frac{2\log{(kT)}}{\sqrt{T}} \right) \bigg(1-4001c\log 2T\sqrt{\frac{ k \log T }{\mu^*T} }-\frac{4000\log (256k)(\log T)^2\log 2T}{(\mu^*)^2 \epsilon^2 T} -\frac{4\log 2T\sqrt{8k\alpha\log T}}{\sqrt T\mu^*\epsilon} \\&\qquad\qquad\qquad\qquad\qquad\qquad-\frac{(2\alpha)^{\frac{1}{4}}(k\log T)^{\frac{3}{4}}\log 2T}{T^\frac{3}{4}\mu^*\sqrt{\epsilon}} \bigg) \\ 
        &\geq \mu^* \bigg(1 - \frac{2\log{(kT)}}{\sqrt{T}}-4001c\log 2T\sqrt{\frac{ k \log T }{\mu^*T} }-\frac{4000\log (256k)(\log T)^2\log 2T}{(\mu^*)^2 \epsilon^2 T} -\frac{4\log 2T\sqrt{8k\alpha\log T}}{\sqrt T\mu^*\epsilon} \\
        &\qquad\qquad-\frac{(2\alpha)^{\frac{1}{4}}(k\log T)^{\frac{3}{4}}\log 2T}{T^\frac{3}{4}\mu^*\sqrt{\epsilon}}\bigg).
    \end{align*}
Thus, Algorithm~\ref{algo:ncb:anytime} achieves the following upper bound on its Nash regret (when LDP-NCB is executed):  
\begin{align*}
\NRg_T& = \mu^*- \left( \prod_{t=1}^{T} \E\left[\mu_{I_t}\right] \right)^{\frac{1}{T}} \leq  \frac{2\mu^*\log{(kT)}}{\sqrt{T}} + 4001c\log 2T\sqrt{\frac{ \mu^*k \log T }{T} } + \frac{4000\log (256k)(\log T)^2\log 2T}{\mu^* \epsilon^2 T} \\ &\qquad \qquad \qquad \qquad \qquad \quad+ \frac{4\log 2T\sqrt{8k\alpha\log T}}{\sqrt T\epsilon}+\frac{(2\alpha)^{\frac{1}{4}}(k\log T)^{\frac{3}{4}}\log 2T}{T^\frac{3}{4}\sqrt{\epsilon}}\\
& \leq  \frac{2\mu^*\log{(kT)}}{\sqrt{T}} + 4001c\log 2T\sqrt{\frac{ \mu^*k \log T }{T} } + \frac{4(\log T)\log 2T}{10 \epsilon \sqrt T} \\ &\qquad \qquad \qquad \qquad \qquad \quad+ \frac{4\log 2T\sqrt{8k\alpha\log T}}{\sqrt T\epsilon}+\frac{(2\alpha)^{\frac{1}{4}}(k\log T)^{\frac{3}{4}}\log 2T}{T^\frac{3}{4}\sqrt{\epsilon}} \tag{$\mu^* \geq \frac{10000\log (256k)(\log T)}{\epsilon \sqrt T}$}\\
&=O\bigg(\sqrt{\frac{k\log T}{T}}\log T + \frac{\sqrt k (\log T)^2}{\epsilon\sqrt{T}}\bigg).\tag{$\mu^* \leq 1$}
\end{align*}
The theorem stands proved. 


\begin{thebibliography}{25}
\providecommand{\natexlab}[1]{#1}
\providecommand{\url}[1]{\texttt{#1}}
\expandafter\ifx\csname urlstyle\endcsname\relax
  \providecommand{\doi}[1]{doi: #1}\else
  \providecommand{\doi}{doi: \begingroup \urlstyle{rm}\Url}\fi

\bibitem[Azize and Basu(2022)]{azize2022privacy}
Achraf Azize and Debabrota Basu.
\newblock When privacy meets partial information: A refined analysis of differentially private bandits.
\newblock \emph{Advances in Neural Information Processing Systems}, 35:\penalty0 32199--32210, 2022.

\bibitem[Barman et~al.(2023)Barman, Khan, Maiti, and Sawarni]{barman2023fairness}
Siddharth Barman, Arindam Khan, Arnab Maiti, and Ayush Sawarni.
\newblock Fairness and welfare quantification for regret in multi-armed bandits.
\newblock In \emph{Proceedings of the AAAI Conference on Artificial Intelligence}, volume~37, pages 6762--6769, 2023.

\bibitem[Basu et~al.(2019)Basu, Dimitrakakis, and Tossou]{basu2019differential}
Debabrota Basu, Christos Dimitrakakis, and Aristide Tossou.
\newblock Differential privacy for multi-armed bandits: What is it and what is its cost?
\newblock \emph{arXiv preprint arXiv:1905.12298}, 2019.

\bibitem[Bubeck et~al.(2012)Bubeck, Cesa-Bianchi, et~al.]{bubeck2012regret}
S{\'e}bastien Bubeck, Nicolo Cesa-Bianchi, et~al.
\newblock Regret analysis of stochastic and nonstochastic multi-armed bandit problems.
\newblock \emph{Foundations and Trends{\textregistered} in Machine Learning}, 5\penalty0 (1):\penalty0 1--122, 2012.

\bibitem[Chan et~al.(2011)Chan, Shi, and Song]{chan2011private}
T-H~Hubert Chan, Elaine Shi, and Dawn Song.
\newblock Private and continual release of statistics.
\newblock \emph{ACM Transactions on Information and System Security (TISSEC)}, 14\penalty0 (3):\penalty0 1--24, 2011.

\bibitem[Clement et~al.(2015)Clement, Roy, Oudeyer, and Lopes]{clement2015multi}
Benjamin Clement, Didier Roy, Pierre-Yves Oudeyer, and Manuel Lopes.
\newblock Multi-armed bandits for intelligent tutoring systems.
\newblock \emph{Journal of Educational Data Mining}, 7\penalty0 (2), 2015.

\bibitem[Dubhashi and Panconesi(2009)]{dubhashi2009concentration}
Devdatt~P Dubhashi and Alessandro Panconesi.
\newblock \emph{Concentration of measure for the analysis of randomized algorithms}.
\newblock Cambridge University Press, 2009.

\bibitem[Dwork and Roth(2014)]{dwork2014}
Cynthia Dwork and Aaron Roth.
\newblock The algorithmic foundations of differential privacy.
\newblock \emph{Foundations and Trends in Theoretical Computer Science}, 9\penalty0 (3-4):\penalty0 211--407, 2014.

\bibitem[Dwork et~al.(2009)Dwork, Naor, Reingold, Rothblum, and Vadhan]{dwork2009complexity}
Cynthia Dwork, Moni Naor, Omer Reingold, Guy~N Rothblum, and Salil Vadhan.
\newblock On the complexity of differentially private data release: efficient algorithms and hardness results.
\newblock In \emph{Proceedings of the forty-first annual ACM symposium on Theory of computing}, pages 381--390, 2009.

\bibitem[Hu et~al.(2021)Hu, Huang, and Mehta]{hu2021optimal}
Bingshan Hu, Zhiming Huang, and Nishant~A Mehta.
\newblock Optimal algorithms for private online learning in a stochastic environment.
\newblock \emph{arXiv preprint arXiv:2102.07929}, 2021.

\bibitem[Krishna et~al.(2025)Krishna, John, Barik, and Tan]{krishna2025p}
Anand Krishna, Philips~George John, Adarsh Barik, and Vincent~YF Tan.
\newblock p-mean regret for stochastic bandits.
\newblock In \emph{Proceedings of the AAAI Conference on Artificial Intelligence}, volume~39, pages 17966--17973, 2025.

\bibitem[Lattimore and Szepesv´ari(2020)]{lattimorebanditalgorithms}
Tor Lattimore and Csaba Szepesv´ari.
\newblock \emph{Bandit Algorithms}.
\newblock Cambridge University Press, 2020.

\bibitem[Li et~al.(2010)Li, Chu, Langford, and Schapire]{li2010contextual}
Lihong Li, Wei Chu, John Langford, and Robert~E Schapire.
\newblock A contextual-bandit approach to personalized news article recommendation.
\newblock In \emph{Proceedings of the 19th international conference on World wide web}, pages 661--670, 2010.

\bibitem[Mishra and Thakurta(2015)]{mishra2015nearly}
Nikita Mishra and Abhradeep Thakurta.
\newblock (nearly) optimal differentially private stochastic multi-arm bandits.
\newblock In \emph{Proceedings of the Thirty-First Conference on Uncertainty in Artificial Intelligence}, pages 592--601, 2015.

\bibitem[Moulin(2004)]{moulin2004fair}
Herv{\'e} Moulin.
\newblock \emph{Fair division and collective welfare}.
\newblock MIT press, 2004.

\bibitem[Pan et~al.(2019)Pan, Wang, Zhang, Li, Yi, and Song]{pan2019you}
Xinlei Pan, Weiyao Wang, Xiaoshuai Zhang, Bo~Li, Jinfeng Yi, and Dawn Song.
\newblock How you act tells a lot: Privacy-leaking attack on deep reinforcement learning.
\newblock In \emph{Aamas}, volume~19, pages 368--376, 2019.

\bibitem[Sawarni et~al.(2023)Sawarni, Pal, and Barman]{sawarni2023nash}
Ayush Sawarni, Soumyabrata Pal, and Siddharth Barman.
\newblock Nash regret guarantees for linear bandits.
\newblock \emph{Advances in Neural Information Processing Systems}, 36:\penalty0 33288--33318, 2023.

\bibitem[Schwartz et~al.(2017)Schwartz, Bradlow, and Fader]{schwartz2017customer}
Eric~M Schwartz, Eric~T Bradlow, and Peter~S Fader.
\newblock Customer acquisition via display advertising using multi-armed bandit experiments.
\newblock \emph{Marketing Science}, 36\penalty0 (4):\penalty0 500--522, 2017.

\bibitem[Shariff and Sheffet(2018)]{shariff2018differentially}
Roshan Shariff and Or~Sheffet.
\newblock Differentially private contextual linear bandits.
\newblock \emph{Advances in Neural Information Processing Systems}, 31, 2018.

\bibitem[Tenenbaum et~al.(2021)Tenenbaum, Kaplan, Mansour, and Stemmer]{tenenbaum2021differentially}
Jay Tenenbaum, Haim Kaplan, Yishay Mansour, and Uri Stemmer.
\newblock Differentially private multi-armed bandits in the shuffle model.
\newblock \emph{Advances in Neural Information Processing Systems}, 34:\penalty0 24956--24967, 2021.

\bibitem[Tewari and Murphy(2017)]{tewari2017ads}
Ambuj Tewari and Susan~A Murphy.
\newblock From ads to interventions: Contextual bandits in mobile health.
\newblock In \emph{Mobile health: sensors, analytic methods, and applications}, pages 495--517. Springer, 2017.

\bibitem[Thompson(1933)]{thompson1933likelihood}
William~R Thompson.
\newblock On the likelihood that one unknown probability exceeds another in view of the evidence of two samples.
\newblock \emph{Biometrika}, 25\penalty0 (3/4):\penalty0 285--294, 1933.

\bibitem[Tossou and Dimitrakakis(2016)]{tossou2016algorithms}
Aristide Tossou and Christos Dimitrakakis.
\newblock Algorithms for differentially private multi-armed bandits.
\newblock In \emph{Proceedings of the AAAI Conference on Artificial Intelligence}, volume~30, 2016.

\bibitem[Villar et~al.(2015)Villar, Bowden, and Wason]{villar2015multi}
Sof{\'\i}a~S Villar, Jack Bowden, and James Wason.
\newblock Multi-armed bandit models for the optimal design of clinical trials: benefits and challenges.
\newblock \emph{Statistical science: a review journal of the Institute of Mathematical Statistics}, 30\penalty0 (2):\penalty0 199, 2015.

\bibitem[Wenbo~Ren and Shroff(2020)]{ren2020localdifferentialprivacy}
Jia~Liu Wenbo~Ren, Xingyu~Zhou and Ness~B. Shroff.
\newblock Multi-armed bandits with local differential privacy, 2020.
\newblock URL \url{https://arxiv.org/abs/2007.03121}.

\end{thebibliography}
\end{document}